\documentclass[letterpaper]{article}
\usepackage{aaai2027}
\usepackage{helvet}
\usepackage{courier}
\usepackage{graphicx}
\usepackage{amsmath}
\usepackage{amssymb}
\usepackage{booktabs}
\usepackage{multirow}
\usepackage{xcolor}
\usepackage{microtype}
\usepackage{natbib}
\usepackage{dblfloatfix}
\usepackage{caption}
\usepackage{url}
\usepackage{longtable}

\newcommand{\methodtypes}[1]{\textsuperscript{\ensuremath{#1}}}

\title{OVEarth-Bench: Evaluating Category Breadth and Query Diversity for Open-Vocabulary Earth Observation}

\author{
    Kaiyu Li,
    Zepeng Xin,
    Zixuan Jiang,
    Jing Fu,
    Lanxuan Xue,
    Lingyu Zhang,
    Xiangyong Cao\corresponding
}
\affiliations{
    Xi'an Jiaotong University

}

\begin{document}

\maketitle



\begin{abstract}
Open-vocabulary Earth observation (EO) aims to localize geospatial concepts specified in natural language rather than a fixed label set. Existing benchmarks, however, usually cover narrow category vocabularies or limited query forms. To fill this gap, we introduce OVEarth-Bench, which extends existing evaluation in two directions: \emph{category breadth}, through broad hierarchical category coverage with positive and negative expressions, and \emph{query diversity}, through vocabulary, referring, and reasoning queries. The benchmark supports mask and box localization under a unified zero-shot protocol. We evaluate a broad set of general and EO-specific methods. The evaluation reveals that: (1) the performance of current methods remains limited, while broader category coverage yields more stable model rankings; (2) MLLM-based methods achieve the strongest overall performance; and (3) EO-specific methods generally underperform general models and rarely match the strongest methods. These findings provide guidance for future open-vocabulary EO method design and highlight the importance of developing more realistic, diverse, high-quality, and large-scale benchmarks for reliable evaluation. Our data and evaluation package are released at \url{https://earth-insights.github.io/OVEarth-bench}.
\end{abstract}

\section{Introduction}

Earth observation (EO) images encompass a broad semantic space, covering land cover, transportation networks, industrial facilities, public services, and agricultural structures at diverse scales and orientations. Open-vocabulary EO perception is therefore important for localizing concepts specified by users beyond a predefined label set~\cite{li2025annotation}.

However, existing open-vocabulary EO benchmarks cannot comprehensively assess this capability. They commonly reuse conventional EO datasets with limited category spaces~\cite{lam2018xview,wang2021loveda,zhang2021shiprsimagenet,xia2023openearthmap}, providing insufficient coverage of fine-grained and long-tail EO concepts. Moreover, most benchmarks support only limited query forms and cannot evaluate vocabulary phrases, referring expressions, and reasoning queries simultaneously~\cite{zhan2023rsvg,dong2024cross,yuan2024rrsis,li2025segearthr1,xin2026segearthr2}. Consequently, they provide an incomplete assessment on whether a model possesses genuine open-vocabulary EO capability.

To address these limitations, we introduce a new open-vocabulary EO benchmark (i.e., OVEarth-Bench), built from newly collected EO images. Besides, OVEarth-Bench differs from existing benchmarks from two aspects, i.e., \emph{category breadth} and \emph{query diversity}. Specifically, it evaluates \emph{category breadth} through a hierarchical taxonomy of 172 categories and 1,346 unique vocabulary strings for positive and negative queries, and \emph{query diversity} through vocabulary phrases, referring expressions, and reasoning queries. Additionally, this benchmark provides manually created and verified mask annotations, together with reviewed horizontal and oriented bounding boxes (HBBs and OBBs), enabling mask and box localization under a unified zero-shot protocol.


\begin{figure}[t]
  \centering
  \captionsetup{type=figure}
  \includegraphics[width=\linewidth]{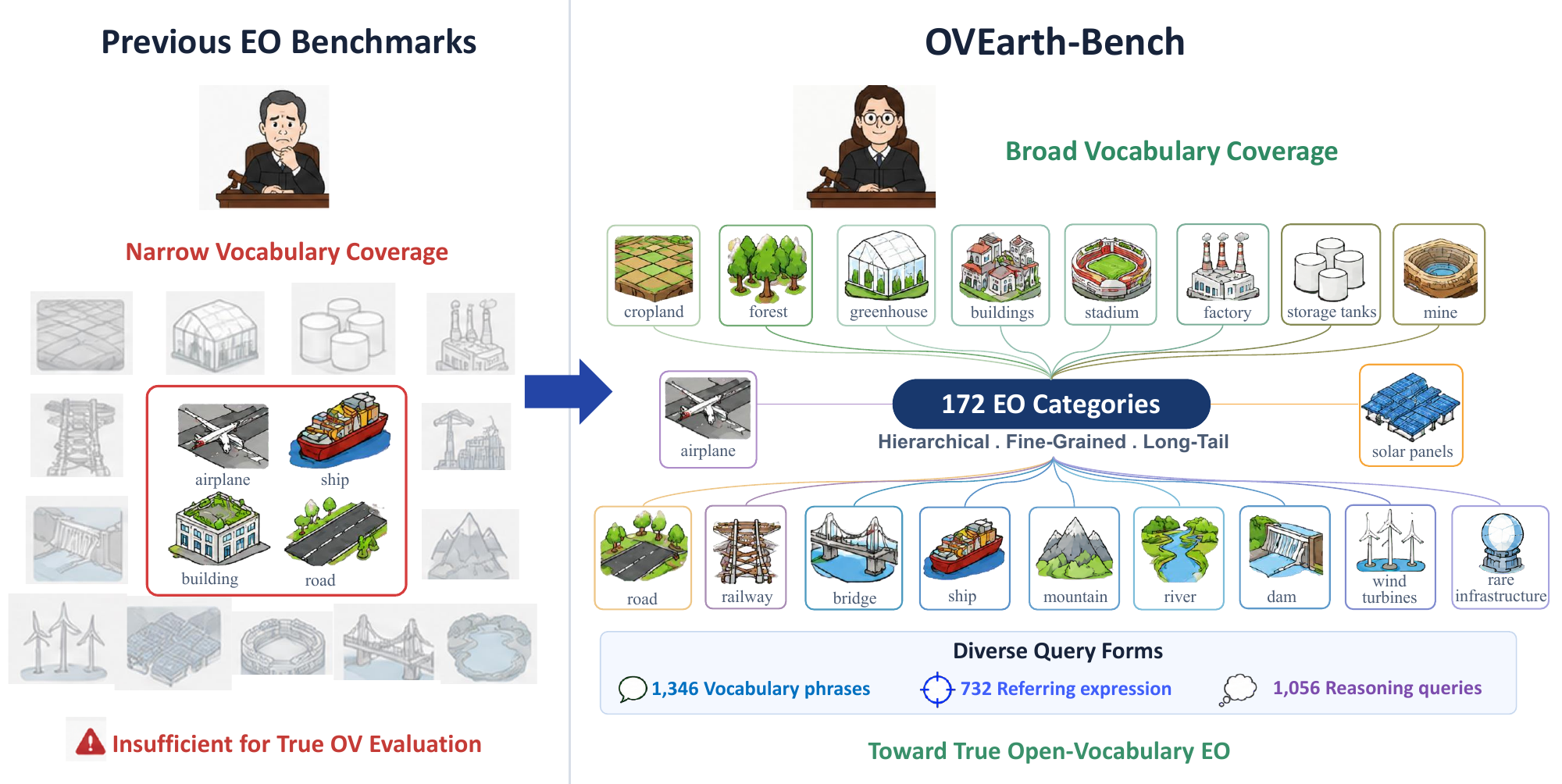}
  \captionof{figure}{Compared with previous EO benchmarks with narrow category scopes, OVEarth-Bench covers broad EO categories and evaluates diverse query forms.}
  \label{fig:motivation}
\end{figure}

\begin{figure*}[t]
  \centering
  \captionsetup{type=figure}
  \includegraphics[width=\linewidth]{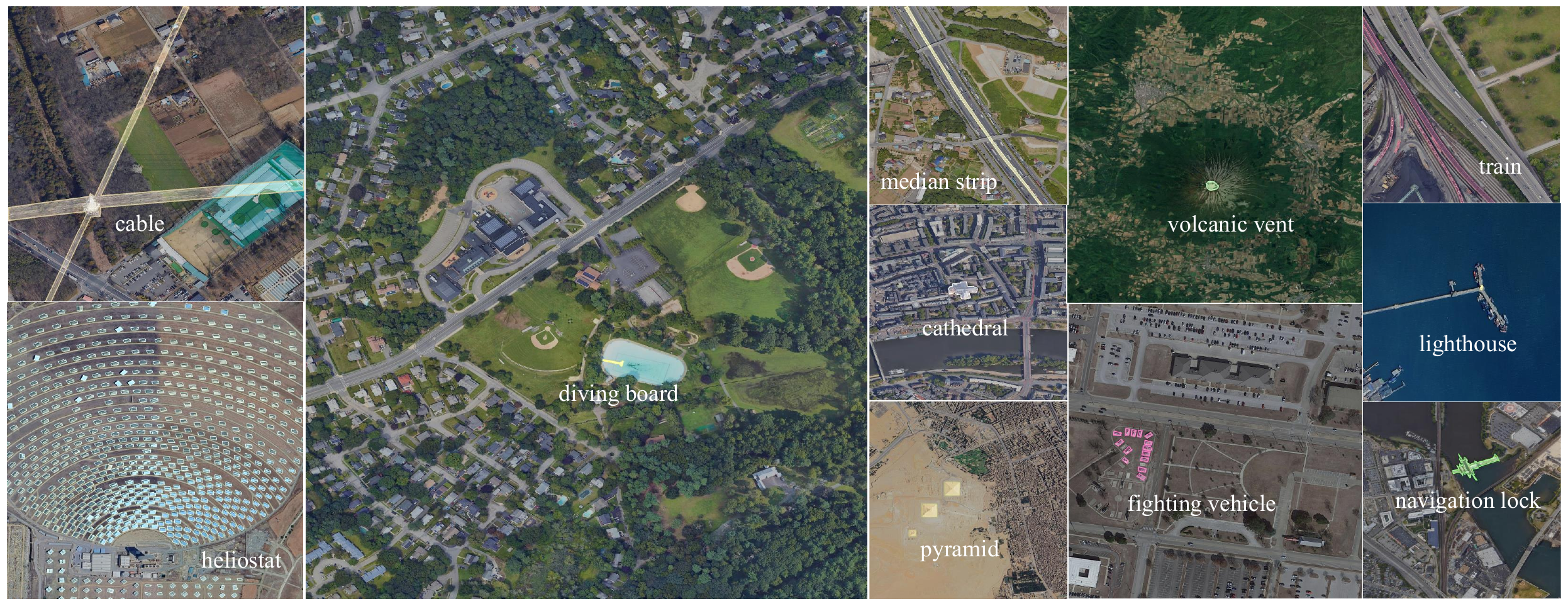}
  \captionof{figure}{Visual examples of some object categories in the OVEarth-Bench dataset. See Figure~\ref{fig:sample3} for more samples.}
  \label{fig:sample1}
\end{figure*}

Based on the OVEarth-Bench, we evaluate general contrastive segmentation, MLLM-based segmentation, and detection-and-grounding models, together with EO-specific methods spanning these approaches. The results reveal that model ranking based on narrow category sets are substantially more sensitive to category composition, MLLM-based methods dominate the leading results, and EO specialization alone does not confer a consistent advantage. Moreover, accurate localization does not necessarily imply reliable rejection of absent concepts. These observations provide guidance for future open-vocabulary EO method design. Although EO images have unique characteristics, focusing more on the broad generalization capabilities of MLLM and combining them with richer EO-specific corpora is a promising direction.

Our contributions are summarized as follows:
\begin{itemize}
  \item We formulate open-vocabulary EO evaluation along category breadth and query diversity, and show that narrow category sets can substantially affect model comparisons.
  \item We construct OVEarth-Bench from newly collected EO images, providing broad category coverage, diverse language queries, and human-annotated masks with reviewed HBBs and OBBs.
  \item We conduct a systematic zero-shot evaluation of 49 general and EO-specific model variants across segmentation and detection paradigms, providing guidance for future open-vocabulary EO method design.
\end{itemize}

\section{Related Work}

\begin{figure*}[t]
  \centering
  \includegraphics[width=\linewidth]{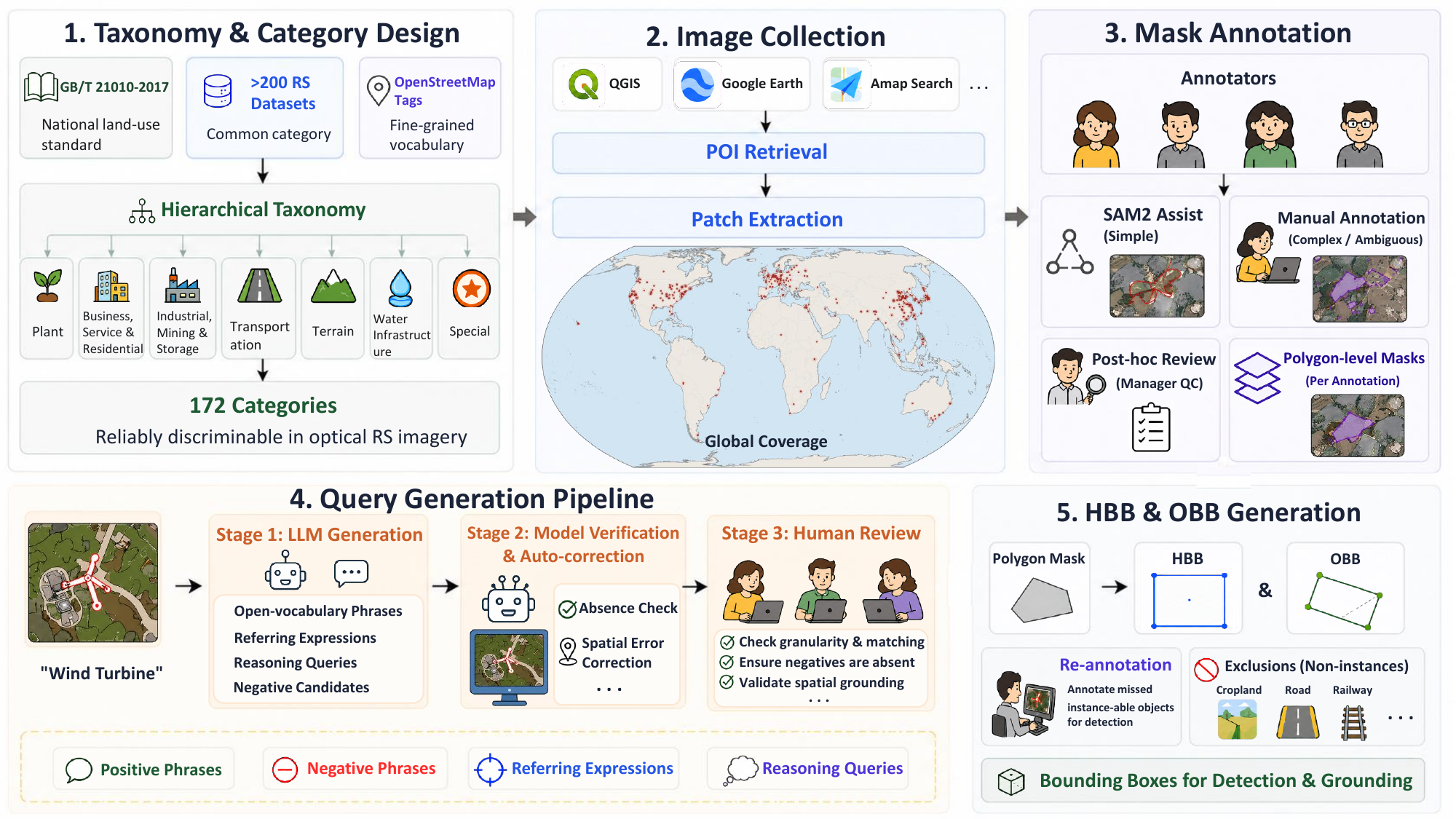}
  \caption{Construction pipeline of OVEarth-Bench: taxonomy design, POI-guided image collection, human mask annotation and review, LLM-assisted query generation with human review, and instance review and box derivation.}
  \label{fig:pipeline}
\end{figure*}

\begin{figure}[t]
  \centering
  \includegraphics[width=0.95\linewidth]{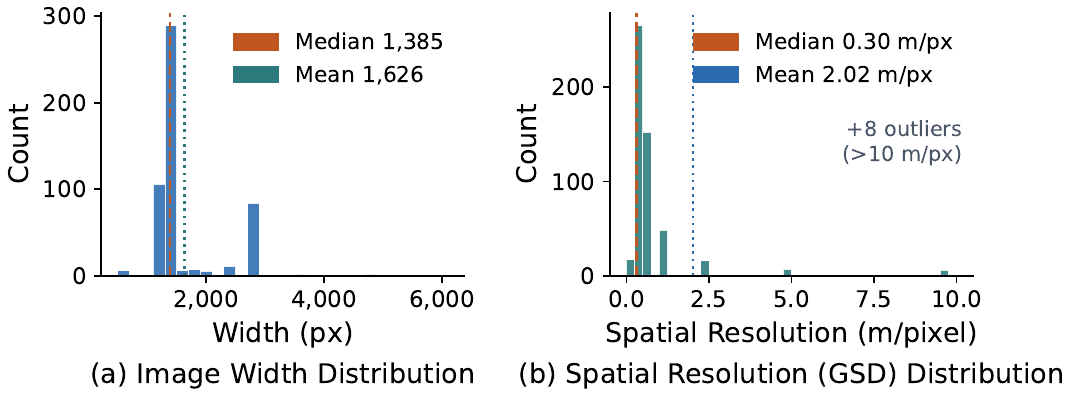}
  \caption{Image width and spatial-resolution distributions.}
  \label{fig:imgstats}
\end{figure}

\subsection{Remote-sensing Localization Benchmarks}


Traditional RS benchmarks evaluate segmentation and detection within narrow category spaces. Representative datasets support semantic segmentation~\cite{rottensteiner2012isprs,wang2021loveda,xia2023openearthmap}, box localization~\cite{cheng2014nwpu,xia2018dota,lam2018xview,li2020dior,zhang2021shiprsimagenet,sun2022fair1m}, or instance segmentation~\cite{waqas2019isaid}. Although they capture EO-specific challenges such as scale, orientation, and density variation, they remain tied to predefined taxonomies.

Language-conditioned benchmarks introduce a different evaluation axis. RSVGD studies box-level visual grounding, while RISBench~\cite{dong2024cross} and RRSIS-D/RefSegRS~\cite{yuan2024rrsis} evaluate referring segmentation. EarthReason~\cite{li2025segearthr1} and LaSeRS~\cite{xin2026segearthr2} further incorporate implicit or reasoning-oriented descriptions. These resources broaden how EO targets are specified, but they reuse public benchmark images, limiting source diversity and risking overlap with pre-training data~\cite{zhou2026data}. They also evaluate category recognition, absent-concept rejection, spatial reference, and functional reasoning in separate datasets rather than within a shared benchmark.

\subsection{Open-Vocabulary Understanding Methods}

Open-vocabulary methods use language-aligned representations to localize concepts beyond a fixed inference vocabulary. Building on CLIP's image--text alignment~\cite{radford2021clip}, dense methods such as CAT-Seg~\cite{cho2024catseg}, ClearCLIP~\cite{lan2024clearclip}, and ProxyCLIP~\cite{lan2024proxyclip} adapt aligned representations to pixel prediction. MLLM-based models such as LISA~\cite{lai2024lisa}, Sa2VA~\cite{yuan2025sa2va}, UniPixel~\cite{liu2026unipixel}, SAMTok~\cite{zhou2026samtok}, ALToLLM~\cite{wang2026alto}, and X2SAM~\cite{wang2026x2sam} couple language understanding with mask generation, while detection and grounding methods~\cite{liu2024grounding,cheng2024yoloworld,wang2026locateanything,jiang2026rex} localize regions from text.


EO-specific methods address domain characteristics such as overhead appearance, small objects, and arbitrary orientation. SegEarth-OV~\cite{li2025segearthov1}, SegEarth-OV3~\cite{li2025segearthov3}, and SkySense-O~\cite{zhu2025skysenseo} adapt open-vocabulary segmentation or vision--language pretraining to EO images; LAE-DINO~\cite{pan2024laedino}, OpenRSD~\cite{huang2025openrsd}, and SLIP-RS~\cite{wang2026sliprs} emphasize category-rich localization; and RemoteSAM~\cite{yao2025remotesam}, GeoPixel~\cite{shabbir2025geopixel}, UniGeoSeg~\cite{ni2026unigeoseg}, SegEarth-R1~\cite{li2025segearthr1}, and SegEarth-R2~\cite{xin2026segearthr2} support referring or reasoning-guided segmentation. Despite these advances, open-vocabulary EO methods are commonly evaluated on conventional or task-specific datasets, with limited category coverage or query diversity. Thus, existing results do not comprehensively characterize model generalization across broad category spaces and diverse query forms. These limitations motivate OVEarth-Bench, which evaluates both dimensions under a unified zero-shot protocol.




\section{OVEarth-Bench}

\begin{table}[t]
  \centering
  \setlength{\tabcolsep}{4pt}
  \scalebox{0.9}{
  \begin{tabular}{lccccccc}
    \toprule
    Dataset & \#Cat & \#Voc & Fresh & Ref & Rea & M & B \\
    \midrule
    DOTA-v2 & 18 & -- & \checkmark & -- & -- & -- & \checkmark \\
    DIOR & 20 & -- & \checkmark & --  & -- & -- &  \checkmark \\
    FAIR1M & 37 & -- & \checkmark & --  & -- & -- &  \checkmark \\
    NWPU & 10 & -- & \checkmark & --  & -- & -- &  \checkmark \\
    ShipRSImageNet & 50 & -- & -- & -- & --  & -- &  \checkmark \\
    xView & 60 & -- & \checkmark & -- & --  & -- &  \checkmark \\
    \midrule
    LoveDA & 7 & -- & \checkmark & -- & --  & \checkmark & -- \\
    iSAID & 16 & -- & -- & -- & --  & \checkmark & -- \\
    Potsdam & 6 & -- & \checkmark & -- & --  & \checkmark & -- \\
    Vaihingen & 6 & -- & \checkmark & -- & --  & \checkmark & -- \\
    OpenEarthMap & 9 & -- & -- & -- & --  & \checkmark & -- \\
    \midrule
    RSVGD & 20 & -- & -- & \checkmark & -- & -- & \checkmark \\
    RISBench & 26 & -- & -- & \checkmark & -- & \checkmark & \checkmark \\
    RRSIS-D & 20 & -- & -- & \checkmark & -- & \checkmark & \checkmark \\
    RefSegRS & 14 & -- & -- & \checkmark & -- & \checkmark & -- \\
    EarthReason & 28 & -- & -- & -- & \checkmark & \checkmark & -- \\
    LaSeRS & 122 & -- & -- & \checkmark & \checkmark & \checkmark & \checkmark \\
    \midrule
    \textbf{OVEarth-Bench} & \textbf{172} & \textbf{1,346} & \checkmark & \checkmark & \checkmark & \checkmark & \checkmark \\
    \bottomrule
  \end{tabular}}
  \caption{Comparison with representative EO segmentation, detection, and language-conditioned benchmarks. ``\#Cat'' follows each source's label-count; ``\#Voc'' counts unique natural-language vocabulary strings; and ``Fresh'' denotes newly collected images. ``Ref'' and ``Rea'' denote referring and reasoning queries; ``M'' and ``B'' denote mask and box annotations.}
  \label{tab:comparison}
\end{table}

In this work, open-vocabulary EO refers to zero-shot localization from natural-language queries without benchmark-specific adaptation or a fixed inference label set. To our knowledge, OVEarth-Bench is the first open-vocabulary EO benchmark to jointly evaluate category and query diversity on newly collected images with mask and box annotations. As shown in Figure~\ref{fig:pipeline}, its construction includes taxonomy design, image collection, mask annotation, query construction, and box derivation, which are detailed below.


\subsection{Taxonomy and Category Design}

We define the category hierarchy before collecting images so that sampling is guided by an explicit semantic scope rather than only by targets that are easy to find or annotate. The hierarchy draws on three complementary sources: (1) the Chinese national land-use standard GB/T~21010-2017 for a domain-level organization; (2) the category inventories of more than 200 RS datasets in EarthNets~\cite{xiong2022earthnets}, used to identify recurring concepts and gaps in existing coverage; and (3) OpenStreetMap tags\footnote{https://wiki.openstreetmap.org/wiki/Map\_Features}, used to refine practically occurring subcategories.
The resulting hierarchy organizes objects into seven top-level domains (\emph{``plant''}, \emph{``business, service and residential''}, \emph{``industrial, mining and storage''}, \emph{``transportation''}, \emph{``terrain''}, \emph{``water infrastructure''}, and \emph{``special''}\footnote{\ \emph{``special''} indicates categories like military, religion, etc.}), subdivided into thematic branches and final 172 categories (see Appendix Table~\ref{tab:category_hierarchy}). Each category was reviewed to ensure that it could be reliably distinguished in optical images.

\subsection{Image Collection}

All images for our OVEarth-Bench are newly collected rather than reused from existing RS benchmarks. Collection proceeds via point-of-interest (POI) retrieval using a custom retrieval tool with QGIS, followed by patch extraction at the retrieved coordinates.
The selected images cover six continents, with denser sampling in Asia, Europe, and North America, and fewer samples in regions where suitable high-quality images were harder to obtain (see Figure~\ref{fig:pipeline}). Several same-location seasonal pairs provide temporal variation (see Figure~\ref{fig:sample2}).
Image widths range from 692 to 4,713 pixels. The ground sampling distance (GSD) ranges from 0.08 to 305.7 m/pixel, with a median of 0.30 m/pixel.
Images are stored as GeoTIFF files with coordinate metadata to support provenance tracking and future extension. Figure~\ref{fig:imgstats} and Appendix~\ref{app:statistics} provide the detailed descriptions. The licensing and redistribution conditions will be released with benchmark.

\subsection{Mask Annotation}

Our annotators create polygon masks for the selected targets. For geometrically simple targets, SAM~2~\cite{ravi2025sam2} provides an initial mask that an annotator must inspect and correct; complex, concave, or ambiguous targets are annotated entirely manually. For non-instanceable regions, such as cropland, grassland, and road sections, annotators only draw the corresponding areas without distinguishing between instances. A separate manager reviews category assignment, boundary quality, and missed target instances, and returns uncertain cases for correction.

\subsection{Query Generation Pipeline}
\label{sec:query}



Each annotation is processed by a two-stage pipeline. Given the category name and an image with target mask, the first LLM stage generates open-vocabulary phrases, referring expressions, reasoning queries, and negative candidate phrases. Negative candidates indicate objects that are visually or conceptually similar to the target but absent from the scene; referring expressions ground the target through spatial context; reasoning prompts describe its function without naming the category. The complete prompt text and the structured output schema are given in Appendix~\ref{app:prompts}.

We identified two main failure modes in LLM-generated annotations: (1) objects proposed as absent are actually present, and (2) referring expressions contain incorrect spatial directions. The second stage serves as automated pre-screening by presenting the model with the raw image and the generated candidates to verify absence and correct spatial errors. Across all annotations, Stage~2 flagged 3.9\% of negative candidates as present in the image and auto-corrected 28.6\% of referring expressions for spatial errors. These automated checks are not treated as final validation, and all candidate language annotations are subsequently reviewed manually.

Three expert annotators then perform a final quality pass, cross-checking that OV positive phrases match the mask at the correct granularity, that negative phrases are unambiguously absent, and that each referring and reasoning query uniquely identifies the full annotated target region and cannot be resolved to another visible object. When an annotation contains multiple instances of the target category, the query must not describe only a subset of them. Human reviewers changed or removed 57.3\% of referring expressions and modified 16.0\% of negative phrases. The high revision rate for referring expressions reflects both conservative quality control and the difficulty of judging spatial relations in RS.

\subsection{HBB and OBB Annotations}

To further support detection and grounding tasks, we first review and refine the polygon annotations of instanceable targets. We then derive each HBB from the axis-aligned envelope of its polygon and compute the corresponding minimum-area OBB using the rotating-calipers algorithm~\cite{toussaint1983solving}. Area-type and linear categories (e.g., cropland, road sections, and railway) are excluded as non-instanceable; see Appendix~\ref{app:excluded} for the full list.

\subsection{Dataset Statistics}

\begin{figure}[t]
  \centering
  \includegraphics[width=0.95\linewidth]{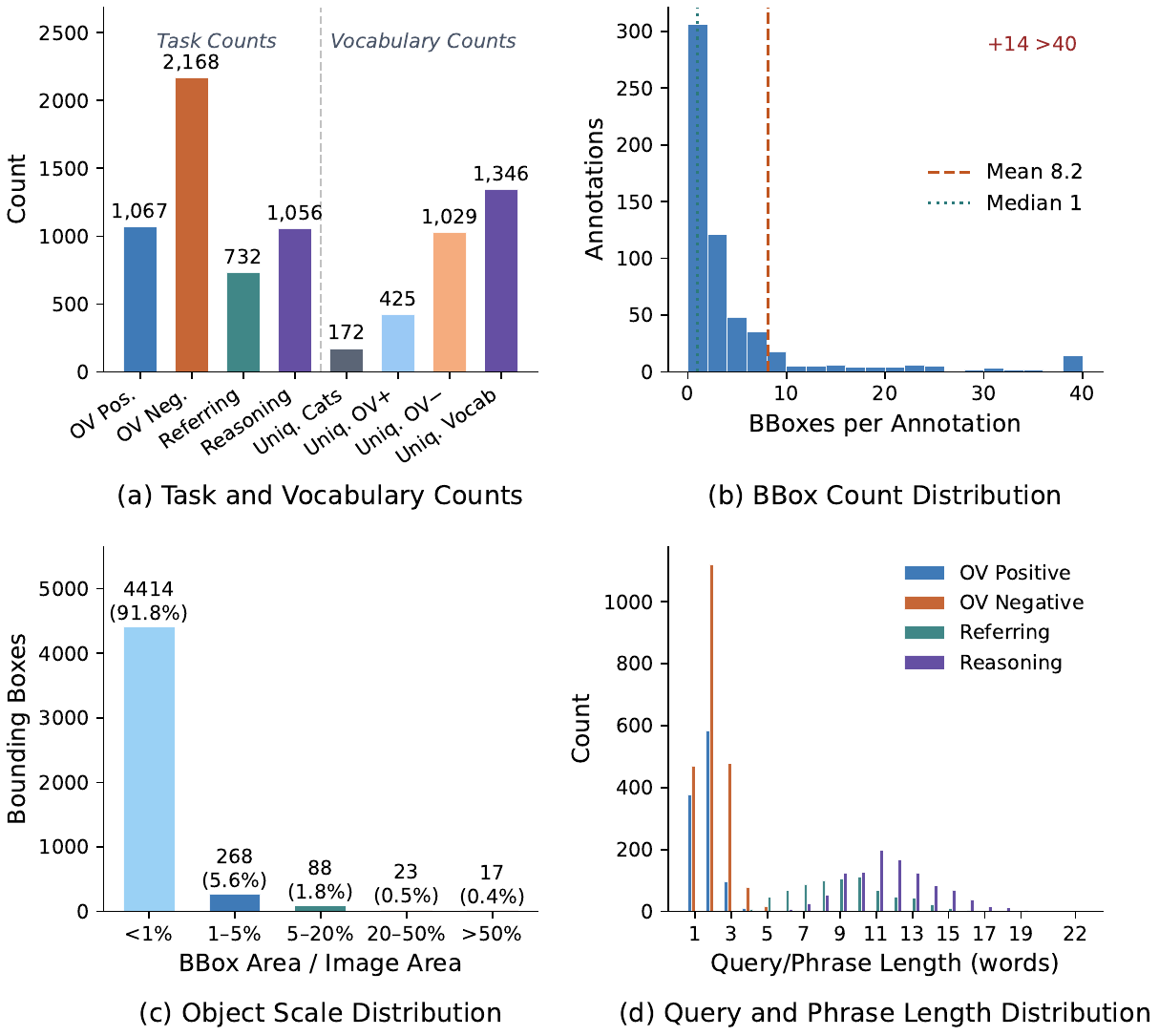}
  \caption{Annotation statistics of OVEarth-Bench.}
  \label{fig:annstats}
\end{figure}

Table~\ref{tab:comparison} compares OVEarth-Bench with representative RS segmentation, detection, and grounding benchmarks. Unlike most prior datasets, which focus on a single annotation form, OVEarth-Bench jointly provides fresh images, category names, open-vocabulary phrases, referring expressions, reasoning queries, masks, and boxes.
As shown in Figure~\ref{fig:annstats}, OVEarth-Bench contains 590 mask annotations from 520 images. Across its 172 categories, the benchmark provides 425 unique positive and 1,029 unique negative vocabulary phrases, 732 referring expressions, and 1,056 reasoning queries. These annotations evaluate complementary abilities: vocabulary-level recognition, rejection of plausible absent categories, spatial grounding, and implicit functional reasoning. The detection subset contains 4,810 boxes after excluding non-instanceable classes. Small targets predominate, with 91.8\% of boxes occupying less than 1\% of the image area. See Appendix~\ref{app:statistics} for more details.

\subsection{Evaluation Protocol}

OVEarth-Bench defines three tasks corresponding to the three query types. In the open-vocabulary task, each annotation includes positive noun phrases and negative phrases; a model must localize the target for positive queries and reject the negative queries. The referring task provides natural-language expressions with spatial context, and the reasoning task provides implicit functional descriptions without naming the category. The evaluator supports segmentation masks, HBBs, and OBBs as separate localization modalities, each scored with its corresponding metrics.

For segmentation, following prior work~\cite{lai2024lisa}, we report macro Precision, Recall, and IoU ($ma$-P, $ma$-R, and $ma$-IoU) to avoid bias toward samples with large target areas. Each query is treated as an evaluation sample because language variation is itself part of the capability assessed by OVEarth-Bench. We further report annotation- and category-macro IoU to assess the robustness of model rankings to query and category imbalance (Appendix~\ref{app:aggregation-robustness}).
We also report micro IoU ($mi$-IoU), which pools intersections and unions across positive queries. For detection, we report micro Precision, Recall, and F1 ($mi$-P$_{0.5}$, $mi$-R$_{0.5}$, and $mi$-F1$_{0.5}$), alongside the mean micro F1 ($mi$-F1$_{0.5:0.95}$) averaged across IoU thresholds from $0.5$ to $0.95$ in steps of $0.05$, following the SAM 3 protocol~\cite{carion2025sam3}. To evaluate presence/absence discrimination across positive and negative vocabulary queries, we report the Matthews Correlation Coefficient (MCC $\in [-1, 1]$), which assesses binary presence predictions regardless of localization quality. Detailed definitions are in Appendix~\ref{app:metrics}.

\section{Experiments}

\subsection{Experimental Setup}

Our evaluation includes four broad groups of general models: (1) \methodtypes{\circ} open-vocabulary segmentation models based on contrastive pre-training; (2) \methodtypes{\diamond} MLLM-based methods; and (3) \methodtypes{\triangle} open-vocabulary detection and grounding models. We additionally evaluate (4) \methodtypes{\star} EO-specific open-vocabulary models, which span segmentation, detection / grounding, and MLLM approaches and are distinguished by domain specialization rather than model type. We evaluate 49 model variants in total. All 49 models provide vocabulary outputs, while 32 and 31 models provide referring and reasoning outputs. For segmentation evaluation, outputs from detection models are converted to masks using SAM2.1-L. Although the evaluation protocol supports both HBB and OBB outputs, the box-localization results reported in this paper use HBB to provide a common output format across the evaluated models. All methods are evaluated zero-shot with respect to OVEarth-Bench, with no training, fine-tuning, or threshold selection on the benchmark. We use default hyper-parameters from the original papers or official implementations. Evaluation is performed using our open-source package \texttt{ovearth-eval}. Detailed settings and model descriptions are provided in Appendix~\ref{app:models}.

\begin{table}[t]
\centering
\scalebox{0.7}{
\begin{tabular}{lccccc}
\toprule
Model & $ma$-P & $ma$-R & \textbf{$ma$-IoU} & $mi$-IoU & MCC \\
\midrule
Text4Seg-7B\methodtypes{\diamond} & 0.30 & 0.50 & 0.25 & 0.16 & 0.00 \\
YOLO-World2-L+SAM\methodtypes{\triangle} & 2.74 & 4.70 & 1.62 & 3.82 & 0.01 \\
Qwen3-VL-4B+SAM\methodtypes{\diamond\triangle} & 5.55 & 3.09 & 2.27 & 11.50 & \textbf{0.60} \\
CLIP\methodtypes{\circ} & 5.54 & 31.83 & 4.55 & 8.73 & 0.04 \\
SLIP-RS-L+SAM\methodtypes{\triangle\star} & 13.32 & 16.17 & 8.70 & 3.98 & 0.16 \\
SCLIP\methodtypes{\circ} & 9.48 & 85.29 & 8.75 & 10.80 & 0.06 \\
SLIP-RS-T+SAM\methodtypes{\triangle\star} & 14.00 & 16.39 & 9.03 & 5.19 & 0.16 \\
GSNet\methodtypes{\circ\star} & 30.63 & 14.68 & 9.15 & 10.21 & 0.07 \\
UniGeoSeg\methodtypes{\diamond\star} & 16.27 & 14.59 & 9.26 & 12.35 & 0.04 \\
OpenRSD+SAM\methodtypes{\triangle\star} & 23.20 & 14.21 & 10.24 & 8.76 & 0.18 \\
GEM\methodtypes{\circ} & 11.79 & \textbf{89.08} & 10.42 & 11.35 & 0.07 \\
MaskCLIP\methodtypes{\circ} & 13.11 & 69.12 & 10.43 & 13.99 & 0.03 \\
Pi-Seg-v1-B\methodtypes{\circ\star} & 12.33 & 66.74 & 10.57 & 14.61 & 0.06 \\
CAT-Seg-B\methodtypes{\circ} & 12.48 & 82.24 & 10.90 & 10.94 & 0.08 \\
CASS\methodtypes{\circ} & 16.17 & 70.85 & 12.61 & 15.72 & 0.18 \\
LISA-7B\methodtypes{\diamond} & 18.93 & 27.02 & 12.91 & 26.45 & 0.05 \\
SkySense-O\methodtypes{\circ\star} & 25.52 & 19.10 & 13.02 & 34.55 & 0.35 \\
SegEarth-R1\methodtypes{\diamond\star} & 21.79 & 27.82 & 13.55 & 9.51 & 0.04 \\
ClearCLIP\methodtypes{\circ} & 16.54 & 76.32 & 13.69 & 16.78 & 0.12 \\
GroundingDINO+SAM\methodtypes{\triangle} & 17.62 & 41.32 & 13.90 & 11.38 & -0.03 \\
ProxyCLIP\methodtypes{\circ} & 15.89 & 84.88 & 14.25 & 17.26 & 0.11 \\
Pi-Seg-v1-L\methodtypes{\circ\star} & 28.29 & 36.58 & 15.47 & 25.44 & 0.30 \\
CAT-Seg-L\methodtypes{\circ} & 21.10 & 57.97 & 16.28 & 23.62 & 0.22 \\
RemoteSAM\methodtypes{\circ\star} & 23.17 & 33.26 & 16.45 & 14.46 & 0.07 \\
LAE-DINO+SAM\methodtypes{\triangle\star} & 24.40 & 26.23 & 17.41 & 15.44 & 0.08 \\
SegEarth-OV\methodtypes{\circ\star} & 22.23 & 68.02 & 17.50 & 22.48 & 0.07 \\
ALToLLM-8B\methodtypes{\diamond} & 20.85 & 60.77 & 19.43 & 23.40 & -- \\
SELF1E-2B\methodtypes{\diamond} & 24.94 & 51.34 & 19.53 & 24.51 & 0.02 \\
Pi-Seg-v2-B\methodtypes{\circ\star} & 24.81 & 65.39 & 20.27 & 24.69 & 0.11 \\
SAM3\methodtypes{\circ} & 23.59 & 31.18 & 21.15 & 38.39 & 0.35 \\
Sa2VA-Qwen2.5-VL-3B\methodtypes{\diamond} & 28.38 & 44.82 & 21.26 & 33.60 & 0.15 \\
Pi-Seg-v2-L\methodtypes{\circ\star} & 28.45 & 56.21 & 21.55 & 33.48 & 0.21 \\
GeoPixel-7B-RES\methodtypes{\diamond\star} & 39.87 & 33.71 & 23.06 & 23.52 & 0.02 \\
SAMTok-PLM-1B-co\methodtypes{\diamond} & 34.14 & 44.62 & 23.25 & 20.36 & 0.00 \\
Sa2VA-Qwen2.5-VL-7B\methodtypes{\diamond} & 32.60 & 41.67 & 23.39 & 44.71 & 0.12 \\
LocateAnything+SAM\methodtypes{\diamond\triangle} & 32.27 & 40.56 & 23.95 & 13.26 & 0.07 \\
SegEarth-R2\methodtypes{\diamond\star} & 41.78 & 34.07 & 25.51 & 36.11 & 0.13 \\
UniPixel-7B\methodtypes{\diamond} & 31.79 & 53.24 & 26.64 & 38.37 & -0.01 \\
SELF1E-8B\methodtypes{\diamond} & 34.57 & 55.03 & 27.05 & 39.93 & 0.01 \\
Sa2VA-Qwen3-VL-2B\methodtypes{\diamond} & 38.71 & 48.11 & 27.25 & 43.61 & 0.20 \\
Sa2VA-Qwen3-VL-4B\methodtypes{\diamond} & 40.78 & 49.40 & 28.34 & 45.68 & 0.22 \\
Sa2VA-Qwen3-VL-4B-SAM3\methodtypes{\diamond} & 43.39 & 59.80 & 30.45 & 44.80 & 0.12 \\
UniPixel-3B\methodtypes{\diamond} & 37.80 & 58.92 & 30.88 & 35.40 & 0.01 \\
SegEarth-OV3\methodtypes{\circ\star} & 36.34 & 63.92 & 31.68 & 24.36 & 0.22 \\
X2SAM\methodtypes{\diamond} & 41.76 & 53.19 & 31.82 & 42.89 & 0.01 \\
SAMTok-Qwen3-VL-4B-co\methodtypes{\diamond} & 51.77 & 52.53 & 33.31 & 34.74 & 0.01 \\
Rex-Omni+SAM\methodtypes{\diamond\triangle} & \textbf{53.55} & 51.62 & \textbf{38.75} & \textbf{46.01} & 0.14 \\
\bottomrule
\end{tabular}}
\caption{Quantitative zero-shot performance on the open-vocabulary segmentation task. Localization metrics are reported as percentages; MCC is unitless in $[-1,1]$. A dash indicates undefined MCC when a model predicts every query as present. Bold indicates the best.}
\label{tab:results}
\end{table}

\begin{table}[t]
\centering
\scalebox{0.79}{
\begin{tabular}{lcccc}
\toprule
Model & $ma$-P & $ma$-R & \textbf{$ma$-IoU} & $mi$-IoU \\
\midrule
Text4Seg-7B\methodtypes{\diamond} & 0.63 & 0.54 & 0.50 & 0.17 \\
Qwen3-VL-4B+SAM\methodtypes{\diamond\triangle} & 4.93 & 2.74 & 1.92 & 11.99 \\
SLIP-RS-T+SAM\methodtypes{\triangle\star} & 2.91 & 3.70 & 2.00 & 1.07 \\
SLIP-RS-L+SAM\methodtypes{\triangle\star} & 3.81 & 4.41 & 2.50 & 1.64 \\
YOLO-World2-L+SAM\methodtypes{\triangle} & 5.60 & 28.72 & 4.87 & 11.24 \\
OpenRSD+SAM\methodtypes{\triangle\star} & 19.11 & 12.80 & 9.30 & 9.82 \\
GroundingDINO+SAM\methodtypes{\triangle} & 13.28 & 34.02 & 11.76 & 16.81 \\
RemoteSAM\methodtypes{\circ\star} & 16.18 & 57.82 & 12.49 & 10.22 \\
LISA-7B\methodtypes{\diamond} & 18.71 & 26.57 & 13.62 & 38.59 \\
SegEarth-R1\methodtypes{\diamond\star} & 21.22 & 38.72 & 15.39 & 9.11 \\
UniGeoSeg\methodtypes{\diamond\star} & 27.39 & 25.82 & 16.29 & 13.46 \\
LAE-DINO+SAM\methodtypes{\triangle\star} & 23.84 & 36.93 & 18.38 & 15.03 \\
SegEarth-R2\methodtypes{\diamond\star} & 28.75 & 27.45 & 19.77 & 27.20 \\
SELF1E-2B\methodtypes{\diamond} & 26.88 & 49.68 & 20.80 & 35.84 \\
SAM3-Agent\methodtypes{\circ\diamond} & 29.94 & 30.81 & 21.88 & 9.61 \\
ALToLLM-8B\methodtypes{\diamond} & 25.93 & 54.48 & 22.84 & 41.61 \\
Sa2VA-Qwen2.5-VL-3B\methodtypes{\diamond} & 31.12 & 39.50 & 22.93 & 55.27 \\
Sa2VA-Qwen2.5-VL-7B\methodtypes{\diamond} & 32.75 & 38.33 & 23.09 & 52.76 \\
UniPixel-7B\methodtypes{\diamond} & 33.55 & 48.26 & 26.34 & 54.05 \\
GeoPixel-7B-RES\methodtypes{\diamond\star} & 40.02 & 39.35 & 26.39 & 25.37 \\
Sa2VA-Qwen3-VL-2B\methodtypes{\diamond} & 40.23 & 45.69 & 27.17 & 53.97 \\
SELF1E-8B\methodtypes{\diamond} & 37.85 & 59.13 & 29.69 & 50.05 \\
SAMTok-PLM-1B-co\methodtypes{\diamond} & 42.08 & 45.64 & 29.93 & 46.59 \\
LocateAnything+SAM\methodtypes{\diamond\triangle} & 43.24 & 43.75 & 32.05 & 13.42 \\
UniPixel-3B\methodtypes{\diamond} & 40.65 & 58.54 & 32.37 & 54.31 \\
Sa2VA-Qwen3-VL-4B\methodtypes{\diamond} & 46.35 & 54.51 & 32.63 & 59.06 \\
X2SAM\methodtypes{\diamond} & 43.09 & 52.28 & 32.86 & 52.32 \\
Sa2VA-Qwen3-VL-4B-SAM3\methodtypes{\diamond} & 44.75 & \textbf{68.01} & 35.04 & \textbf{61.08} \\
SAMTok-Qwen3-VL-4B-co\methodtypes{\diamond} & \textbf{58.60} & 57.73 & 40.21 & 60.81 \\
Rex-Omni+SAM\methodtypes{\diamond\triangle} & 58.15 & 54.23 & \textbf{41.32} & 49.63 \\
\bottomrule
\end{tabular}
}
\caption{Quantitative zero-shot performance on the referring segmentation task. All values are reported as percentages. Bold indicates the best.}
\label{tab:ref_result}
\vspace{-1em}
\end{table}

\begin{table}[t]
\centering
\scalebox{0.75}{
\begin{tabular}{lcccc}
\toprule
Model & $ma$-P & $ma$-R & \textbf{$ma$-IoU} & $mi$-IoU \\
\midrule
Text4Seg-7B\methodtypes{\diamond} & 0.59 & 0.60 & 0.47 & 0.20 \\
SLIP-RS-T+SAM\methodtypes{\triangle\star} & 1.79 & 1.47 & 0.94 & 0.41 \\
SLIP-RS-L+SAM\methodtypes{\triangle\star} & 2.34 & 2.01 & 1.50 & 0.65 \\
Qwen3-VL-4B+SAM\methodtypes{\diamond\triangle} & 6.06 & 3.56 & 2.61 & 12.76 \\
YOLO-World2-L+SAM\methodtypes{\triangle} & 6.19 & 28.47 & 5.10 & 9.62 \\
OpenRSD+SAM\methodtypes{\triangle\star} & 18.28 & 10.85 & 7.58 & 9.76 \\
GroundingDINO+SAM\methodtypes{\triangle} & 12.63 & 32.86 & 10.58 & 13.37 \\
LISA-7B\methodtypes{\diamond} & 19.46 & 33.99 & 14.87 & 28.75 \\
LAE-DINO+SAM\methodtypes{\triangle\star} & 21.94 & 28.32 & 15.47 & 14.27 \\
SegEarth-R1\methodtypes{\diamond\star} & 25.02 & 32.16 & 15.79 & 10.60 \\
UniGeoSeg\methodtypes{\diamond\star} & 33.40 & 24.18 & 16.85 & 18.16 \\
SELF1E-2B\methodtypes{\diamond} & 26.20 & 48.56 & 19.42 & 28.99 \\
SAM3-Agent\methodtypes{\circ\diamond} & 31.42 & 28.28 & 20.26 & 11.57 \\
Sa2VA-Qwen2.5-VL-3B\methodtypes{\diamond} & 31.12 & 35.30 & 20.64 & 46.38 \\
ALToLLM-8B\methodtypes{\diamond} & 24.70 & 50.95 & 20.73 & 30.84 \\
Sa2VA-Qwen2.5-VL-7B\methodtypes{\diamond} & 32.86 & 36.25 & 22.42 & 46.93 \\
SegEarth-R2\methodtypes{\diamond\star} & 34.77 & 30.35 & 22.52 & 30.36 \\
GeoPixel-7B-RES\methodtypes{\diamond\star} & 35.99 & 34.28 & 22.65 & 22.99 \\
SAMTok-PLM-1B-co\methodtypes{\diamond} & 40.21 & 44.23 & 26.56 & 34.55 \\
Sa2VA-Qwen3-VL-2B\methodtypes{\diamond} & 40.77 & 43.23 & 27.11 & 47.89 \\
UniPixel-7B\methodtypes{\diamond} & 34.10 & 50.30 & 27.25 & 46.89 \\
SELF1E-8B\methodtypes{\diamond} & 38.64 & 54.18 & 29.54 & \textbf{52.15} \\
Sa2VA-Qwen3-VL-4B\methodtypes{\diamond} & 43.54 & 46.48 & 29.93 & 52.08 \\
UniPixel-3B\methodtypes{\diamond} & 39.37 & 56.90 & 30.71 & 48.31 \\
LocateAnything+SAM\methodtypes{\diamond\triangle} & 40.18 & 43.38 & 30.95 & 17.31 \\
Sa2VA-Qwen3-VL-4B-SAM3\methodtypes{\diamond} & 47.35 & \textbf{59.68} & 34.11 & 52.14 \\
SAMTok-Qwen3-VL-4B-co\methodtypes{\diamond} & \textbf{55.58} & 50.07 & 35.07 & 45.18 \\
X2SAM\methodtypes{\diamond} & 46.40 & 55.09 & 36.08 & 47.19 \\
Rex-Omni+SAM\methodtypes{\diamond\triangle} & 53.16 & 50.90 & \textbf{37.96} & 51.21 \\
\bottomrule
\end{tabular}
}
\caption{Quantitative zero-shot performance on the reasoning segmentation task. All values are reported as percentages. Bold indicates the best.}
\label{tab:results_exp3}
\vspace{-1em}
\end{table}

\begin{figure}[t]
  \centering
  \includegraphics[width=\linewidth]{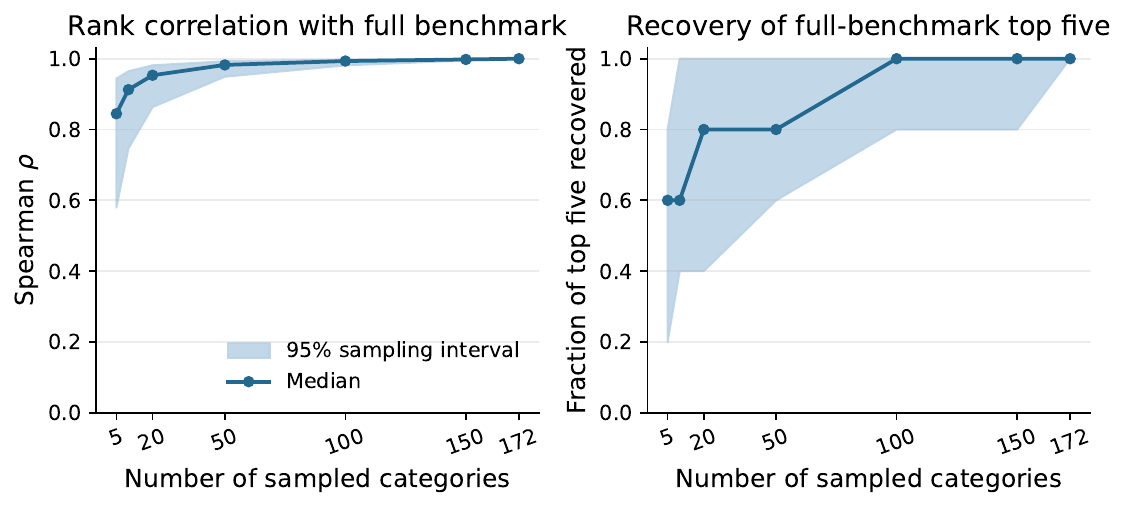}
  \caption{Sensitivity of vocabulary-segmentation model comparisons to category-set size and composition. Curves show medians over 1,000 subsets, and shaded regions show empirical 2.5th--97.5th percentiles. (a) Spearman's $\rho$ between the subset and full category rankings. (b) Fraction of the full-benchmark top five recovered.}
  \label{fig:category-scaling}
\end{figure}

\begin{table}[t]
\centering
\scalebox{0.78}{
\begin{tabular}{lcccc}
\toprule
Model & $mi$-P$_{0.5}$ & $mi$-R$_{0.5}$ & \textbf{$mi$-F1$_{0.5}$} & $mi$-F1$_{0.5:0.95}$ \\
\midrule
\multicolumn{5}{c}{\textbf{Vocabulary}} \\
\midrule
Qwen3-VL-4B\methodtypes{\diamond\triangle} & 0.37 & 0.15 & 0.22 & 0.06 \\
YOLO-World2-S\methodtypes{\triangle} & 21.90 & 0.76 & 1.47 & 0.67 \\
YOLO-World2-M\methodtypes{\triangle} & 27.03 & 1.46 & 2.78 & 1.27 \\
YOLO-World2-L\methodtypes{\triangle} & 34.76 & 1.69 & 3.23 & 1.53 \\
OpenRSD\methodtypes{\triangle\star} & \textbf{48.51} & 2.51 & 4.77 & 2.85 \\
LAE-DINO\methodtypes{\triangle\star} & 44.56 & 5.09 & 9.14 & 4.57 \\
GroundingDINO\methodtypes{\triangle} & 34.98 & 5.29 & 9.19 & 6.04 \\
RemoteSAM\methodtypes{\circ\star} & 16.04 & 9.68 & 12.08 & 7.07 \\
SLIP-RS-L\methodtypes{\triangle\star} & 9.83 & 18.03 & 12.72 & 6.33 \\
SLIP-RS-T\methodtypes{\triangle\star} & 15.98 & 18.33 & 17.07 & 8.68 \\
LocateAnything\methodtypes{\diamond\triangle} & 23.59 & 21.52 & 22.51 & \textbf{14.23} \\
Rex-Omni\methodtypes{\diamond\triangle} & 21.37 & \textbf{28.79} & \textbf{24.53} & 13.07 \\
\midrule
\multicolumn{5}{c}{\textbf{Referring}} \\
\midrule
Qwen3-VL-4B\methodtypes{\diamond\triangle} & 0.55 & 0.16 & 0.25 & 0.04 \\
RemoteSAM\methodtypes{\circ\star} & 9.12 & 1.95 & 3.21 & 2.02 \\
SLIP-RS-L\methodtypes{\triangle\star} & 19.42 & 2.37 & 4.22 & 2.51 \\
SLIP-RS-T\methodtypes{\triangle\star} & 29.63 & 2.30 & 4.27 & 2.60 \\
YOLO-World2-S\methodtypes{\triangle} & 6.26 & 3.87 & 4.78 & 2.39 \\
YOLO-World2-M\methodtypes{\triangle} & 7.22 & 5.31 & 6.12 & 3.32 \\
GroundingDINO\methodtypes{\triangle} & 17.31 & 3.87 & 6.32 & 3.44 \\
OpenRSD\methodtypes{\triangle\star} & 34.55 & 3.64 & 6.59 & 4.13 \\
YOLO-World2-L\methodtypes{\triangle} & 7.74 & 6.36 & 6.98 & 3.96 \\
LAE-DINO\methodtypes{\triangle\star} & 26.13 & 7.77 & 11.98 & 8.38 \\
Rex-Omni\methodtypes{\diamond\triangle} & 11.51 & \textbf{33.34} & 17.11 & 9.97 \\
LocateAnything\methodtypes{\diamond\triangle} & \textbf{53.50} & 26.63 & \textbf{35.56} & \textbf{21.47} \\
\midrule
\multicolumn{5}{c}{\textbf{Reasoning}} \\
\midrule
Qwen3-VL-4B\methodtypes{\diamond\triangle} & 0.51 & 0.17 & 0.25 & 0.07 \\
SLIP-RS-L\methodtypes{\triangle\star} & 38.46 & 0.40 & 0.78 & 0.55 \\
SLIP-RS-T\methodtypes{\triangle\star} & \textbf{38.93} & 0.58 & 1.14 & 0.82 \\
OpenRSD\methodtypes{\triangle\star} & 38.25 & 1.88 & 3.58 & 2.25 \\
GroundingDINO\methodtypes{\triangle} & 18.13 & 2.20 & 3.92 & 2.22 \\
YOLO-World2-S\methodtypes{\triangle} & 9.92 & 2.74 & 4.29 & 2.25 \\
YOLO-World2-M\methodtypes{\triangle} & 10.50 & 3.45 & 5.20 & 2.90 \\
LAE-DINO\methodtypes{\triangle\star} & 24.43 & 3.17 & 5.61 & 3.94 \\
YOLO-World2-L\methodtypes{\triangle} & 10.07 & 3.96 & 5.69 & 3.25 \\
Rex-Omni\methodtypes{\diamond\triangle} & 13.55 & \textbf{24.73} & 17.51 & 9.62 \\
LocateAnything\methodtypes{\diamond\triangle} & 31.34 & 22.39 & \textbf{26.12} & \textbf{14.91} \\
\bottomrule
\end{tabular}
}
\caption{Zero-shot object detection and visual grounding results across vocabulary, referring, and reasoning queries. All values are reported as percentages. Bold indicates the best result within each query block.}
\label{tab:results_exp4}
\end{table}

\subsection{Results and Analysis}

\paragraph{Open-vocabulary EO remains challenging.}
On OVEarth-Bench, current methods reach at most 38.75\%, 41.32\%, and 37.96\% segmentation $ma$-IoU on vocabulary, referring, and reasoning queries, respectively, all achieved by Rex-Omni+SAM. The best box-localization scores across the same query types are 24.53\%, 35.56\%, and 26.12\% for $mi$-F1$_{0.5}$ and 14.23\%, 21.47\%, and 14.91\% for $mi$-F1$_{0.5:0.95}$. These results indicate substantial room for improving open-vocabulary EO performance on OVEarth-Bench.

\paragraph{Limited gains from EO specialization.}
Only one EO-specific variant ranks among the top ten vocabulary-segmentation results, and none does so for referring or reasoning queries. The best EO-specific results remain 7.07--15.31 segmentation points and 7.46--23.58 detection points below the task maxima. Against related general baselines, SegEarth-OV exceeds ClearCLIP by 3.81 $ma$-IoU points and LAE-DINO+SAM exceeds GroundingDINO+SAM by 3.52--6.61 points, whereas GSNet and Pi-Seg-v1-L do not outperform CAT-Seg-L, although Pi-Seg-v2-L does. Thus, EO specialization can improve related baselines but does not consistently reach the best evaluated results.


\paragraph{MLLM-based methods perform best overall.}
MLLM-based methods account for nine of the ten best $ma$-IoU results on vocabulary queries and all ten best results on both referring and reasoning queries. This concentration is especially pronounced for the latter two tasks, where models must interpret compositional spatial relations or infer a target from its function rather than match an explicit category name. A plausible explanation is that MLLMs are often pretrained and instruction-tuned on larger and more heterogeneous mixtures of image--text, grounding, segmentation, and general multimodal data than task-specific EO systems. Such exposure may provide both broad visual--semantic alignment for long-tail category expressions and transferable language understanding for spatial and functional descriptions. Their advantage may therefore arise partly from the general representations learned by the MLLM, rather than solely from the downstream localization design.


\paragraph{Narrow category sets yield category-sensitive ranking.}
Do model comparisons remain consistent when evaluation is restricted to a narrow category set? We answer this question by repeatedly sampling smaller category sets from OVEarth-Bench and comparing the model rankings with the ranking obtained from all categories. As shown in Figure~\ref{fig:category-scaling}, with only five categories, the median Spearman's $\rho$ is 0.84 and the median top-five recovery is 0.6, meaning that three of the full benchmark's top five models are typically retained. At 20 categories, the corresponding values increase to 0.95 and 0.8. At 50 categories, the median rank correlation reaches 0.98. Thus, the overall model ordering becomes increasingly consistent with the full ranking as more categories are included. These results show that narrow category sets make model comparisons strongly dependent on which concepts are included, motivating broader coverage in open-vocabulary evaluation. Appendix~\ref{app:additional-analysis} reports more details.

\paragraph{Grounding and direct segmentation.}
Grounding-based pipelines map queries to boxes before applying SAM, whereas segmentation-based methods generate masks directly. Because the groups are unequal, we compare their two highest-scoring models. Rex-Omni+SAM exceeds the best direct segmentation model by 5.44, 1.11, and 1.88 $ma$-IoU points, while LocateAnything+SAM trails the corresponding second-best model by 7.87, 2.99, and 4.12 points. This crossed pattern establishes no general advantage for either paradigm. Training scale, backbone, and supervision format jointly impact the model. Further details are provided in Appendix~\ref{app:paradigm-comparison}.

\paragraph{Model scaling trends.}
Within the reported variants, SELF1E-8B exceeds SELF1E-2B by 7.52--10.12 $ma$-IoU points across the three query types. Sa2VA-Qwen3-VL-4B improves over its 2B counterpart by 1.08, 5.46, and 2.82 points on vocabulary, referring, and reasoning queries, respectively. Conversely, UniPixel-7B is 3.46--6.04 points below UniPixel-3B across all three tasks. Because these variants may differ beyond parameter count, the results show only that larger published variants do not consistently perform better on OVEarth-Bench.


\paragraph{Effects of query form.}
Does replacing an explicit category name with a spatial referring expression or an implicit functional description affect model performance? To isolate the effect of query form, we compare each model only on annotation targets shared by the query types. This comparison shows opposite effects: SegEarth-R2 declines on referring expressions, whereas UniGeoSeg improves. Reasoning descriptions likewise improve some models while reducing others, although the overall rankings remain similar across query types. Appendix~\ref{app:query-effects} reports the full comparison and cross-query rank correlations. Query form therefore reveals model-specific strengths and weaknesses rather than defining one query type as universally harder.

\paragraph{Presence discrimination remains weak}
MCC scores are generally low. Qwen3-VL-4B+SAM reaches 0.60 as a clear outlier, whereas every other model with a defined MCC scores at most 0.35. Moreover, across the 48 models with defined MCC, vocabulary $ma$-IoU and MCC have almost no rank association (Spearman $\rho=0.107$). Thus, accurately localizing present concepts does not imply reliable rejection of absent concepts (i.e., hallucination).

\section{Conclusion}

We presented OVEarth-Bench, a unified zero-shot benchmark for evaluating category and query diversity in open-vocabulary EO. Its broad taxonomy, diverse query types, and spatial annotations support a comprehensive evaluation of recognition, reasoning, and localization. Our experiments yield several key findings. MLLM-based methods achieve the strongest overall performance across query types. EO-specific models can outperform comparable general baselines, but still fall behind the strongest general methods. Meanwhile, the SOTA grounding-based and segmentation-based methods both perform competitively, demonstrating the potential of both approaches. These findings highlight the remaining challenges in open-vocabulary EO and establish OVEarth-Bench as a foundation for future model development. Further discussion and limitations of these findings are in Appendix~\ref{app:discussion-limitations}.

\bibliography{aaai2027}

\newpage

\appendix
\setcounter{secnumdepth}{1}

\section{Query Generation Prompt Templates}
\label{app:prompts}

The pipeline uses GPT-5-mini with structured JSON output.

\subsection{Stage 1: Generation}

\noindent\textit{System message:}
\begin{quote}
\small
You are an expert remote sensing image annotator. Analyze the red outlined target regions in satellite/aerial imagery. Generate useful text prompts for open-vocabulary segmentation and challenging absent negative candidates.
\end{quote}

\noindent\textit{User message} (\texttt{\{category\}} is substituted with the canonical category name; image is the visualization with the red-outline mask):
\begin{quote}
\small
The red outline marks the target mask boundary, expanded slightly outward so the original object edge remains visible.\\
Category Name: \texttt{\{category\}}\\[3pt]
Output strictly as a JSON object with these keys:\\
\texttt{"positive\_ovs\_phrases"}: exactly 2 short, common category noun phrases for the outlined target. At least one phrase must be exactly or directly include the original category name. Prefer standard object/category names over visual descriptions. Do not force extra attributes such as color, material, or shape unless they are part of a common category phrase.\\
\texttt{"positive\_res\_phrases"}: exactly 2 referring expressions locating the outlined target using spatial or geographic context. Each referring expression must directly include the category name.\\
\texttt{"positive\_reasoning\_phrases"}: exactly 2 prompts describing the target implicitly by function or purpose without naming the category.\\
\texttt{"negative\_candidates"}: 3 to 5 challenging negative noun phrases that are visually or conceptually similar but absent from the whole image.
\end{quote}

\subsection{Stage 2: Verification}

\noindent\textit{System message:}
\begin{quote}
\small
You are an expert quality control inspector for remote sensing datasets. Verify negative candidates and review referring expressions for factual correctness in satellite/aerial imagery.
\end{quote}

\noindent\textit{User message} (image is the raw image without mask overlay; \texttt{\{negatives\}} and \texttt{\{refs\}} are the JSON-serialised outputs from Stage~1):
\begin{quote}
\small
Carefully inspect the entire image. Remote sensing imagery can contain very small or densely packed objects.\\[3pt]
Category Name: \texttt{\{category\}}\\
Negative candidates to verify: \texttt{\{negatives\}}\\
Positive referring expressions to check: \texttt{\{refs\}}\\[3pt]
Tasks:\\
1.~Verify whether any negative candidates are actually present in the image.\\
2.~Check each referring expression for factual correctness, especially spatial relations and direction words (north/south/east/west, left/right, adjacent, near).\\
3.~If a referring expression contains an incorrect or overly specific location claim, rewrite it into a correct, safer expression that still names the category.\\
4.~If a referring expression is correct, keep it unchanged and mark it valid.\\[3pt]
Output strictly as a JSON object with these keys:\\
\texttt{"verification\_results"}: maps each negative candidate to \texttt{\{"status": "ABSENT"|"PRESENT", "reason": ...\}}.\\
\texttt{"positive\_res\_phrase\_qc"}: list with one entry per referring expression, each containing \texttt{"original\_phrase"}, \texttt{"status"} (\texttt{"valid"} or \texttt{"corrected"}), \texttt{"issue"}, and \texttt{"corrected\_phrase"}.\\
\texttt{"corrected\_positive\_res\_phrases"}: exactly 2 final referring expressions after correction.
\end{quote}

\section{Dataset Statistics}
\label{app:statistics}

\subsection{Image Statistics}

\begin{figure}[t]
  \centering
  \includegraphics[width=\linewidth]{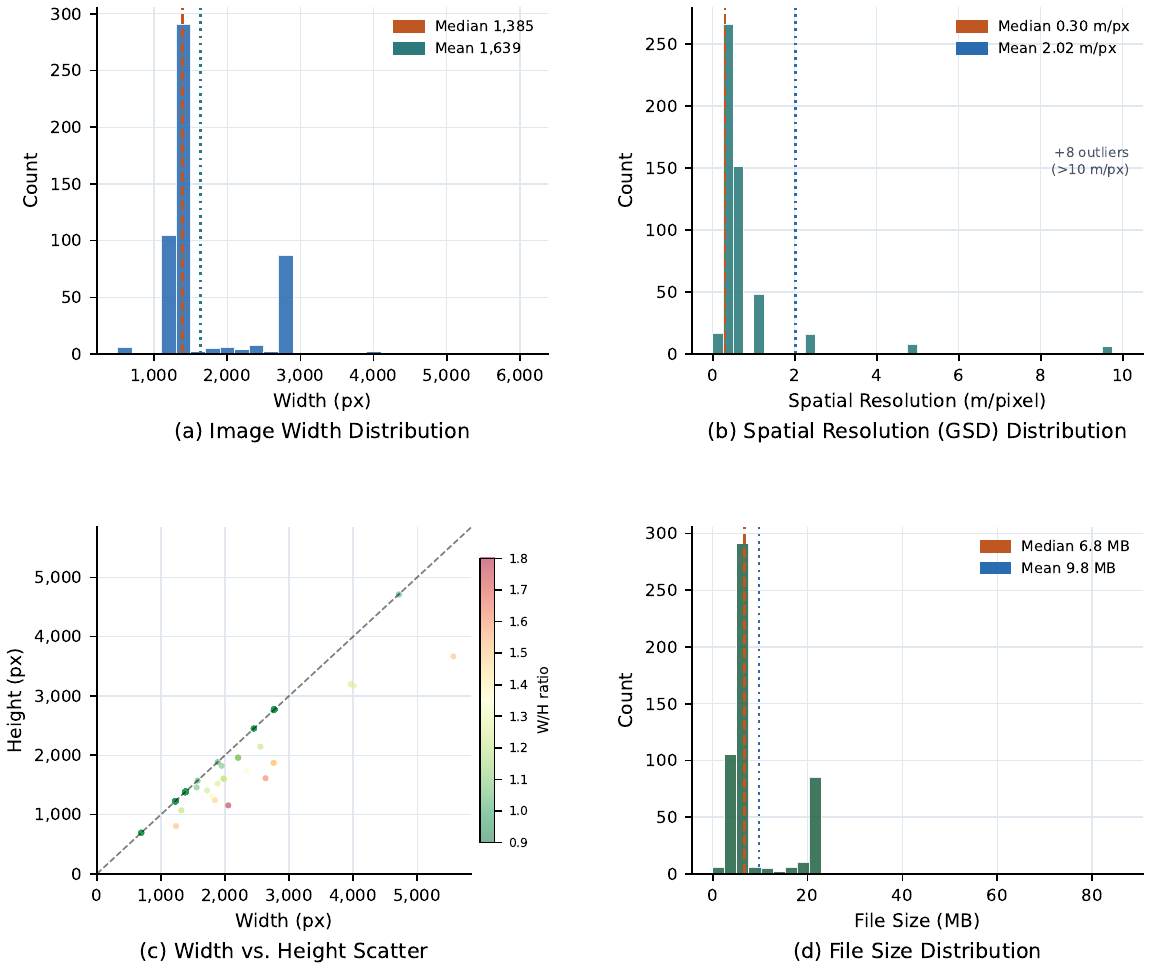}
  \caption{Detailed image statistics of OVEarth-Bench. The panels show the image-width distribution, spatial resolution (GSD) distribution, width--height relationship, and GeoTIFF file-size distribution.}
  \label{fig:imgstats-all}
\end{figure}

Figure~\ref{fig:imgstats-all} provides the complete image-level statistics. The collected images have diverse spatial extents and resolutions: image widths have a median of 1,385 pixels and a mean of 1,626 pixels, while GSD values have a median of 0.30 m/pixel and a mean of 2.02 m/pixel. This gap between the GSD median and mean is caused by a small number of coarse-resolution outliers, whereas most images are very-high-resolution Earth observation tiles. The width--height scatter further shows that many images are close to square crops.

\subsection{Annotation Statistics}

\begin{figure*}[t]
  \centering
  \includegraphics[width=\linewidth]{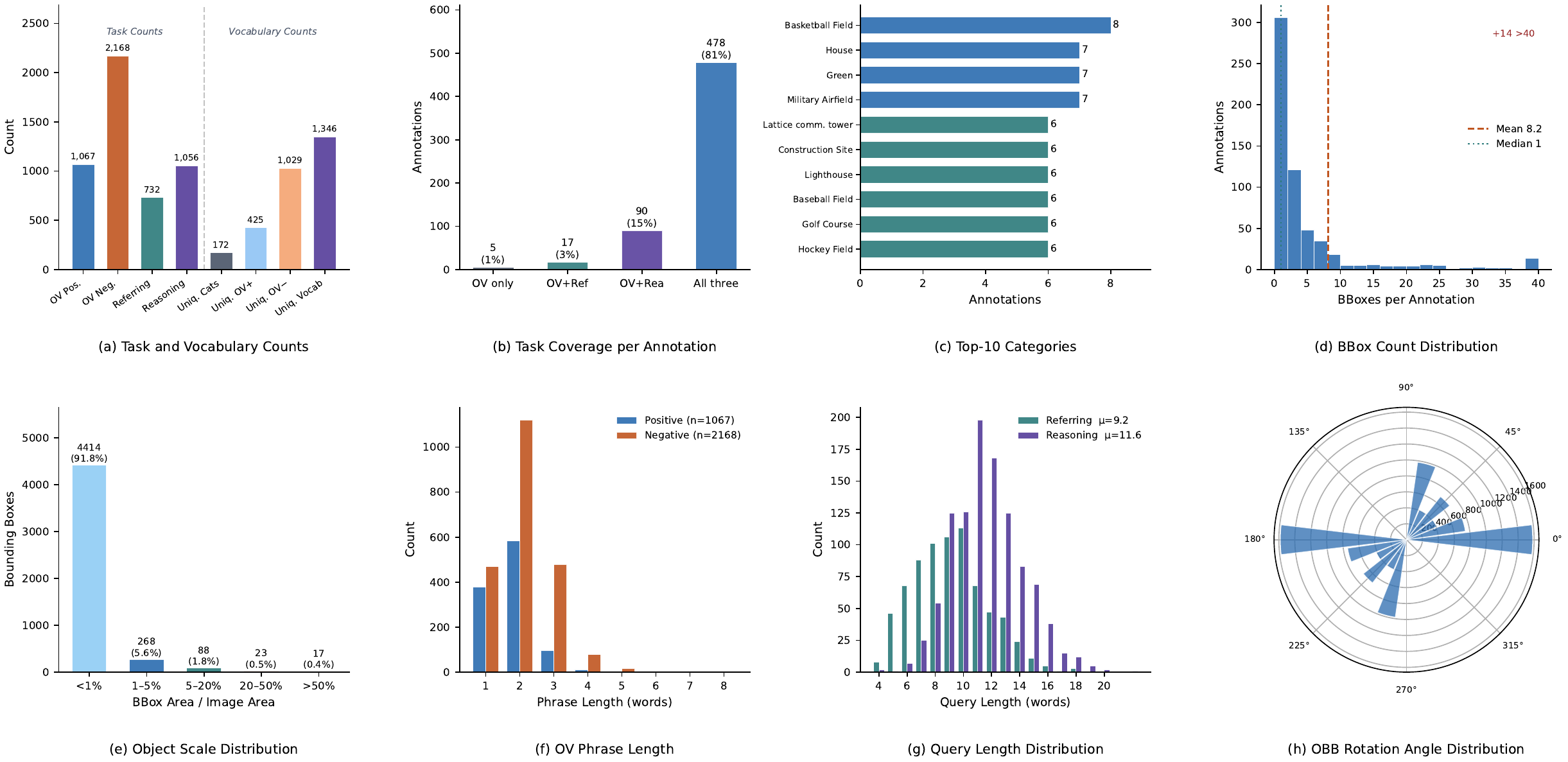}
  \caption{Detailed annotation statistics of OVEarth-Bench. The panels summarize task and vocabulary counts, task coverage per annotation, the most frequent categories, bounding-box counts, object-scale distribution, phrase and query lengths, and OBB rotation angles.}
  \label{fig:annstats-all}
\end{figure*}

Figure~\ref{fig:annstats-all} provides the complete annotation statistics. OVEarth-Bench contains 1,067 positive open-vocabulary phrases, 2,168 negative phrases, 732 referring expressions, and 1,056 reasoning queries. After deduplication within each polarity, these correspond to 425 unique positive and 1,029 unique negative phrases. Because 108 strings occur with both polarities in different samples, their union contains 1,346 unique vocabulary strings over 172 categories. Most annotations support the full set of language-query tasks: 478 annotations (81\%) contain open-vocabulary, referring, and reasoning queries; 90 annotations (15\%) contain open-vocabulary and reasoning queries; 17 annotations (3\%) contain open-vocabulary and referring queries; and only 5 annotations (1\%) contain open-vocabulary phrases alone.

For box-based evaluation, the detection subset contains 4,810 bounding boxes. The number of boxes per annotation is highly skewed, with a median of 1 and a mean of 8.2; 14 annotations contain more than 40 boxes. Object scales are also dominated by small targets: 4,414 boxes (91.8\%) occupy less than 1\% of the image area, while only 17 boxes (0.4\%) occupy more than half of the image. The language annotations remain concise, with most open-vocabulary phrases containing one to three words, and average lengths of 9.2 words for referring expressions and 11.6 words for reasoning queries. The OBB rotation-angle distribution further demonstrates the arbitrary orientations characteristic of RS objects.

\begin{figure}[t]
  \centering
  \includegraphics[width=\linewidth]{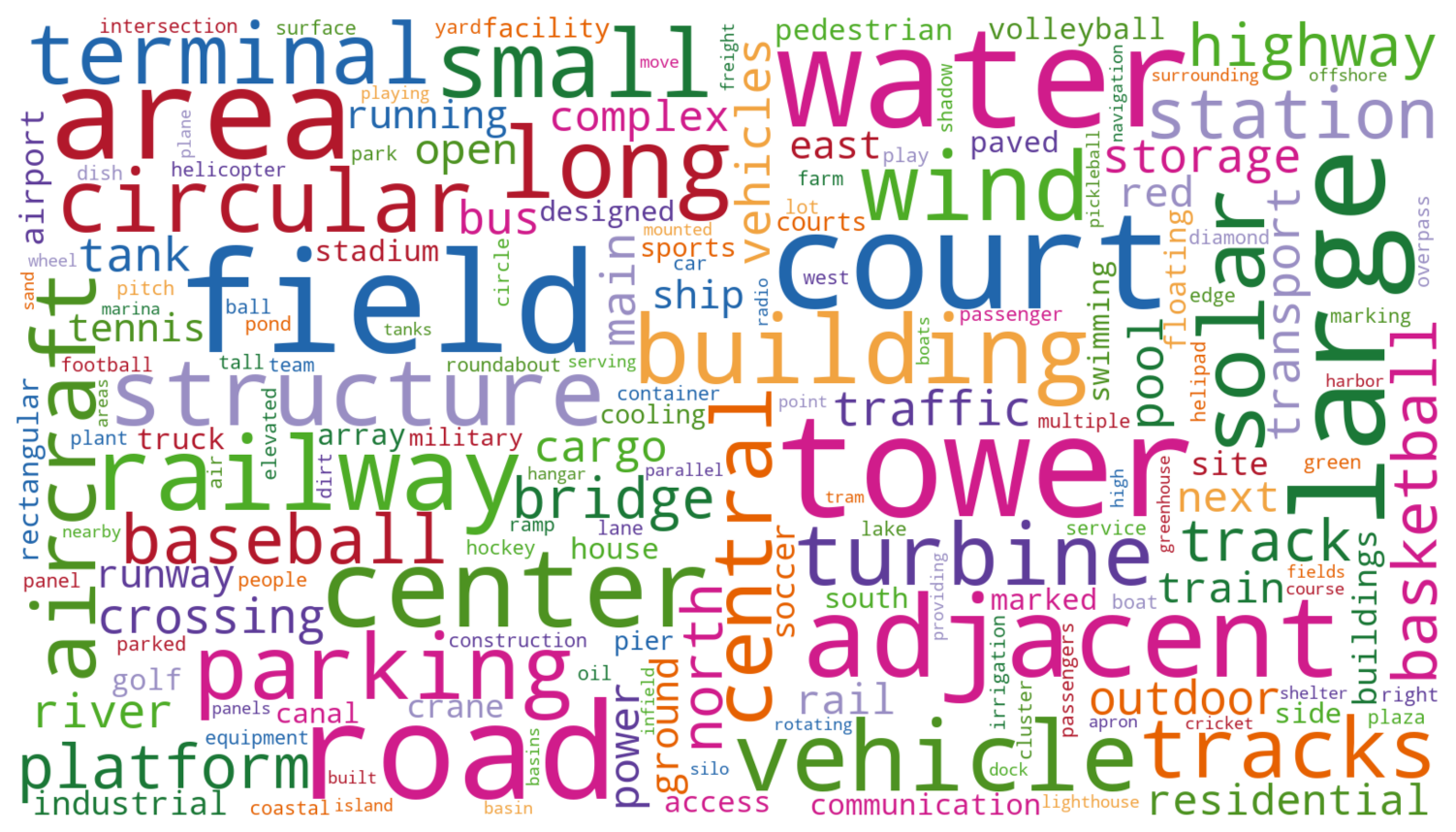}
  \caption{Word cloud of terms used in the language annotations of OVEarth-Bench. Larger words indicate more frequent terms.}
  \label{fig:wordcloud}
\end{figure}

\section{Evaluation Metrics}
\label{app:metrics}

We release an open-source Python evaluation package, \texttt{ovearth-eval}, to facilitate reproducible comparison on OVEarth-Bench. This section provides the formal definitions of all metrics reported in the main paper.

\begin{table*}[t]
\centering
\scriptsize
\setlength{\tabcolsep}{3pt}
\resizebox{0.9\textwidth}{!}{%
\begin{tabular}{lrrr@{\hspace{12pt}}lrrr}
\toprule
\multicolumn{4}{c}{Q-ranks 1--25} & \multicolumn{4}{c}{Q-ranks 26--49} \\
\cmidrule(lr){1-4}\cmidrule(lr){5-8}
Model & Q-mIoU & Ann.-mIoU & Cat.-mIoU & Model & Q-mIoU & Ann.-mIoU & Cat.-mIoU \\
\midrule
Rex-Omni+SAM & \textbf{38.75} & \textbf{38.98} & \textbf{39.32} & Pi-Seg-v1-L & 15.47 & 15.56 & 13.94 \\
SAMTok-Qwen3-VL-4B-co & 33.31 & 32.81 & 32.23 & ProxyCLIP & 14.25 & 14.04 & 14.12 \\
X2SAM & 31.82 & 31.50 & 31.74 & GroundingDINO+SAM & 13.90 & 13.84 & 14.55 \\
SegEarth-OV3 & 31.68 & 31.36 & 33.36 & ClearCLIP & 13.69 & 13.51 & 13.22 \\
UniPixel-3B & 30.88 & 30.46 & 29.79 & SegEarth-R1 & 13.55 & 13.15 & 12.48 \\
Sa2VA-Qwen3-VL-4B-SAM3 & 30.45 & 30.10 & 30.73 & SkySense-O & 13.02 & 12.98 & 10.69 \\
Sa2VA-Qwen3-VL-4B & 28.34 & 27.92 & 27.85 & LISA-7B & 12.91 & 12.29 & 11.27 \\
Sa2VA-Qwen3-VL-2B & 27.25 & 26.74 & 27.38 & CASS & 12.61 & 12.43 & 12.58 \\
SELF1E-8B & 27.05 & 26.74 & 24.77 & CAT-Seg-B & 10.90 & 10.63 & 10.33 \\
UniPixel-7B & 26.64 & 26.00 & 25.48 & Pi-Seg-v1-B & 10.57 & 10.42 & 9.62 \\
SegEarth-R2 & 25.51 & 24.56 & 23.75 & MaskCLIP & 10.43 & 10.17 & 9.54 \\
LocateAnything+SAM & 23.95 & 23.17 & 23.85 & GEM & 10.42 & 10.23 & 10.38 \\
Sa2VA-Qwen2.5-VL-7B & 23.39 & 22.99 & 23.11 & OpenRSD+SAM & 10.24 & 10.24 & 10.16 \\
SAMTok-PLM-1B-co & 23.25 & 22.57 & 23.29 & UniGeoSeg & 9.26 & 9.65 & 8.36 \\
GeoPixel-7B-RES & 23.06 & 22.45 & 21.12 & GSNet & 9.15 & 9.76 & 9.19 \\
Pi-Seg-v2-L & 21.55 & 21.20 & 18.77 & SLIP-RS-T+SAM & 9.03 & 9.00 & 8.80 \\
Sa2VA-Qwen2.5-VL-3B & 21.26 & 20.75 & 20.23 & SCLIP & 8.75 & 8.58 & 8.33 \\
SAM3 & 21.15 & 20.97 & 19.37 & SLIP-RS-L+SAM & 8.70 & 8.74 & 8.33 \\
Pi-Seg-v2-B & 20.27 & 20.06 & 18.75 & CLIP & 4.55 & 4.38 & 4.06 \\
SELF1E-2B & 19.53 & 19.16 & 18.55 & Qwen3-VL-4B+SAM & 2.27 & 2.24 & 2.33 \\
ALToLLM-8B & 19.43 & 19.02 & 17.94 & YOLO-World2-M+SAM & 1.73 & 1.79 & 2.27 \\
SegEarth-OV & 17.50 & 17.26 & 16.09 & YOLO-World2-L+SAM & 1.62 & 1.90 & 2.77 \\
LAE-DINO+SAM & 17.41 & 16.48 & 16.66 & YOLO-World2-S+SAM & 1.25 & 1.27 & 1.27 \\
RemoteSAM & 16.45 & 16.47 & 14.59 & Text4Seg-7B & 0.25 & 0.24 & 0.23 \\
CAT-Seg-L & 16.28 & 16.07 & 14.98 &  &  &  &  \\
\bottomrule
\end{tabular}}
\caption{Complete vocabulary-segmentation results under query-, annotation-, and category-level aggregation. Q-mIoU is the $ma$-IoU reported in the main table. Models follow the Q-mIoU ranking from top to bottom in the left panel and then the right panel. All values are percentages.}
\label{tab:aggregation-vocabulary}
\end{table*}

\begin{table*}[t]
\centering
\scriptsize
\setlength{\tabcolsep}{3pt}
\resizebox{0.9\textwidth}{!}{%
\begin{tabular}{lrrr@{\hspace{12pt}}lrrr}
\toprule
\multicolumn{4}{c}{Referring} & \multicolumn{4}{c}{Reasoning} \\
\cmidrule(lr){1-4}\cmidrule(lr){5-8}
Model & Q-mIoU & Ann.-mIoU & Cat.-mIoU & Model & Q-mIoU & Ann.-mIoU & Cat.-mIoU \\
\midrule
Rex-Omni+SAM & \textbf{41.32} & \textbf{40.58} & \textbf{41.20} & Rex-Omni+SAM & \textbf{37.96} & \textbf{37.98} & \textbf{37.19} \\
SAMTok-Qwen3-VL-4B-co & 40.21 & 38.08 & 36.47 & X2SAM & 36.08 & 35.79 & 34.31 \\
Sa2VA-Qwen3-VL-4B-SAM3 & 35.04 & 33.78 & 34.29 & SAMTok-Qwen3-VL-4B-co & 35.07 & 34.81 & 34.04 \\
X2SAM & 32.86 & 31.33 & 31.61 & Sa2VA-Qwen3-VL-4B-SAM3 & 34.11 & 33.97 & 33.78 \\
Sa2VA-Qwen3-VL-4B & 32.63 & 30.82 & 30.75 & LocateAnything+SAM & 30.95 & 30.38 & 29.06 \\
UniPixel-3B & 32.37 & 31.47 & 30.85 & UniPixel-3B & 30.71 & 30.38 & 27.83 \\
LocateAnything+SAM & 32.05 & 30.18 & 29.63 & Sa2VA-Qwen3-VL-4B & 29.93 & 29.88 & 29.45 \\
SAMTok-PLM-1B-co & 29.93 & 28.17 & 28.61 & SELF1E-8B & 29.54 & 29.05 & 26.42 \\
SELF1E-8B & 29.69 & 28.67 & 27.34 & UniPixel-7B & 27.25 & 26.95 & 25.04 \\
Sa2VA-Qwen3-VL-2B & 27.17 & 25.43 & 26.81 & Sa2VA-Qwen3-VL-2B & 27.11 & 27.07 & 26.94 \\
GeoPixel-7B-RES & 26.39 & 25.01 & 23.01 & SAMTok-PLM-1B-co & 26.56 & 26.12 & 26.07 \\
UniPixel-7B & 26.34 & 25.06 & 25.03 & GeoPixel-7B-RES & 22.65 & 22.24 & 20.04 \\
Sa2VA-Qwen2.5-VL-7B & 23.09 & 21.97 & 22.25 & SegEarth-R2 & 22.52 & 22.80 & 21.12 \\
Sa2VA-Qwen2.5-VL-3B & 22.93 & 21.76 & 21.64 & Sa2VA-Qwen2.5-VL-7B & 22.42 & 22.30 & 21.87 \\
ALToLLM-8B & 22.84 & 22.15 & 21.85 & ALToLLM-8B & 20.73 & 20.31 & 17.71 \\
SAM3-Agent & 21.88 & 20.58 & 19.91 & Sa2VA-Qwen2.5-VL-3B & 20.64 & 20.35 & 19.37 \\
SELF1E-2B & 20.80 & 20.02 & 20.05 & SAM3-Agent & 20.26 & 20.26 & 18.53 \\
SegEarth-R2 & 19.77 & 18.67 & 17.40 & SELF1E-2B & 19.42 & 19.20 & 17.69 \\
LAE-DINO+SAM & 18.38 & 17.45 & 16.70 & UniGeoSeg & 16.85 & 17.04 & 15.92 \\
UniGeoSeg & 16.29 & 15.97 & 13.36 & SegEarth-R1 & 15.79 & 15.51 & 14.81 \\
SegEarth-R1 & 15.39 & 14.52 & 13.75 & LAE-DINO+SAM & 15.47 & 15.41 & 15.38 \\
LISA-7B & 13.62 & 12.93 & 11.95 & LISA-7B & 14.87 & 14.40 & 13.27 \\
RemoteSAM & 12.49 & 12.31 & 11.18 & GroundingDINO+SAM & 10.58 & 10.79 & 11.14 \\
GroundingDINO+SAM & 11.76 & 10.98 & 11.24 & OpenRSD+SAM & 7.58 & 7.77 & 7.03 \\
OpenRSD+SAM & 9.30 & 9.24 & 8.14 & YOLO-World2-L+SAM & 5.10 & 4.91 & 5.31 \\
YOLO-World2-L+SAM & 4.87 & 4.76 & 5.14 & YOLO-World2-M+SAM & 4.85 & 4.68 & 5.14 \\
YOLO-World2-M+SAM & 4.57 & 4.37 & 3.94 & YOLO-World2-S+SAM & 3.53 & 3.42 & 4.04 \\
YOLO-World2-S+SAM & 4.14 & 3.47 & 3.43 & Qwen3-VL-4B+SAM & 2.61 & 2.53 & 2.49 \\
SLIP-RS-L+SAM & 2.50 & 2.59 & 1.99 & SLIP-RS-L+SAM & 1.50 & 1.65 & 1.76 \\
SLIP-RS-T+SAM & 2.00 & 1.86 & 1.56 & SLIP-RS-T+SAM & 0.94 & 1.05 & 1.20 \\
Qwen3-VL-4B+SAM & 1.92 & 1.79 & 2.12 & Text4Seg-7B & 0.47 & 0.44 & 0.30 \\
Text4Seg-7B & 0.50 & 0.37 & 0.27 &  &  &  &  \\
\bottomrule
\end{tabular}}
\caption{Complete referring- and reasoning-segmentation results under query-, annotation-, and category-level aggregation. Models are ordered by Q-mIoU within each task. All values are percentages.}
\label{tab:aggregation-referring-reasoning}
\end{table*}

\subsection{Segmentation Metrics}

Localization metrics are computed only on positive queries. For each positive sample $i$, let $G_i$ denote the non-empty set of ground-truth foreground pixels and $P_i$ the set of predicted foreground pixels. A missing, null, empty, or all-zero prediction is represented by an empty $P_i$ and is assigned precision, recall, and IoU of zero. Otherwise, per-query precision ($\text{P}_i$), recall ($\text{R}_i$), and IoU ($\text{IoU}_i$) are defined as:
\begin{equation}
\text{P}_i = \frac{|G_i \cap P_i|}{|P_i|}, \quad
\text{R}_i = \frac{|G_i \cap P_i|}{|G_i|},
\end{equation}
\begin{equation}
\text{IoU}_i = \frac{|G_i \cap P_i|}{|G_i \cup P_i|}
\end{equation}

The $ma$-P, $ma$-R, and $ma$-IoU are obtained by averaging these metrics across all $N$ positive query samples:
\begin{equation}
ma\text{-P} = \frac{1}{N}\sum_{i=1}^N \text{P}_i, \quad
ma\text{-R} = \frac{1}{N}\sum_{i=1}^N \text{R}_i
\end{equation}
\begin{equation}
ma\text{-IoU} = \frac{1}{N}\sum_{i=1}^N \text{IoU}_i
\end{equation}
Negative queries are excluded from mask-overlap aggregation and are evaluated separately through presence detection.

Additionally, we report $mi$-IoU, which pools intersection and union areas over all positive query samples:
\begin{equation}
mi\text{-IoU} = \frac{\sum_{i=1}^N |G_i \cap P_i|}{\sum_{i=1}^N |G_i \cup P_i|}
\end{equation}

\begin{table}[t]
\centering
\small
\setlength{\tabcolsep}{5pt}
\begin{tabular}{lccc}
\toprule
Task & Q--Ann. $\rho$ & Q--Cat. $\rho$ & Top-10 overlap \\
\midrule
Vocabulary & 0.999 & 0.996 & 10/10 \\
Referring & 0.997 & 0.997 & 10/10 \\
Reasoning & 0.998 & 0.994 & 9/10 \\
\bottomrule
\end{tabular}
\caption{Robustness of model rankings to alternative aggregation. Spearman's $\rho$ compares the original query-macro ranking with annotation- and category-macro rankings. Top-10 overlap compares query- and category-macro rankings.}
\label{tab:aggregation-rank-robustness}
\end{table}

\subsection{Robustness to Evaluation Aggregation}
\label{app:aggregation-robustness}

The main tables average IoU over positive queries, directly measuring model performance across different language expressions. Since multiple queries may refer to the same image--mask annotation and several categories contain only one or two annotations, query-level averaging may assign different effective weights to visual targets and categories. We therefore additionally compute annotation-macro IoU (Ann.-mIoU) and category-macro IoU (Cat.-mIoU). The former assigns equal weight to each annotated visual target, while the latter assigns equal weight to each represented category.

For task $t$, let $\mathcal{A}_t$ denote its annotations, $\mathcal{Q}_t(a)$ the positive queries associated with annotation $a$, and $\operatorname{IoU}(q)$ the IoU obtained for query $q$. Ann.-mIoU is defined as
\begin{equation}
\operatorname{Ann\text{-}mIoU}^{(t)}
=
\frac{1}{|\mathcal{A}_t|}
\sum_{a\in\mathcal{A}_t}
\frac{1}{|\mathcal{Q}_t(a)|}
\sum_{q\in\mathcal{Q}_t(a)}
\operatorname{IoU}(q).
\end{equation}
Let $\mathcal{C}_t$ be the categories represented in task $t$, and let $\mathcal{A}_t(c)$ be the annotations of category $c$. Cat.-mIoU is defined as
\begin{equation}
\begin{aligned}
\operatorname{Cat\text{-}mIoU}^{(t)}
&= \frac{1}{|\mathcal{C}_t|}
\sum_{c\in\mathcal{C}_t}
\frac{1}{|\mathcal{A}_t(c)|}
\sum_{a\in\mathcal{A}_t(c)} \\
&\quad \frac{1}{|\mathcal{Q}_t(a)|}
\sum_{q\in\mathcal{Q}_t(a)}
\operatorname{IoU}(q).
\end{aligned}
\end{equation}

Per-query IoUs are taken from the same predictions used in the main evaluation. Empty predictions retain an IoU of zero. Negative vocabulary queries are excluded from localization aggregation and remain evaluated by MCC.

Model rankings remain highly consistent under the three aggregation schemes. The highest-ranked model remains unchanged in all three tasks. Query- and category-macro rankings share all ten leading models for vocabulary and referring queries, and nine of ten for reasoning queries. Annotation-macro rankings are even more closely aligned with the original results. Although category-level aggregation changes several local rank positions, it does not alter the overall comparison among the evaluated models. The consistent results across aggregation schemes indicate that the main findings are not driven by repeated query variants or unequal category frequencies. Tables~\ref{tab:aggregation-vocabulary} and~\ref{tab:aggregation-referring-reasoning} report the complete model-level results.

\subsection{Detection Metrics}

For each positive query and IoU threshold $\tau$, we construct a bipartite graph whose vertices are ground-truth and predicted boxes and whose edges connect pairs with IoU at least $\tau$. We compute a maximum-cardinality one-to-one matching on this graph. Higher-IoU edges are visited first during augmenting-path search, but maximizing the number of valid matches takes precedence over maximizing the sum of matched IoUs. Each matched pair is a true positive (TP); an unmatched prediction is a false positive (FP); and an unmatched ground truth is a false negative (FN). Let $\text{TP}_\tau$, $\text{FP}_\tau$, and $\text{FN}_\tau$ denote the counts pooled across all positive query samples. Missing predictions therefore contribute false negatives, whereas predictions on negative queries are excluded from localization counts and are assessed by the presence metric.

The micro Precision ($mi$-P$_\tau$), micro Recall ($mi$-R$_\tau$), and micro F1 ($mi$-F1$_\tau$) at threshold $\tau$ are defined as:
\begin{equation}
mi\text{-P}_{\tau} = \frac{\text{TP}_\tau}{\text{TP}_\tau+\text{FP}_\tau}, \quad
mi\text{-R}_{\tau} = \frac{\text{TP}_\tau}{\text{TP}_\tau+\text{FN}_\tau}
\end{equation}
\begin{equation}
mi\text{-F1}_{\tau} = \frac{2 \cdot mi\text{-P}_{\tau} \cdot mi\text{-R}_{\tau}}{mi\text{-P}_{\tau}+mi\text{-R}_{\tau}}
\end{equation}

If no boxes are predicted at threshold $\tau$, then $\text{TP}_\tau+\text{FP}_\tau=0$ while positive ground-truth boxes remain; in this case, we define $mi$-P$_\tau$ and $mi$-F1$_\tau$ as zero rather than leaving them undefined.

We report these metrics at the standard threshold of $\tau = 0.5$ as $mi$-P$_{0.5}$, $mi$-R$_{0.5}$, and $mi$-F1$_{0.5}$.

\paragraph{Why micro averaging.} Unlike segmentation, each detection sample in OVEarth-Bench may contain multiple ground truth instances. Macro-averaging would give equal weight to samples regardless of how many instances they contain, causing low-instance samples to exert disproportionate influence. Micro-averaging pools TP, FP, and FN counts across all samples before computing precision, recall, and F1, so every ground truth instance and every prediction contributes equally to the final score.

\paragraph{Mean Micro F1 ($mi$-F1$_{0.5:0.95}$).} Following the SAM 3 protocol~\cite{carion2025sam3}, we report $mi$-F1$_{0.5:0.95}$, which averages the micro F1 score over ten IoU thresholds from $0.50$ to $0.95$ in steps of $0.05$:
\begin{equation}
mi\text{-F1}_{0.5:0.95} = \frac{1}{10} \sum_{k=0}^{9} mi\text{-F1}_{0.5 + 0.05k}
\end{equation}
The matching and pooled counts are recomputed independently at every threshold. This metric is not average precision and does not integrate over confidence thresholds.

\paragraph{OBBs.} For OBB submissions, the evaluator applies the same protocol after replacing axis-aligned box IoU with polygon IoU between submitted four-corner quadrilaterals. It provides the same $mi$-P$_{0.5}$, $mi$-R$_{0.5}$, $mi$-F1$_{0.5}$, and $mi$-F1$_{0.5:0.95}$ metrics.

\subsection{Presence Detection Metric (MCC)}

In the open-vocabulary task, each sample is either positive (the queried concept is present in the image) or negative (it is absent). A model's output is treated as a binary prediction: present if any enabled modality contains a non-empty output after submission filtering, namely at least one foreground mask pixel, HBB, or OBB; otherwise it is absent. The task contains 1,067 positive and 2,168 negative queries. Because the classes are imbalanced, accuracy can obscure asymmetric behavior. We therefore use MCC:
\begin{equation}
\text{MCC} = \frac{\text{TP}\cdot\text{TN} - \text{FP}\cdot\text{FN}}
{\sqrt{(\text{TP}+\text{FP})(\text{TP}+\text{FN})(\text{TN}+\text{FP})(\text{TN}+\text{FN})}}
\end{equation}
where TP and TN are correct positive and negative predictions, and FP and FN are hallucinations and misses, respectively. MCC $\in [-1, 1]$; a value of $0$ corresponds to no binary association and $1$ to perfect prediction. If any marginal term in the denominator is zero, for example when the ground truth contains one class only or the model predicts the same presence label for every query, MCC is undefined and the evaluator returns a null value rather than assigning zero.

\section{Additional Analyses}
\label{app:additional-analysis}

\subsection{Category-Subset Stability}

We use the full category vocabulary-segmentation ranking as the reference. For each subset size $K\in\{5,10,20,50,100,150\}$, we draw 1,000 category subsets uniformly without replacement. For each model and subset, the score is the mean IoU over all positive vocabulary queries whose categories are selected. All 49 model variants have predictions for all 172 categories, and using all categories exactly reproduces the main vocabulary-segmentation ranking.

For each subset, we compare its model ranking with the full ranking using two measures. Spearman's $\rho$ measures agreement across all 49 model variants. Top-five recovery is defined as $|T_K\cap T_{172}|/5$, where $T_K$ and $T_{172}$ are the top-five model sets from the subset and full benchmark, respectively. A recovery of 0.6 means that three of the full benchmark's top five models are retained. The shaded bands in Figure~\ref{fig:category-scaling} show the empirical 2.5th--97.5th percentiles across the sampled category subsets; they are not confidence intervals for a population parameter.

With five categories, the median rank correlation is 0.84 (0.58--0.94), and the median top-five recovery is 0.6 (0.2--0.8). At 20 categories, the corresponding values increase to 0.95 (0.85--0.98) and 0.8 (0.4--1.0). At 50 categories, the median rank correlation reaches 0.98 (0.95--0.99), but the top-five recovery remains 0.8 (0.6--1.0). At 100 categories, the corresponding values are 0.99 (0.98--1.00) and 1.0 (0.8--1.0); at 150 categories, they are 1.00 (0.99--1.00) and 1.0 (0.8--1.0). These results show that agreement in the broad model ranking does not guarantee agreement on the leading models. The pronounced ranking changes at small subset sizes, including a median top-five recovery of only 0.6 with five categories, show that evaluations based on narrow category sets can materially change conclusions about the leading models.

\subsection{Model and Variant Comparisons}

\begin{figure}[t]
  \centering
  \includegraphics[width=\linewidth]{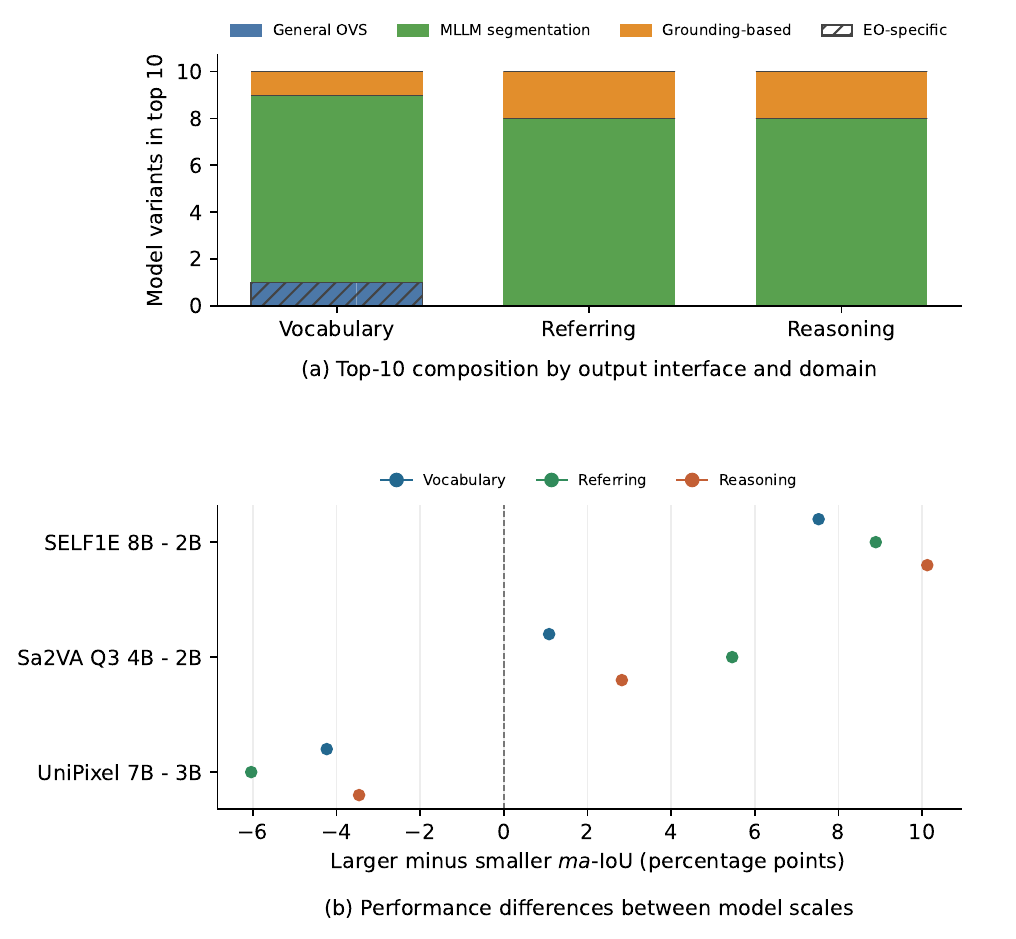}
  \caption{Model comparisons. (a) Composition of the ten best-performing segmentation models. Colors denote output interfaces, while hatching marks EO-specific methods; MLLM-based grounders are categorized as Grounding-based according to their evaluated mask-generation route. (b) Full-benchmark $ma$-IoU differences between larger and smaller variants of three model families across query types; positive values favor the larger variant.}
  \label{fig:model-comparisons}
\end{figure}

Figure~\ref{fig:model-comparisons} summarizes two complementary comparisons. In Figure~\ref{fig:model-comparisons}(a), colors distinguish three output interfaces. General OVS denotes generic dense open-vocabulary segmentation methods; MLLM segmentation denotes MLLM-based methods that directly emit or decode masks; and Grounding-based denotes a language-conditioned grounding model followed by an external SAM mask generator. Hatching independently marks methods designed or trained specifically for EO images, avoiding the use of domain specialization as a competing interface category. A method is considered MLLM-based whenever an MLLM jointly processes its image and language query, irrespective of its final output. Rex-Omni and LocateAnything are therefore MLLM-based methods, although they remain in the Grounding-based interface group in the figure. Related model sizes are counted as separate variants.

Figure~\ref{fig:model-comparisons}(a) shows that direct MLLM segmentation methods contribute eight of the ten best results for every query type. After also counting the MLLM-based grounding pipelines, MLLM-based methods account for nine of the ten best vocabulary results and all ten best referring and reasoning results. Only one EO-specific method enters the vocabulary top ten, and none enters the referring or reasoning top ten. Figure~\ref{fig:model-comparisons}(b) compares three model families using their full-benchmark $ma$-IoU differences, computed with the same $ma$-IoU definition as the main results. SELF1E-8B scores 7.52--10.12 points above SELF1E-2B across the three query types. Sa2VA-Qwen3-VL-4B exceeds its 2B counterpart by 1.08, 5.46, and 2.82 points on vocabulary, referring, and reasoning queries, respectively, whereas UniPixel-7B scores 3.46--6.04 points below UniPixel-3B. Larger variants therefore do not consistently transfer better to OVEarth-Bench. Because these published variants may differ in training data or recipe in addition to parameter count, this is a descriptive comparison rather than a controlled estimate of the causal effect of scale.

\begin{table}[t]
\centering
\scalebox{0.68}{
\begin{tabular}{lccccc}
\toprule
Contrast & \#Models & \#Ann. & $\rho$ & First higher & First lower \\
\midrule
Referring $-$ Vocabulary & 31 & 495 & 0.95 & 20 & 11 \\
Reasoning $-$ Vocabulary & 30 & 568 & 0.96 & 17 & 13 \\
Reasoning $-$ Referring & 31 & 478 & 0.98 & 12 & 19 \\
\bottomrule
\end{tabular}}
\caption{Query-form comparison on shared models and annotation targets. Each contrast is the first query type minus the second using the same $ma$-IoU definition as the main results. ``First higher'' and ``First lower'' count models with positive and negative direct $ma$-IoU differences, respectively. Spearman's $\rho$ is computed from the overall $ma$-IoU rankings of the common models.}
\label{tab:query-effects}
\end{table}

\subsection{Query-Form Effects}
\label{app:query-effects}

To determine whether query form imposes a consistent difficulty ordering or instead interacts with model design, we perform two complementary comparisons. First, for each query-type comparison, we rank only the models evaluated on both tasks and compute Spearman's $\rho$ to measure whether their broad ordering is preserved. Second, we restrict both tasks to annotation targets shared by the two query types and compute the segmentation metrics for each task on the shared target subset. Using the same spatial targets holds image and target content fixed across the two tasks. Table~\ref{tab:query-effects} reports the common model and annotation counts, rank correlations, and the numbers of models for which the first query type scores higher or lower than the second. Table~\ref{tab:query-effects-model} provides the corresponding model-level scores and differences.

\begin{table*}[t]
\centering
\scriptsize
\setlength{\tabcolsep}{3pt}
\renewcommand{\arraystretch}{0.95}
\begin{tabular}{lrrr@{\hspace{8pt}}rrr@{\hspace{8pt}}rrr}
\toprule
& \multicolumn{3}{c}{Vocabulary--Referring} & \multicolumn{3}{c}{Vocabulary--Reasoning} & \multicolumn{3}{c}{Referring--Reasoning} \\
\cmidrule(lr){2-4}\cmidrule(lr){5-7}\cmidrule(lr){8-10}
Model & Voc. & Ref. & $\Delta$ & Voc. & Rea. & $\Delta$ & Ref. & Rea. & $\Delta$ \\
\midrule
ALToLLM-8B                    & 18.15 & 22.84 & $+4.69$ & 19.73 & 20.73 & $+1.00$ & 23.30 & 20.46 & $-2.83$ \\
GeoPixel-7B-RES               & 25.01 & 26.39 & $+1.38$ & 23.27 & 22.65 & $-0.62$ & 26.87 & 24.25 & $-2.62$ \\
GroundingDINO+SAM             & 13.52 & 11.76 & $-1.76$ & 14.08 & 10.58 & $-3.50$ & 11.97 & 11.05 & $-0.92$ \\
LAE-DINO+SAM                  & 18.74 & 18.38 & $-0.36$ & 17.82 & 15.47 & $-2.35$ & 18.86 & 16.25 & $-2.61$ \\
LISA-7B                       & 13.19 & 13.62 & $+0.43$ & 13.18 & 14.87 & $+1.69$ & 14.08 & 15.20 & $+1.12$ \\
LocateAnything+SAM            & 23.93 & 32.05 & $+8.12$ & 24.18 & 30.95 & $+6.77$ & 32.69 & 30.51 & $-2.18$ \\
OpenRSD+SAM                   & 11.16 &  9.30 & $-1.85$ & 10.32 &  7.58 & $-2.74$ &  9.44 &  8.16 & $-1.28$ \\
Qwen3-VL-4B+SAM               &  2.12 &  1.92 & $-0.20$ &  2.31 &  2.61 & $+0.30$ &  2.00 &  2.50 & $+0.50$ \\
RemoteSAM                     & 15.06 & 12.49 & $-2.57$ & --    & --    & --      & --    & --    & --      \\
Rex-Omni+SAM                  & 38.41 & 41.32 & $+2.91$ & 38.71 & 37.96 & $-0.75$ & 41.29 & 37.96 & $-3.33$ \\
Sa2VA-Qwen2.5-VL-3B          & 22.15 & 22.93 & $+0.78$ & 21.46 & 20.64 & $-0.82$ & 23.36 & 22.01 & $-1.35$ \\
Sa2VA-Qwen2.5-VL-7B          & 24.24 & 23.09 & $-1.14$ & 23.64 & 22.42 & $-1.22$ & 23.47 & 23.26 & $-0.21$ \\
Sa2VA-Qwen3-VL-2B            & 29.32 & 27.17 & $-2.15$ & 27.55 & 27.11 & $-0.44$ & 27.49 & 29.38 & $+1.89$ \\
Sa2VA-Qwen3-VL-4B            & 30.46 & 32.63 & $+2.17$ & 28.52 & 29.93 & $+1.41$ & 32.93 & 32.53 & $-0.40$ \\
Sa2VA-Qwen3-VL-4B-SAM3       & 32.86 & 35.04 & $+2.18$ & 30.75 & 34.11 & $+3.36$ & 35.39 & 36.59 & $+1.20$ \\
SAM3-Agent                    & --    & --    & --      & --    & --    & --      & 22.37 & 19.85 & $-2.52$ \\
SAMTok-PLM-1B-co             & 24.24 & 29.93 & $+5.68$ & 23.44 & 26.56 & $+3.12$ & 30.17 & 28.13 & $-2.04$ \\
SAMTok-Qwen3-VL-4B-co        & 36.03 & 40.21 & $+4.18$ & 33.61 & 35.07 & $+1.45$ & 40.56 & 37.74 & $-2.82$ \\
SegEarth-R1                   & 13.88 & 15.39 & $+1.51$ & 13.68 & 15.79 & $+2.11$ & 15.84 & 15.51 & $-0.33$ \\
SegEarth-R2                   & 27.21 & 19.77 & $-7.43$ & 25.70 & 22.52 & $-3.17$ & 20.16 & 23.70 & $+3.54$ \\
SELF1E-2B                     & 19.86 & 20.80 & $+0.94$ & 19.81 & 19.42 & $-0.38$ & 21.22 & 19.90 & $-1.32$ \\
SELF1E-8B                     & 27.48 & 29.69 & $+2.22$ & 27.26 & 29.54 & $+2.28$ & 29.91 & 30.10 & $+0.19$ \\
SLIP-RS-L+SAM                 &  8.00 &  2.50 & $-5.51$ &  8.50 &  1.50 & $-7.00$ &  2.29 &  1.50 & $-0.79$ \\
SLIP-RS-T+SAM                 &  8.36 &  2.00 & $-6.36$ &  8.92 &  0.94 & $-7.98$ &  1.88 &  1.01 & $-0.88$ \\
Text4Seg-7B                   &  0.30 &  0.50 & $+0.20$ &  0.26 &  0.47 & $+0.21$ &  0.52 &  0.55 & $+0.03$ \\
UniGeoSeg                     &  8.50 & 16.29 & $+7.80$ &  9.09 & 16.85 & $+7.76$ & 16.47 & 17.83 & $+1.36$ \\
UniPixel-3B                   & 31.34 & 32.37 & $+1.03$ & 31.09 & 30.71 & $-0.38$ & 32.62 & 31.57 & $-1.06$ \\
UniPixel-7B                   & 26.79 & 26.34 & $-0.46$ & 26.86 & 27.25 & $+0.39$ & 26.46 & 27.71 & $+1.25$ \\
X2SAM                         & 32.40 & 32.86 & $+0.46$ & 32.16 & 36.08 & $+3.92$ & 33.18 & 36.71 & $+3.53$ \\
YOLO-World2-L+SAM             &  1.55 &  4.87 & $+3.32$ &  1.61 &  5.10 & $+3.49$ &  5.06 &  5.23 & $+0.17$ \\
YOLO-World2-M+SAM             &  1.67 &  4.57 & $+2.90$ &  1.73 &  4.85 & $+3.12$ &  4.73 &  4.90 & $+0.16$ \\
YOLO-World2-S+SAM             &  1.37 &  4.14 & $+2.77$ &  1.27 &  3.53 & $+2.26$ &  4.30 &  3.58 & $-0.72$ \\
\bottomrule
\end{tabular}
\caption{Model-level query-form comparison on shared annotation targets. The Vocabulary--Referring, Vocabulary--Reasoning, and Referring--Reasoning comparisons contain 495, 568, and 478 shared annotations, respectively. Scores are percentages, and each $\Delta$ is the second score minus the first. Dashes indicate that a model was not evaluated on both query types.}
\label{tab:query-effects-model}
\end{table*}

The high rank correlations show that broad performance tiers remain stable across query types. The direct differences nevertheless have both signs. Referring scores are higher than vocabulary scores for 20 of 31 models and lower for 11, while reasoning scores are higher than vocabulary scores for 17 of 30 models and lower for 13. Reasoning scores are higher than referring scores for 12 of 31 models and lower for 19. Thus, no query form is uniformly easier or harder across the evaluated models. The model-level changes can also be substantial. From vocabulary to referring queries, SegEarth-R2 decreases by 7.43 $ma$-IoU points, whereas UniGeoSeg and LocateAnything+SAM increase by 7.80 and 8.12 points, respectively. These are descriptive comparisons on the shared benchmark targets rather than population-level tests of model families.

\subsection{Comparison of Mask-Generation Paradigms}
\label{app:paradigm-comparison}

We compare two mask-generation paradigms on the common segmentation evaluation. The grounding-based group first predicts boxes or grounded regions and then uses a SAM model to obtain masks. The segmentation-based group directly produces dense masks through a segmentation head, mask decoder, dense vision-language matching, or another native mask interface. Because the paradigms contain unequal numbers of evaluated models and the grounding-based group includes several early or lightweight baselines, an aggregate mean or median would largely reflect which models were included. We therefore select the two highest-scoring models within each paradigm and task according to overall $ma$-IoU and report them individually.

\begin{table}[t]
\centering
\scalebox{0.58}{
\begin{tabular}{lllccc}
\toprule
Query & Paradigm & Model & $ma$-P & $ma$-R & $ma$-IoU \\
\midrule
Vocabulary & Grounding-based & Rex-Omni+SAM\methodtypes{\diamond\triangle} & 53.55 & 51.62 & 38.75 \\
Vocabulary & Grounding-based & LocateAnything+SAM\methodtypes{\diamond\triangle} & 32.27 & 40.56 & 23.95 \\
Vocabulary & Segmentation-based & SAMTok-Qwen3-VL-4B-co\methodtypes{\diamond} & 51.77 & 52.53 & 33.31 \\
Vocabulary & Segmentation-based & X2SAM\methodtypes{\diamond} & 41.76 & 53.19 & 31.82 \\
Referring & Grounding-based & Rex-Omni+SAM\methodtypes{\diamond\triangle} & 58.15 & 54.23 & 41.32 \\
Referring & Grounding-based & LocateAnything+SAM\methodtypes{\diamond\triangle} & 43.24 & 43.75 & 32.05 \\
Referring & Segmentation-based & SAMTok-Qwen3-VL-4B-co\methodtypes{\diamond} & 58.60 & 57.73 & 40.21 \\
Referring & Segmentation-based & Sa2VA-Qwen3-VL-4B-SAM3\methodtypes{\diamond} & 44.75 & 68.01 & 35.04 \\
Reasoning & Grounding-based & Rex-Omni+SAM\methodtypes{\diamond\triangle} & 53.16 & 50.90 & 37.96 \\
Reasoning & Grounding-based & LocateAnything+SAM\methodtypes{\diamond\triangle} & 40.18 & 43.38 & 30.95 \\
Reasoning & Segmentation-based & X2SAM\methodtypes{\diamond} & 46.40 & 55.09 & 36.08 \\
Reasoning & Segmentation-based & SAMTok-Qwen3-VL-4B-co\methodtypes{\diamond} & 55.58 & 50.07 & 35.07 \\
\bottomrule
\end{tabular}}
\caption{Comparison of the two best-performing models from each evaluated mask-generation paradigm. Values are percentages. Models are selected independently for each query type using overall $ma$-IoU.}
\label{tab:paradigm-comparison}
\end{table}

\begin{figure}[t]
  \centering
  \includegraphics[width=\linewidth]{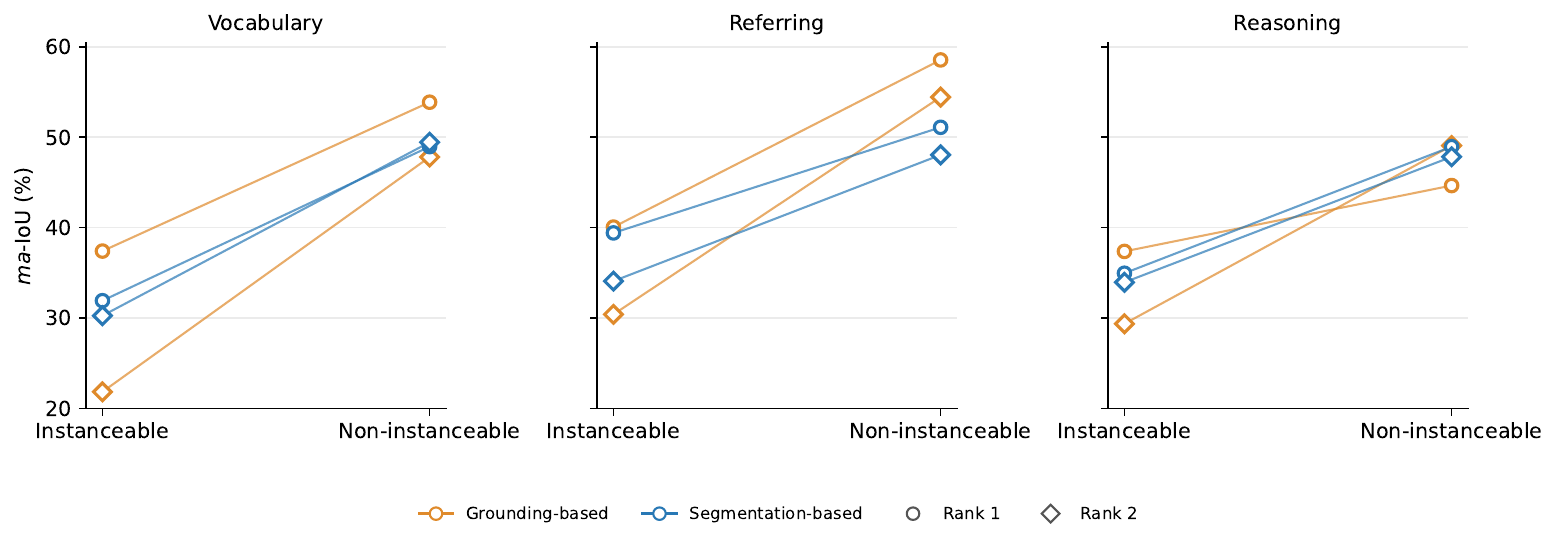}
  \caption{Performance of the two models selected within each mask-generation paradigm by overall task $ma$-IoU. Each point reports subset-specific $ma$-IoU; a line connects the same model across the two subsets only to aid reading and does not denote uncertainty. Non-instanceable targets are the 12 area-type or linear categories excluded from box evaluation.}
  \label{fig:paradigm-instanceability}
\end{figure}

\begin{figure}[t]
  \centering
  \includegraphics[width=\linewidth]{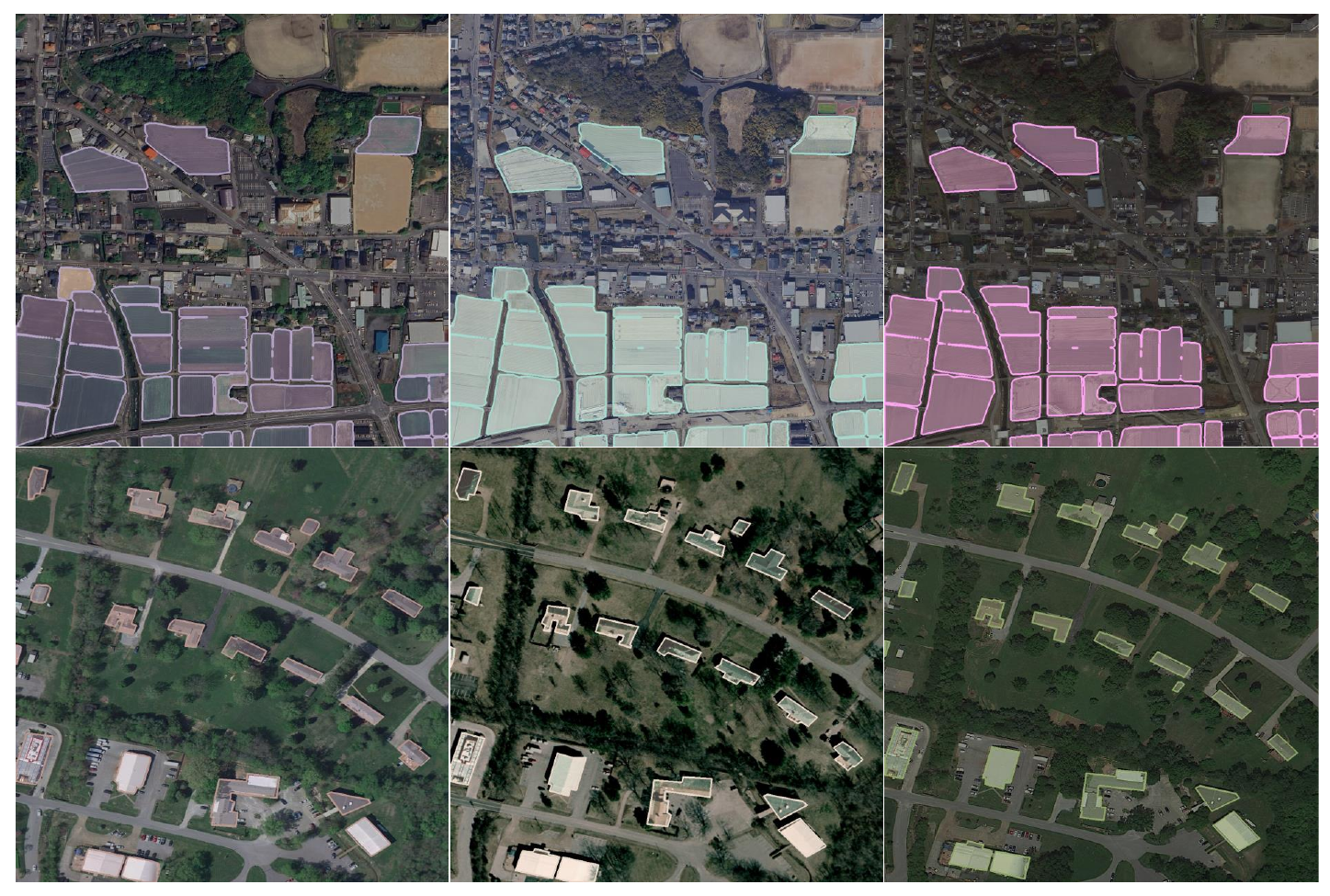}
  \caption{Same-location OVEarth-Bench examples acquired at different times and under different seasonal conditions.}
  \label{fig:sample2}
\end{figure}

Table~\ref{tab:paradigm-comparison} shows a consistent pattern between the best and second-best models across tasks but does not favor either paradigm uniformly. Rex-Omni+SAM, the best-performing grounding-based model, scores 38.75\%, 41.32\%, and 37.96\% $ma$-IoU, exceeding the best-performing segmentation-based models by 5.44, 1.11, and 1.88 points on vocabulary, referring, and reasoning queries. Conversely, the second-best grounding-based model, LocateAnything+SAM, scores 23.95\%, 32.05\%, and 30.95\%, trailing the corresponding second-best segmentation-based model by 7.87, 2.99, and 4.12 points. The highest observed score therefore comes from the grounding-based paradigm, but it is driven by a single model rather than a general advantage of that paradigm.

\paragraph{Effect of target instanceability.}
We next separate the 12 area-type or linear categories excluded from box evaluation (Appendix~\ref{app:excluded}) from the remaining categories. These categories contain 45 annotations in total; 45, 32, and 43 have vocabulary, referring, and reasoning queries, respectively. For each selected model and task, we report $ma$-IoU separately for the instanceable and non-instanceable subsets. Figure~\ref{fig:paradigm-instanceability} reports each of the four selected models individually; connecting lines only aid reading and do not denote uncertainty.

The four individual results do not reveal a consistent segmentation advantage on non-instanceable targets. Rex-Omni+SAM and LocateAnything+SAM obtain vocabulary $ma$-IoUs of 53.88\% and 47.81\%, while SAMTok-Qwen3-VL-4B-co and X2SAM obtain 48.96\% and 49.45\%. On referring queries, the two grounding-based models score 58.55\% and 54.44\%, both above the two selected segmentation-based models at 51.10\% and 48.05\%. On reasoning queries, the four scores overlap closely: Rex-Omni+SAM and LocateAnything+SAM obtain 44.66\% and 49.07\%, compared with 48.96\% for X2SAM and 47.83\% for SAMTok-Qwen3-VL-4B-co. A complementary pattern appears on instanceable targets: the best-performing grounding-based model exceeds the best-performing segmentation-based model by 5.49, 0.64, and 2.42 points across the three tasks, whereas the second-best grounding-based model trails its segmentation-based counterpart by 8.42, 3.68, and 4.59 points. Every selected model scores higher on the non-instanceable subset than on the remaining targets, suggesting that these area-type and linear categories are not necessarily harder in pixel-overlap terms. Although a box is geometrically imperfect for such targets, strong grounding-based models followed by SAM can recover competitive masks. Because this diagnostic covers only 12 non-instanceable categories and two benchmark-selected models per paradigm, it is descriptive rather than a paradigm-level comparison.


\paragraph{Possible explanations for the grounding ceiling.}
The best-performing grounding models also report substantially larger region-language training corpora than many mask-decoding MLLMs. LocateAnything is trained on 12M images with more than 138M language queries and 785M boxes~\cite{wang2026locateanything}, while Rex-Omni uses approximately 22M images during supervised fine-tuning~\cite{jiang2026rex}. By comparison, the reported instruction or joint-training scales are approximately 1.1M pairs for Sa2VA~\cite{yuan2025sa2va}, 1.57M samples for X2SAM~\cite{wang2026x2sam}, 561K samples for SELF1E (2.4M after magnification for its segmentation variant)~\cite{zhang2026rethinking}, 5M mask-text pairs for SAMTok~\cite{zhou2026samtok}, and 7.1M supervised samples for ALToLLM~\cite{wang2026alto}. Because the source papers count images, queries, masks, and augmented samples differently, these quantities are not directly comparable. They motivate, but do not test, the hypothesis that region--language training scale contributes to the highest observed grounding result.

A second hypothesis concerns output compatibility. Autoregressive MLLMs can express a bounding box with a short coordinate-token sequence, which may align more closely with their native token interface than dense-mask prediction. Producing a dense mask usually requires an additional mask decoder, mask tokenizer, or pixel-level feature alignment, each of which introduces a specialized output interface and additional training requirements. This may help large grounding models retain broader language generalization. The selected-model instanceability analysis shows that strong grounding-based models can also recover competitive masks for area-type and linear targets after SAM, so the expected geometric advantage of direct segmentation is not evident in the current subset. Neither explanation is isolated by the present benchmark: training scale, architecture, supervision, decoding, and prompting all vary jointly. Controlled studies that match the MLLM backbone and training data while changing only the output representation are required to test these hypotheses causally.

\subsection{Category Overlap with Other Benchmarks}
\label{app:category-overlap}

\begin{figure}[t]
  \centering
  \includegraphics[width=\linewidth]{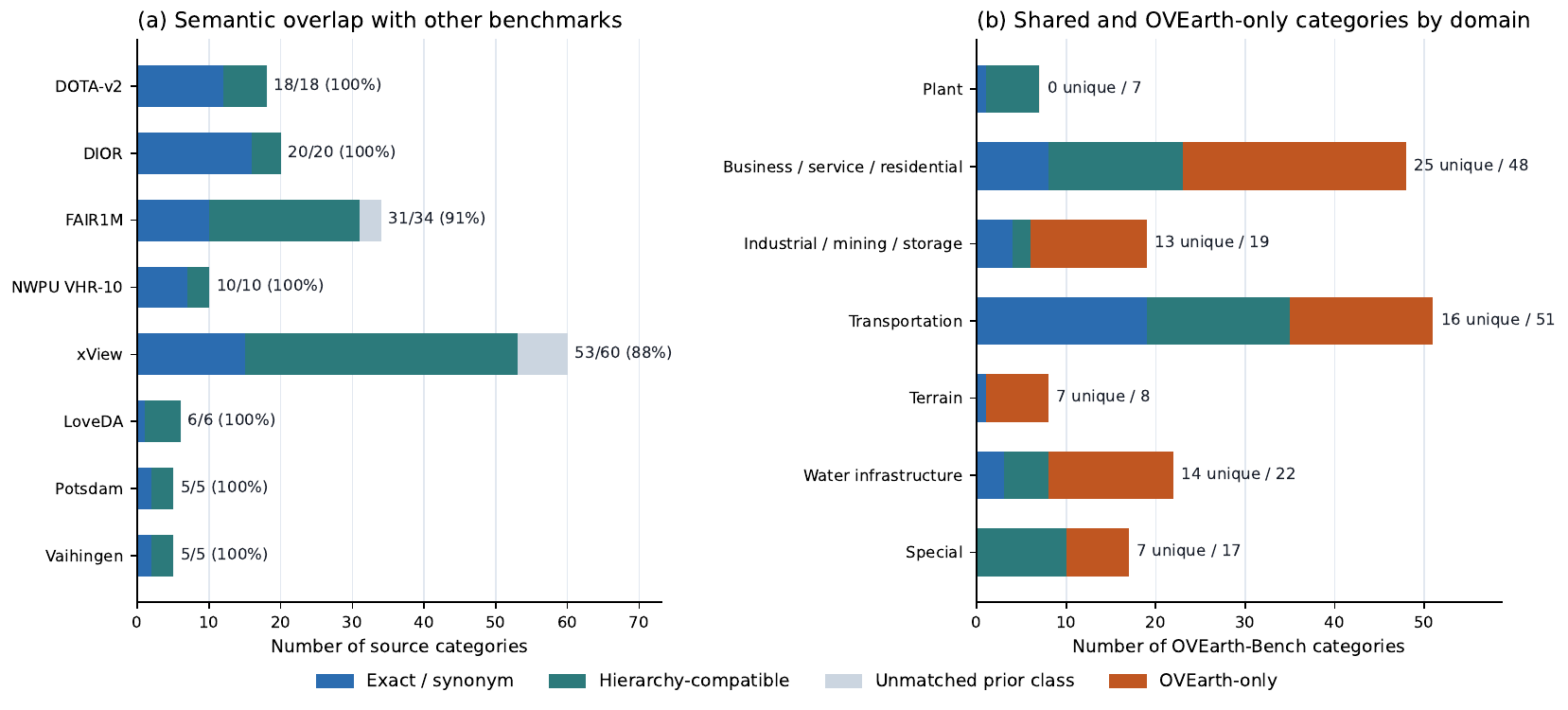}
  \caption{Bidirectional category overlap between OVEarth-Bench and the eight prior datasets marked as using fresh images in Table~\ref{tab:comparison}. (a) Source categories matched by exact/synonym or hierarchy-compatible concepts in OVEarth-Bench. Non-semantic catch-all and background labels are excluded from the denominators. (b) OVEarth-Bench categories shared with any selected benchmark or unique to OVEarth-Bench, grouped by domain.}
  \label{fig:fresh-category-overlap}
\end{figure}

We manually align the category inventories of DOTA-v2, DIOR, FAIR1M, NWPU VHR-10, xView, LoveDA, Potsdam, and Vaihingen with the OVEarth-Bench taxonomy. Each category is assigned as an exact/synonym match, a broader or narrower hierarchy-compatible match, or unmatched. Part--whole relations are generally not treated as matches. Non-specific catch-all and background labels, including \emph{other-vehicle} and \emph{clutter/background}, are excluded from the semantic-overlap denominators. 

As shown in Figure~\ref{fig:fresh-category-overlap}, all categories in six of the eight benchmarks have an OVEarth-Bench match; FAIR1M and xView reach 31/34 (91\%) and 53/60 (88\%), respectively. In the reverse direction, 36 of the 172 OVEarth-Bench categories have an exact/synonym match, 54 have only a hierarchy-compatible match, and 82 are not represented by the selected benchmarks. This bidirectional comparison indicates that OVEarth-Bench retains most established EO concepts while substantially extending the evaluated category space.

\section{Detection Subset: Excluded Categories}
\label{app:excluded}

The following 12 categories are excluded from the detection and OBB evaluation subsets because they correspond to area-type or linear features that are inherently non-instanceable or unsuitable for bounding-box evaluation: pipeline, common cropland, vineyard, arbor, grass, bareland, ice, asphalt road, mountain road, railway, port container, and river. Four additional annotations (heliostat $\times 2$, fish raft house $\times 1$, moat $\times 1$) were also excluded after post-hoc review determined that their instances were not reliably separable at the available image resolution. All remaining annotations contribute to the 4,810 bounding boxes used for detection scoring.

\section{Model Descriptions and Experimental Settings}
\label{app:models}

This section provides detailed descriptions of the evaluated models and the experimental settings used for each.

\subsection{General Open-Vocabulary Segmentation Models}

\paragraph{CLIP~\cite{radford2021clip}.} CLIP trains an image encoder and a text encoder jointly on 400 million image-text pairs collected from the internet using a contrastive objective, learning aligned visual and linguistic representations. The resulting model transfers zero-shot to downstream tasks by measuring cosine similarity between image patches and text embeddings without any additional training.

\paragraph{MaskCLIP~\cite{ding2022maskclip}.} MaskCLIP adapts CLIP for dense pixel-level prediction by extracting value features from the last attention layer of the frozen CLIP ViT encoder and matching them against text embeddings for direct semantic assignment. No training or fine-tuning is required.

\paragraph{CAT-Seg~\cite{cho2024catseg}.} CAT-Seg constructs a per-pixel cost volume by computing cosine similarity between CLIP patch embeddings and class text embeddings, then aggregates this cost volume through a spatial aggregation module based on per-class Swin Transformer blocks and a class aggregation module to produce dense semantic predictions. The model is fine-tuned on COCO-Stuff, comprising 118K images annotated with 171 semantic categories, with only the query and value projection matrices of CLIP updated during training.

\paragraph{ClearCLIP~\cite{lan2024clearclip}.} ClearCLIP is a training-free adaptation of CLIP for open-vocabulary semantic segmentation. It modifies only the final ViT block of a frozen CLIP model: the residual connection is removed, standard query-key self-attention is replaced with query-query self-attention, and the feed-forward sub-layer is discarded. No parameters are updated.

\paragraph{ProxyCLIP~\cite{lan2024proxyclip}.} ProxyCLIP is a training-free method that substitutes the self-attention maps of a frozen CLIP ViT with attention maps borrowed from a separately frozen vision foundation model such as DINO, DINOv2, SAM, or MAE. Adaptive normalization and activation masking are applied to the proxy attention before it is used for dense prediction. No parameters are updated.

\paragraph{CASS~\cite{kim2025distilling}.} CASS is a training-free approach that distills structural context from a vision foundation model (DINO ViT-B/8) into frozen CLIP features. It aligns structural information between the two models through spectral graph matching using Wasserstein distance on attention-head eigenvalues, injects a dynamically scaled low-rank approximation into CLIP attention during inference, and refines text embeddings using CLIP zero-shot classification scores. No parameters are updated.

\paragraph{GEM~\cite{bousselham2024grounding}.} GEM is a training-free method that improves CLIP spatial sensitivity by generalizing self-self attention to three variants: query-query, key-key, and value-value. Their outputs are ensembled with L2-normalized projections. An adaptive temperature $\tau = N\sqrt{d}/\sum\|x_i\|_2$ is computed from token statistics and applied in parallel at multiple ViT layers. No parameters are updated.

\paragraph{SCLIP~\cite{wang2024sclip}.} SCLIP is a training-free method that replaces standard query-key self-attention in the final ViT block of a frozen CLIP model with Correlative Self-Attention (CSA), which computes separate query-query and key-key similarity matrices to improve spatial correspondence for dense prediction. No parameters are updated.

\paragraph{SAM~3~\cite{carion2025sam3}.} SAM~3 is a promptable segmentation foundation model supporting concept segmentation. The model uses a Hiera image encoder with an image-level detector, a tracker decoder, and vision-text encoders for text-prompted segmentation. Training proceeds in four stages using large-scale data totaling over 2.5 billion training samples: SA-1B (one billion masks), SA-V (video frames with 467K masklets), and multiple SA-Co datasets including SA-Co/HQ (52.3M masks), SA-Co/EXT (70.5M masks), SA-Co/SYN (1.4 billion synthetic masks), and SA-Co/VIDEO (467.1K masklets) containing real human annotations, external datasets, and synthetic pseudo-labels generated by the SAM3 data engine. We evaluate the model for open-vocabulary segmentation and an agentic variant (SAM3-Agent) that employs an agent-based pipeline for referring and reasoning segmentation tasks.

\paragraph{SAM3-Agent inference.} For referring and reasoning queries, we use Qwen2.5-VL-7B-Instruct as an agent that translates image-conditioned complex language into short noun phrases suitable for SAM~3. At each round, the agent receives the original image and query, proposes a previously unused noun phrase, such as ``tower'' or ``greenhouse'', and invokes SAM~3's \texttt{segment\_phrase} tool. If no mask is returned or the candidates do not satisfy the original query, the agent proposes another non-repeated phrase. When several candidates are returned, Qwen inspects the original image, query, and an indexed visualization of the masks, then selects one or more masks or rejects the candidates after individual inspection. The masks returned are unioned into one binary prediction for scoring. The interaction is capped at 100 rounds; failure to return a final result within this budget is recorded as an empty prediction.

\subsection{Multimodal Large Language Models}

\paragraph{LISA~\cite{lai2024lisa}.} LISA extends LLaVA-1.5 with a special \texttt{<SEG>} token appended to the LLM vocabulary. When the model generates this token, the corresponding embedding is projected through a multi-layer perceptron and passed to a SAM mask decoder to produce the output mask. The LLM backbone is end-to-end fine-tuned, while LoRA adapters are applied to SAM to preserve its generalization ability. Training combines semantic segmentation datasets (ADE20K, COCO-Stuff), referring expression segmentation and VQA data, and ReasonSeg, totaling approximately 600K+ segmentation-related and 600K+ VQA training samples. We evaluate the 7B variant.

\paragraph{Sa2VA~\cite{yuan2025sa2va}.} Sa2VA integrates SAM~2 with an MLLM by routing all inputs through a shared LLM token space. The model generates instruction tokens that guide the frozen SAM-2 decoder while maintaining temporal consistency through the SAM~2 memory mechanism. Training combines public image and video segmentation datasets with Ref-SAV, totaling approximately 1.1 million image-text and video-text training pairs. We evaluate variants using Qwen2.5-VL-3B, Qwen2.5-VL-7B, Qwen3-VL-2B, and Qwen3-VL-4B backbone, with an additional Qwen3-VL-4B-SAM3 variant.

\paragraph{UniPixel~\cite{liu2026unipixel}.} UniPixel is built on a Qwen2.5-VL backbone and introduces an object memory bank that stores object-level spatial and temporal features to maintain consistency across video frames. The model is trained through a three-stage process using a combination of public image and video segmentation datasets: Stage 1 uses 851K samples to pre-train the sparse prompt encoder; Stage 2 employs 87K samples to align the LLM and mask decoder via the $L \rightarrow M$ projector; Stage 3 leverages $\sim$2M samples from diverse pixel-level and holistic-level tasks to jointly train the model. We evaluate 3B and 7B variants.

\paragraph{SAMTok~\cite{zhou2026samtok}.} SAMTok tokenizes segmentation masks into discrete tokens using a residual vector-quantized tokenizer (initialized from SAM-2) trained on 209 million mask instances from diverse open-source datasets. Each mask is compactly encoded as a fixed sequence of two codebook entries, which are added to the MLLM vocabulary as new special tokens. The full model is fine-tuned on $\sim$5M mask-text instruction pairs, followed by reinforcement learning with group relative policy optimization (GRPO) using a textual answer-matching reward. Following their framework, we evaluate the PLM-1B-co and Qwen3-VL-4B-co variants.

\paragraph{X2SAM~\cite{wang2026x2sam}.} X2SAM couples Qwen3-VL-4B with SAM2 in a dual-branch architecture, employing a global vision encoder branch for high-level understanding and a fine-grained mask encoder branch for pixel-level prediction. A mask memory module caches guided vision features from prior frames to support temporally consistent video segmentation. The model is trained through a two-phase process: (1) class-agnostic segmentor training on the mask-only SA-1B dataset (comprising 11M images and 1.1B masks) to initialize the mask decoder; and (2) unified joint training over a heterogeneous mixture of image and video datasets totaling $\sim$1.57M samples.

\paragraph{SELF1E~\cite{zhang2026rethinking}.} SELF1E adapts a multimodal large language model for segmentation by introducing a single \texttt{[SEG]} token whose embedding directly produces the segmentation mask via a matrix product with image features, eliminating the need for an auxiliary decoder. Residual Features Refilling (RFR) and Residual Features Amplifier (RFA) modules preserve fine-grained spatial details, and task-specific attention masks improve bidirectional feature interaction. The model is trained on a combination of semantic segmentation datasets, referring expression segmentation datasets, ReasonSeg, and VQA datasets, totaling $\sim$561k samples ($\sim$2.4M magnified samples for the SELF1E-SEG variant). We evaluate 2B and 8B ``-SEG'' variants.

\paragraph{ALToLLM~\cite{wang2026alto}.} ALToLLM integrates an adaptive-length tokenizer (ALTo) into an InternVL-2.5-8B backbone, autonomously encoding masks into variable-length sequences of 1 to 32 tokens based on object complexity via an attention-driven length predictor. For high-fidelity reconstruction, its mask de-tokenizer employs a SAM-ViT-L pixel encoder. The model is trained in four progressive stages with explicit data scales: (1) pretraining the tokenizer and (2) fine-tuning the length predictor on the Multi-Target-SA1B dataset (derived from SA-1B's 11M images and 1.1B masks); (3) supervised fine-tuning on a massive 7.1 million-sample multimodal mixture (including HiMTok and SA-1B datasets); and (4) reinforcement learning via GRPO using Multi-Target-SA1B, gRefCOCO, and the RefCOCO series (totaling over 630k referring expressions) to dynamically balance mask fidelity and token efficiency.

\paragraph{Text4Seg~\cite{lan2025text4seg}.} Text4Seg reformulates image segmentation as text generation by mapping mask patches to text labels, utilizing Row-wise Run-Length Encoding (R-RLE) to compress the resulting descriptor sequence by $\sim$74\% (at 16$\times$16 resolution). It seamlessly fine-tunes MLLM backbones (e.g., LLaVA-1.5, DeepSeek-VL, InternVL2-8B, Qwen-VL) using LoRA (rank 64). To support diverse segmentation tasks, the model is trained on visual instruction datasets for referring expression segmentation ($\sim$800k samples combined from refCLEF/COCO/+/g), generalized referring segmentation ($\sim$419k from grefCOCO), and open-vocabulary segmentation ($\sim$1.16M from COCOStuff).

\subsection{Detection and Grounding Models}

\paragraph{GroundingDINO~\cite{liu2024grounding}.} GroundingDINO combines a Swin Transformer image backbone with a BERT-base text encoder and performs tight cross-modal fusion through feature enhancement, language-guided query selection, and a cross-modality decoder. GroundingDINO-T is pretrained on a combination of detection, grounding, and caption datasets. This includes Objects365 v1 (609K images with 10M bounding box annotations), GoldG (770K image-text pairs), and Cap4M ($\sim$4M image-caption pairs), totaling around 5M unique image samples. This pretrained model enables zero-shot open-vocabulary detection from arbitrary text prompts.

\paragraph{YOLO-World2~\cite{cheng2024yoloworld}.} YOLO-World2 integrates a YOLOv8 detector with a frozen CLIP-base text encoder through a Re-parameterizable Vision-Language Path Aggregation Network (RepVL-PAN) and region-text contrastive pre-training. The model is pre-trained on a combination of detection datasets (Objects365, with 609K images), grounding datasets (GoldG, with 770K images), and an image-text dataset annotated via a pseudo-labeling pipeline (CC3M, with 246K images and 821K pseudo annotations). This pre-training protocol utilizes a total of $\sim$1.62M images containing around 14.7M region-text annotations, enabling zero-shot detection from text prompts without vocabulary constraints at inference. We evaluate the second-version S, M, and L variants.

\paragraph{LocateAnything~\cite{wang2026locateanything}.} LocateAnything is a 3B generative grounding model based on Parallel Box Decoding (PBD), which decodes all bounding-box coordinates and points as atomic units in a single parallel step or forward pass. The architecture pairs a Moon-ViT image encoder with a Qwen2.5 language model backbone. The model is trained on LocateAnything-Data, a composite dataset comprising over 138M natural language queries and 785M annotated bounding boxes across 12M unique images. This dataset integrates over 40 source datasets, including OpenImages, COCO, V3Det, Objects365, LVIS, SA-1B, RefCOCO/+/g, and gRefCOCO, covering both category-name detection and spatial referring expression grounding.

\paragraph{Rex-Omni~\cite{jiang2026rex}.} Rex-Omni is an MLLM with 3B parameters designed for versatile visual perception tasks, including object detection, referring, pointing, OCR, and GUI grounding. Under a unified coordinate prediction framework, normalized spatial coordinates are quantized into 1,000 discrete special tokens mapping to integers in [0, 999]. Depending on the specific task, spatial locations are generated as structured token sequences, such as four coordinate tokens for a bounding box, two tokens for a single point, or multiple tokens for polygons. The model is trained through a two-stage pipeline consisting of SFT on approximately 22M images (combining 8.9M public samples and 13.1M samples generated by custom data engines), followed by GRPO-based reinforcement post-training on a subset of 66K samples to refine spatial precision and correct behavioral deficiencies.

\subsection{RS Open-Vocabulary Models}

\paragraph{GSNet~\cite{ye2025towards}.} GSNet addresses open-vocabulary RS semantic segmentation through a dual-stream image encoder combining a frozen CLIP backbone with a specialist RS backbone. Query-guided feature fusion (QGFF) integrates the two streams, and a residual information preservation decoder (RIPD) aggregates multi-scale representations into dense predictions. The model is trained on LandDiscover50K, a dataset of 51,846 RS images spanning 40 categories.

\paragraph{Pi-Seg~\cite{li2026towards}.} Pi-Seg is a lightweight baseline for open-vocabulary RS image segmentation that enhances cross-domain transferability through a positive-incentive noise mechanism. It regularizes the vision-language alignment space of frozen CLIP encoders using two semantic-aware perturbation modules: a Text Semantic Perturbation Module (Text-SPM) that injects learnable residual perturbations into textual prototypes, and an Image Semantic Perturbation Module (Image-SPM) that applies text-guided cross-attention to generate adaptive, spatially-coherent perturbations in dense visual features. Depending on the evaluation benchmark, Pi-Seg is trained under different data scales: for the OVRSISBenchV1 cross-dataset protocol, it is trained using either the DLRSD dataset (5,601 images) or the iSAID dataset (18,076 images); for the application-oriented OVRSISBenchV2 protocol, it is trained on the large-scale, balanced OVRSIS95K dataset comprising 94,620 image-mask pairs.

\paragraph{SegEarth-OV~\cite{li2025segearthov1}.} SegEarth-OV is a training-free method that adapts frozen CLIP for open-vocabulary RS segmentation. SimFeatUp upsamples low-resolution CLIP patch features to higher resolution, and the global \texttt{[CLS]} token embedding is subtracted from all patch features to suppress scene-level bias. SimFeatUp is trained on 16K RS images in a self-supervised manner. No parameters are updated in CLIP.

\paragraph{SegEarth-OV3~\cite{li2025segearthov3}.} SegEarth-OV3 applies the SAM~3 inference pipeline to open-vocabulary RS segmentation without any training. SegEarth-OV3 produces outputs from three parallel heads: a Presence Head that outputs a scalar confidence score $S_\text{pres}$, an FCN-style Semantic Segmentation Head that outputs $P_\text{sem}$, and a Transformer-based Instance Head that outputs $P_\text{inst}$. The two spatial predictions are fused as $P_\text{fused} = \max(P_\text{sem},\, P_\text{inst\_agg})$, and the final prediction is $P_\text{final} = P_\text{fused} \times S_\text{pres}$, which suppresses predictions for absent categories.

\paragraph{SegEarth-R1~\cite{li2025segearthr1}.} SegEarth-R1 combines a frozen Swin-B visual encoder, a Phi-1.5 language model (1.3B parameters), and a Mask2Former decoder to support pixel-level segmentation from implicit natural language queries. The model is trained on EarthReason, a dataset of 5,434 RS image-mask pairs spanning 28 semantic categories with more than 30K question-answer annotations.

\paragraph{SegEarth-R2~\cite{xin2026segearthr2}.} SegEarth-R2 pairs a frozen Swin-B visual encoder with a Phi-2 language model (2.7B parameters) and a combined pixel decoder and Transformer decoder for hierarchical segmentation. Spatial attention supervision guides cross-attention toward small objects and fine-grained structures. The model is trained on LaSeRS, a dataset of 40,396 image-mask pairs covering 122 categories.

\paragraph{SkySense-O~\cite{zhu2025skysenseo}.} SkySense-O performs open-world RS interpretation using a SwinV2-H-8 vision encoder initialized from SkySense. To retain general visual representation quality, visual self-supervised paradigms (masked image modeling and contrastive learning) are incorporated during pre-training. For fine-tuning, a visual-relevance knowledge graph constructed via GPT-4V models category similarities to derive a vision-centric regularization loss ($\mathcal{L}_{vc}$) that resolves semantic alignment ambiguities. The model is trained on Sky-SA, containing 183,375 segment-text pairs across 1,763 category labels (~500 distinct categories).

\paragraph{GeoPixel~\cite{shabbir2025geopixel}.} GeoPixel is a pixel-grounding multimodal model for high-resolution RS images, supporting up to 4K resolution through dynamic image partitioning. The architecture combines a frozen CLIP ViT-L/14 visual encoder with an InternLM2-7B language model fine-tuned via partial LoRA on visual tokens, alongside a frozen SAM-2 visual encoder. The training dataset, GeoPixelD, contains 53,816 grounded phrases paired with 600,817 object masks drawn from RS scenes. Additionally, we evaluate GeoPixel-7B-RES, a variant fine-tuned on the RRSIS-D dataset (12,181 image-caption-mask triplets).

\paragraph{RemoteSAM~\cite{yao2025remotesam}.} RemoteSAM is a referring expression segmentation foundation model for RS that combines a Swin-Base visual encoder with a BERT text encoder and a fusion decoder, without incorporating a large language model (total 180M parameters). The model is trained on a curated collection of 270K image-text-mask triplets aggregated from multiple RS datasets including iSAID, LoveDA, DOTA, and HRRSD, covering a broad range of semantic categories.

\paragraph{UniGeoSeg~\cite{ni2026unigeoseg}.} UniGeoSeg supports instruction-driven segmentation across multiple RS paradigms, specifically referring, interactive, and reasoning segmentation, within a single architecture. The model pairs a frozen Swin-B visual encoder (initialized from Mask2Former pretrained weights) with a Phi-1.5 language model (1.3B parameters) and a Mask2Former segmentation decoder. Three key components are introduced: Task-Adaptive Text Enhancement (TATE) for refining instruction comprehension across heterogeneous tasks, Latent Knowledge Memory (LKM) for facilitating cross-task knowledge sharing, and Progressive Task Scheduling (PTS) as a training strategy to balance differences in task difficulty. The model is trained on GeoSeg-1M, a composite dataset of $\sim$1.15M image-mask-instruction triplets integrating 16 source datasets.

\paragraph{LAE-DINO~\cite{pan2024laedino}.} LAE-DINO is a DINO-based open-vocabulary detector developed for the ``Locate Anything on Earth''  task. To address RS domain gaps, it introduces Dynamic Vocabulary Construction (DVC) to dynamically sample positive and negative categories per training batch, and Visual-Guided Text Prompt Learning (VisGT) to align scene-level visual features with the text space. The model is trained on LAE-1M, a dataset containing one million annotated instances across 1,600 categories. This dataset is constructed by combining human-labeled datasets  with SAM- and LVLM-annotated unlabeled data.

\paragraph{OpenRSD~\cite{huang2025openrsd}.} OpenRSD is an open-prompt RS object detector based on RTMDet that supports both horizontal and oriented bounding box regression using multimodal (text and image) prompts. For the common HBB comparison reported in this paper, each predicted OBB is converted to its axis-aligned envelope; the original OBB output remains supported by the evaluator. OpenRSD adopts a dual-head architecture, comprising an alignment head for fast vocabulary scaling and a fusion head for deep cross-modal interaction. To facilitate robust cross-domain generalization, the model utilizes a multi-stage training pipeline consisting of pre-training, fine-tuning, and self-training. The detector is trained on ORSD+, a large-scale dataset constructed by the authors that encompasses 474,058 images spanning 200 categories. This dataset aggregates ten heterogeneous RS sources, such as DIOR-R, DOTA-v2.0, FAIR1M-2.0, fMoW, and WHU-Mix, alongside unlabeled data from Million-AID with pseudo-labels refined via a self-training loop filtered by SkyCLIP.

\paragraph{SLIP-RS~\cite{wang2026sliprs}.} SLIP-RS is a language-image pre-training framework that decomposes target semantics into a finite attribute space. It utilizes Structured-Attribute Contrastive Learning (SACL) for decoupled representation learning and the Conformal Attribute Reliability Engine (CARE) for statistical label filtration. Using the progressive ``Tune-Calibrate-Scale'' pipeline of CARE, the authors construct RS-Attribute-15M, a massive attribute-grounded dataset featuring over 15M high-fidelity instance-level attribute annotations across 31M bounding boxes. The data come from two streams: RS-Attribute-O, which enriches 14 public RS object detection datasets with attribute labels, and RS-Attribute-C, which leverages unlabeled images from diverse third-party channels to generate instance-level detection and attribute pseudo-labels.

\subsection{Inference Settings}

All methods are evaluated without training, adaptation, or threshold selection on OVEarth-Bench. Each query is submitted independently, and no ground-truth category list, mask, or box is exposed at inference. Predictions are mapped back to original-image coordinates before serialization. For segmentation scoring, multiple masks selected for one query are unioned into one binary mask. Detector outputs retain separate instances for box evaluation, while masks generated from their SAM prompts are unioned for segmentation evaluation. Missing query records, null masks, empty lists, and all-zero masks are treated as empty predictions. The evaluator is run with explicitly declared output modalities and no additional score filtering unless filtering is part of the model's official inference pipeline.

We use the default hyper-parameters specified in the official implementations or papers. For CLIP-based dense prediction models, CAT-Seg, GSNet, and Pi-Seg use a ViT-L/14 backbone; images are resized to 1008 pixels on the long side and processed with a sliding window of 336 pixels. CLIP, SCLIP, GEM, MaskCLIP, ClearCLIP, and SegEarth-OV use a ViT-B/16 backbone; images are resized to 896 pixels and processed with a sliding window of 224 pixels. Detection models use the confidence and NMS thresholds specified by each official implementation. All evaluations are performed on NVIDIA A100 or 4090 GPUs.

\section{Discussion and Limitations}
\label{app:discussion-limitations}

OVEarth-Bench reveals model behaviors that remain obscured within closed-set leaderboards. An analysis of category subsets demonstrates that broader category coverage renders model comparisons less sensitive to the selection of specific categories. Furthermore, the analysis of query forms uncovers model-specific variations, despite the presence of broadly similar performance rankings. The weak correlation between the localization of positive queries and the rejection of absent concepts further indicates that generating accurate masks does not guarantee reliable presence judgments. These findings advocate for the joint evaluation of category breadth, query diversity, localization accuracy, and presence prediction.

Among the evaluated configurations, methods based on MLLMs achieve the majority of the strongest results, particularly for referring and reasoning queries that necessitate compositional spatial understanding or functional inference. A plausible explanation is that extensive multimodal pretraining and instruction tuning offer transferable visual--semantic and linguistic representations that extend beyond the narrow scope of EO supervision. However, this interpretation remains descriptive rather than causal, as factors such as training data, model scale, output interfaces, external modules, and inference budgets vary significantly across the evaluated methods.

These observations motivate the integration of the semantic breadth of general MLLMs with EO-specific spatial competence. Promising directions include multiscale and high-resolution visual encoding for small targets, orientation-aware region and mask decoding, and EO-focused instruction tuning. Such tuning should cover long-tail categories, spatial relations, functional descriptions, and hard negative queries without eroding the transferability of general language capabilities. Furthermore, presence prediction should be treated as a calibrated objective rather than a mere by-product of localization. This requires explicit negative supervision alongside uncertainty-aware evaluation protocols. Controlled studies that align the MLLM backbone, training data, and inference budget while systematically varying EO adaptation and output representations are necessary. Such studies will help identify the specific components responsible for the observed performance gains and determine the conditions under which direct mask generation is preferable to bounding box grounding followed by a promptable segmenter.

Several limitations bound these conclusions. The proposed taxonomy does not exhaustively cover the open world, the geographic sampling remains uneven, and the reasoning queries address functions and visible context rather than the full scope of geospatial reasoning. Although the collection of new images reduces the risk of direct benchmark memorization, it cannot guarantee complete absence from the pretraining corpora of foundation models. The current evaluation focuses on categories and therefore does not assess localization at different levels of the taxonomy. Separately, detection comparisons primarily use HBBs because only a limited subset of the evaluated methods natively produces OBBs, limiting conclusions about orientation-aware localization. Moreover, due to data collection costs and the difficulty of acquiring examples for certain categories, such as specific aircraft, some categories contain only one or two samples, making their category-level results potentially sensitive to individual examples. We plan to address or mitigate these limitations in future releases of the benchmark.

\begin{table*}[!t]
  \centering
  \scriptsize
  \setlength{\tabcolsep}{2pt}
  \renewcommand{\arraystretch}{0.98}
  \begin{tabular}{p{0.17\textwidth}p{0.17\textwidth}p{0.25\textwidth}p{0.36\textwidth}}
    \toprule
    Hierarchy & Category & Positive vocabulary & Negative vocabulary \\
    \midrule
    \multicolumn{4}{l}{\textbf{Domain: Plant}} \\
    \addlinespace[1pt]
    Agriculture & center-pivot irrigation field & center-pivot irrigation field; circular irrigation field; pivot irrigation field; circular irrigated field & circular reservoir; solar panel array; stadium; natural oasis pond; circular industrial cooling pond; circular solar panel array; round water storage reservoir; sports stadium; circular industrial storage tank; large circular residential development; circular solar panel arrays; industrial circular storage tanks; man-made round water reservoirs; golf course greens; circular water reservoir; open-pit mining crater; circular solar array \\
    Agriculture & common cropland & cropland; crop field; farmland; agricultural field; common cropland & greenhouse complex; aquaculture ponds; vineyard; orchard; solar panel array; rice paddy terraces; center-pivot irrigation circle; wind turbine; airport runway; pastureland with grazing animals; greenhouse; rice paddy \\
    Agriculture & vineyard & vineyard; vineyard plot; grape vines & orchard; cornfield; greenhouse complex; solar panel array; olive grove; row crop field; rice terraces; trellis structures; rice paddies; winery building \\
    Agriculture > agricultural facilities & greenhouse & greenhouse; hoop house; glasshouse; greenhouses; polytunnel & solar panel array; glass conservatory; orchard; botanical conservatory; heated hothouse; industrial warehouse; parking lot; aircraft hangar; wind turbine; center-pivot irrigation system; grain silo; orchard (row-planted fruit trees); silo; cold frame; shade house; shipping container; row of cold frames \\
    Forestland & arbor & arbor; tree; wooded area; woodland & gazebo; trellis; greenhouse; pergola; orchard; vineyard; tree-lined avenue; tree nursery; avenue of trees; palm grove; hedgerow \\
    Forestland & shrub & shrub; bush & tree; large boulder; cactus; fence post \\
    Grassland & grass & grass; lawn & soccer field; golf course; crop field; pasture; wetland marsh; agricultural crop field; pasture with grazing animals; golf course green \\
    \midrule
    \multicolumn{4}{l}{\textbf{Domain: Business, service and residential}} \\
    \addlinespace[1pt]
    -- & bell tower & bell tower; church tower & water tower; windmill; radio mast; lighthouse \\
    -- & cable & cable; power line & suspension bridge cable; zip line; clothesline; conveyor belt; guy wire; railway track; tram tracks; underground pipeline; monorail; overhead pipeline \\
    -- & construction site & construction site; building site & railway yard; shipping container terminal; open-pit mine; large surface parking lot; airplane runway; quarry; landfill; solar panel farm; shipyard; airport runway; stadium; solar farm; port terminal; airstrip; rail yard; parking garage; demolition site; container port; sports stadium \\
    -- & ferris wheel & ferris wheel; observation wheel; amusement ride & carousel; roller coaster; observation tower; drop tower ride; merry-go-round; waterwheel; water tower; drop tower; rotating swing ride; wind turbine; swing ride \\
    -- & geyser & geyser; water fountain; fountain & reflecting pool; natural hot spring; sprinkler play area; swimming pool; volcanic crater; industrial cooling tower; sculpture; manhole cover; small gazebo; geothermal vent; water tower; sprinkler system; natural hot spring pool; industrial steam vent; oil drilling rig; children's splash pad \\
    -- & house & house; residential building; residence & mobile home park; industrial warehouse; shopping mall; skyscraper; agricultural barn; barn; greenhouse; mobile home; farmhouse; apartment building; row house (terraced housing); high-rise condominium; mobile home (trailer); apartment complex; attached row houses; high-rise tower \\
    \bottomrule
  \end{tabular}
  \caption{Category hierarchy and vocabulary of OVEarth-Bench.}
  \label{tab:category_hierarchy}
\end{table*}

\begin{table*}[!t]
  \ContinuedFloat
  \centering
  \scriptsize
  \setlength{\tabcolsep}{2pt}
  \renewcommand{\arraystretch}{0.98}
  \begin{tabular}{p{0.17\textwidth}p{0.17\textwidth}p{0.25\textwidth}p{0.36\textwidth}}
    \toprule
    Hierarchy & Category & Positive vocabulary & Negative vocabulary \\
    \midrule
    \multicolumn{4}{l}{\textbf{Domain: Business, service and residential}} \\
    \addlinespace[1pt]
    -- & lighthouse & lighthouse; beacon tower; light tower; lighthouse tower; beacon & water tower; smokestack; radio mast; watchtower; wind turbine; offshore oil platform; pier; harbor; cellular radio tower; church steeple; navigation buoy; harbor crane; radio antenna mast \\
    -- & sports ground & sports ground; athletic field & indoor swimming pool; ice hockey rink; golf course; stadium with large spectator stands \\
    -- & stele & stele; monument; obelisk & fountain; equestrian statue; clock tower; pagoda; lake; moutain; cropland; headstone; lighthouse; torii gate; statue; pool; football field \\
    -- & swimming pool & swimming pool; backyard pool & natural pond; hot tub; tennis court; indoor swimming pool; garden pond; public swimming pool; boat dock; hot tub (spa); decorative garden pond; blue tarp \\
    -- & village & village; rural settlement & industrial park; airport runway; large shopping mall; seaport or harbor; city center; industrial complex; railway station; high-rise apartment tower; large commercial shopping mall; large factory; shopping mall; seaport; highway interchange \\
    communications tower & dish communication tower & dish communication tower; satellite dish; parabolic antenna & wind turbine; solar panel array; water tower; lattice radio mast; mobile communication van; ground-mounted solar panel arrays; cellular phone tower; large ground-based parabolic antenna arrays; cell phone tower; guyed radio mast; high-voltage transmission tower; radio broadcast mast; microwave relay tower; large solar panel array \\
    communications tower & lattice communication tower & lattice communication tower; cell tower; cellular tower; communication tower; transmission tower & dish communication tower; steel transmission pylon; wind turbine; water storage tank; guyed radio mast; water tower; wind turbines; grain silo \\
    ferris wheel & shadow of ferris wheel & shadow of ferris wheel; ferris wheel shadow & shadow of carousel; shadow of roller coaster; shadow of observation tower; shadow of circus tent \\
    house & stilt house & stilt house; house on stilts; fishing hut; overwater bungalow & houseboat; floating cabin; lighthouse; mobile home; boathouse; skyscraper; apartment building; greenhouse; hotel; barn; garage \\
    house > with color & blue house & blue house; blue building & thatched-roof cottage; mobile home; glass greenhouse; isolated suburban bungalow with yard \\
    house > with color & red house & red house; red building & red barn; church building with a steeple \\
    house > with color & yellow house & yellow house; yellow building & mobile home; farm barn; thatched cottage \\
    sports ground & australian rules football field & australian rules football field; sports oval & athletics running track; baseball diamond; stadium with spectator stands; rectangular soccer pitch; soccer field; stadium with grandstands; golf course; athletics track; tennis courts; horse racing track; stadium with tiered stands; indoor swimming pool complex \\
    sports ground & badminton field & badminton field; badminton court & volleyball court; pickleball court; beach volleyball court; swimming pool; skatepark; pickleball courts \\
    sports ground & baseball field & baseball field; ballpark; baseball diamond & soccer field; american football field; cricket ground; athletics running track; tennis courts; cricket pitch; golf course; ice hockey rink; football field; baseball stadium; running track; softball diamond \\
    sports ground & basketball field & basketball field; basketball court & volleyball court; pickleball court; skatepark; tennis court; baseball diamond; handball court; skate park; outdoor swimming pool; netball court; indoor basketball gym; ice skating rink; covered sports arena; indoor gymnasium; swimming pool; ice rink; baseball field; ice hockey rink \\
    sports ground & diving board & diving board; springboard; pool diving board & water slide; gymnastics balance beam; pier; boat dock; springboard tower \\
    sports ground & golf course & golf course; golf course grounds & soccer field; baseball diamond; athletics running track; horse racetrack; crop field; equestrian center; race track; orchard; vineyard; soccer fields; baseball diamonds; amusement park; tennis courts; airstrip \\
    \bottomrule
  \end{tabular}
  \caption{Category hierarchy and vocabulary of OVEarth-Bench (continued).}
\end{table*}

\begin{table*}[!t]
  \ContinuedFloat
  \centering
  \scriptsize
  \setlength{\tabcolsep}{2pt}
  \renewcommand{\arraystretch}{0.98}
  \begin{tabular}{p{0.17\textwidth}p{0.17\textwidth}p{0.25\textwidth}p{0.36\textwidth}}
    \toprule
    Hierarchy & Category & Positive vocabulary & Negative vocabulary \\
    \midrule
    \multicolumn{4}{l}{\textbf{Domain: Business, service and residential}} \\
    \addlinespace[1pt]
    sports ground & gridiron & gridiron; football field & baseball diamond; tennis courts; ice hockey rink; rugby pitch; soccer stadium; cricket pitch; soccer pitch; lacrosse field; field hockey pitch \\
    sports ground & handball court & handball court & soccer field; baseball diamond; ice hockey rink; basketball court; swimming pool; volleyball court; pickleball court; skatepark \\
    sports ground & hockey field & hockey field; hockey pitch & outdoor basketball court; sand volleyball court; ice rink; baseball diamond; skatepark; rugby pitch; lacrosse field; outdoor ice rink; athletics running track; ice hockey rink; soccer field; american football field; cricket pitch; athletics track \\
    sports ground & korfball court & korfball court & tennis court; basketball court; handball court; beach volleyball court; netball court; volleyball court; basketball courts; netball courts; handball courts; baseball diamonds; baseball diamond; athletics running track; outdoor swimming pool; ice hockey rink; velodrome; baseball field \\
    sports ground & pickleball court & pickleball court & tennis court; basketball court; volleyball court; badminton court; padel court; sand volleyball court; handball court; outdoor badminton court; bocce court; platform tennis court \\
    sports ground & shot put & shot put; shot put circle; throwing circle & pickleball court; mountain; hammer throw cage; discus throwing cage; javelin runway; football field; baseball field; discus cage; long jump sand pit; discus circle; pole vault mat \\
    sports ground & soccer field & soccer field; football pitch & baseball diamond; tennis court; athletics running track; basketball court; cricket ground; ice hockey rink; cricket pitch; golf hole; lacrosse field; rugby field; rugby pitch \\
    sports ground & stadium & stadium; sports stadium; sports arena & baseball diamond; velodrome; horse racing track; horse racecourse; open-air amphitheater; baseball field; airport runway; harbor pier; container shipping terminal \\
    sports ground & tennis court & tennis court; tennis courts; tennis field & basketball court; volleyball court; pickleball court; badminton court; paddle tennis court; pickleball courts; outdoor basketball courts; sand volleyball courts; swimming pool; indoor tennis court; clay tennis court; padel courts; badminton courts \\
    sports ground & volleyball court & volleyball court & tennis court; badminton court; beach volleyball court; pickleball court; beach (sand) volleyball court; indoor wooden gym court; pickleball courts; baseball diamond; sand volleyball court; indoor volleyball court; boat; ship \\
    sports ground > baseball field & infield & baseball infield; infield & soccer field; american football field; cricket pitch; athletics running track; batting cage; baseball dugout; sand volleyball court; playground sandbox; horseshoe pit; golf green; tennis court; basketball court \\
    sports ground > baseball field & outfield & baseball outfield; sports outfield & soccer field; american football field; cricket pitch; athletics running track; golf course; horse racetrack; golf putting green; running track; tennis court; golf green \\
    sports ground > baseball field & pitcher's mound & pitcher's mound; pitching mound & cricket pitch; bowling green; tennis court; soccer penalty spot \\
    sports ground > basketball field & center circle & basketball center circle; basketball midcourt circle & soccer center circle; helipad landing pad; circular fountain; roundabout; helipad; rotunda; circular swimming pool; public fountain; round seating amphitheater; soccer field; bandstand; fountain \\
    \bottomrule
  \end{tabular}
  \caption{Category hierarchy and vocabulary of OVEarth-Bench (continued).}
\end{table*}

\begin{table*}[!t]
  \ContinuedFloat
  \centering
  \scriptsize
  \setlength{\tabcolsep}{2pt}
  \renewcommand{\arraystretch}{0.98}
  \begin{tabular}{p{0.17\textwidth}p{0.17\textwidth}p{0.25\textwidth}p{0.36\textwidth}}
    \toprule
    Hierarchy & Category & Positive vocabulary & Negative vocabulary \\
    \midrule
    \multicolumn{4}{l}{\textbf{Domain: Business, service and residential}} \\
    \addlinespace[1pt]
    sports ground > basketball field & restricted area & basketball restricted area & prison yard; military base; helipad; fuel storage tanks; security checkpoint; military barracks; ammunition depot; watchtower; guard tower; construction site; vehicle impound lot; military compound; tennis courts; tennis court; electrical substation; guard booth; storage container \\
    sports ground > golf course & fairway & golf fairway & soccer field; football field; airport runway; baseball diamond \\
    sports ground > golf course & putting green & golf green; putting green & flagstick; driving range target; golf cart; lawn bowling green; bowling green; soccer field; cricket pitch; baseball diamond; orchard; greenhouse; tennis courts; tennis court \\
    sports ground > golf course & sand trap & sand trap & sand volleyball court; construction sand pile; gravel parking area \\
    sports ground > golf course & tee box & golf teeing area; tee box; teeing ground & soccer field; helipad; river; pond; sea; algae \\
    tourist attraction & christ the redeemer & christ the redeemer; landmark statue & statue of liberty; eiffel tower; obelisk monument; equestrian statue; cathedral \\
    tourist attraction & eiffel tower & eiffel tower; observation tower & obelisk; church spire; radio broadcast tower; construction crane; clock tower \\
    tourist attraction & statue of liberty & statue of liberty; monument & lighthouse; obelisk; equestrian statue; triumphal arch \\
    tourist attraction & sydney opera house & sydney opera house; opera house & sports stadium; cathedral or large church; industrial shipyard \\
    tourist attraction & the colosseum & the colosseum; amphitheater & modern sports stadium; horse racing track; circular shopping mall \\
    \midrule
    \multicolumn{4}{l}{\textbf{Domain: Industrial, mining and storage}} \\
    \addlinespace[1pt]
    Energy & air-cooled condenser & air-cooled condenser; condenser bank & natural draft cooling tower; evaporative wet cooling tower; open cooling pond; solar photovoltaic array \\
    Energy & cooling tower & cooling tower; industrial cooling tower & industrial smokestack; water tower; grain silo; wind turbine; elevated water tower; smokestack; satellite dish array \\
    Energy & heliostat & heliostat; solar reflector; heliostat array; solar mirror & dish stirling concentrator; wind turbine; satellite communication dish; photovoltaic solar panels; satellite radio antenna array; tracking photovoltaic ground array; parabolic trough collectors; fresnel linear reflectors; satellite dishes; wind turbines \\
    Energy & heliostat field & heliostat field; solar power plant & center-pivot irrigation field; circular crop field; solar photovoltaic panel array; parabolic dish solar array; wind turbine farm \\
    Energy & molten salt storage tank & molten salt storage tank; storage tank & cooling tower; lng storage sphere; solar photovoltaic array; wind turbine \\
    Energy & pv panels & pv panels; solar farm; solar panels; solar arrays & rooftop solar panels; greenhouses; heliostat mirror field; wind turbines; ground-mounted solar farm; solar carport canopy; concentrated solar power tower; small wind turbine; industrial rooftop hvac units; solar thermal mirror arrays; solar carport canopies; rooftop photovoltaic panels on houses; carports; solar thermal collectors; shipping containers; roof-mounted solar panels; solar carports \\
    Energy & solar power tower & solar power tower; central receiver; central receiver tower; solar tower & wind turbine; ground-mounted photovoltaic farm; radio transmission mast; industrial cooling tower; hyperboloid cooling tower; ground-mounted photovoltaic panels; parabolic trough collector; solar dish concentrator; radio communications tower \\
    \bottomrule
  \end{tabular}
  \caption{Category hierarchy and vocabulary of OVEarth-Bench (continued).}
\end{table*}

\begin{table*}[!t]
  \ContinuedFloat
  \centering
  \scriptsize
  \setlength{\tabcolsep}{2pt}
  \renewcommand{\arraystretch}{0.98}
  \begin{tabular}{p{0.17\textwidth}p{0.17\textwidth}p{0.25\textwidth}p{0.36\textwidth}}
    \toprule
    Hierarchy & Category & Positive vocabulary & Negative vocabulary \\
    \midrule
    \multicolumn{4}{l}{\textbf{Domain: Industrial, mining and storage}} \\
    \addlinespace[1pt]
    Energy & substation & substation; transformer station; electrical substation & wind turbine farm; railway switching yard; large parking garage; construction equipment depot; solar panel array; open-air switchyard with lattice towers; power plant with cooling towers; wind farm; railway depot; transmission tower; wind turbine; railway station; residential house \\
    Energy & tower & tower; transmission tower & water tower; wind turbine; silo; lighthouse; oil derrick; grain silo; solar panel array \\
    Energy & wind turbine & wind turbine; wind turbine tower; windmill; turbine & traditional windmill; lattice transmission tower; guyed meteorological mast; radio communication antenna tower; solar panel array; electrical transmission pylon; radio communication mast; water tower; agricultural silo; radio antenna mast; construction crane; vertical-axis wind turbine; radio communications tower; decorative windmill; lighthouse; cellular tower; high-voltage transmission tower; electric transmission tower; cellular communication mast; grain silo \\
    Energy > wind turbine & shadow of wind turbine & shadow of wind turbine; wind turbine shadow & tree shadow; building shadow; crane shadow; airplane shadow; cell tower shadow; shadow of high-voltage transmission tower; shadow of traditional windmill; shadow of small airplane; shadow of multi-story barn; solar panel array; high-voltage transmission tower; construction crane shadow \\
    Energy > wind turbine & wind turbine blade & wind turbine blade; turbine blade & airplane wing; helicopter rotor; crane boom; solar panel array; airplane; helicopter; satellite dish \\
    Industrial & chimney & chimney; smokestack & cooling tower; water tower; grain silo; radio mast; observation tower; electrical transmission tower; silo; wind turbine; flare stack; residential house chimney; construction crane \\
    Industrial & crane & crane; container crane & tower crane; truck-mounted crane; wind turbine; mobile crane truck; shipboard derrick; excavator; mobile truck crane; floating crane barge; offshore drilling rig; floating crane; reach stacker \\
    Industrial & pipeline & pipeline; piping; pipe network; oil pipeline & railway tracks; conveyor belts; overhead power transmission towers; irrigation canals; conveyor belt; overhead power lines; stormwater drain channel; sewer main; railway track; overhead power line; solar panel array \\
    Industrial & ship loader & ship loader; ship unloader & floating crane barge; bucket-wheel reclaimer; stacker-reclaimer; mobile harbor crane \\
    Industrial & storage tank & storage tank; water tank; water storage tank; fuel tank & grain silo; elevated water tower; tanker truck; industrial oil storage farm; water tower; oil storage dome; industrial smokestack; agricultural silo; circular swimming pool; oil storage tank farm; propane tank; domed gas holder; fuel tanker truck; cooling tower; swimming pool; gas holder (gasholder) \\
    Mining and quarrying & headframe & headframe; mining headframe; mine headframe & water tower; construction crane; wind turbine; industrial silo; cellular tower; observation tower \\
    Mining and Quarrying & mine & mine; open-pit mine; mining site & solar farm; wind farm; rail yard; residential neighborhood; landfill; tailings pond; smelting plant; solar panel array; airport runway; crop fields; solar panel farm; sewage treatment plant; natural freshwater lake; wind turbine array; residential subdivision; railway yard \\
    \midrule
    \multicolumn{4}{l}{\textbf{Domain: Transportation}} \\
    \addlinespace[1pt]
    -- & bicycle parking & bicycle parking; bike rack & painted bike lane; covered bicycle shed; bicycle locker; motorcycle parking area; curbside bicycle corral \\
    \bottomrule
  \end{tabular}
  \caption{Category hierarchy and vocabulary of OVEarth-Bench (continued).}
\end{table*}

\begin{table*}[!t]
  \ContinuedFloat
  \centering
  \scriptsize
  \setlength{\tabcolsep}{2pt}
  \renewcommand{\arraystretch}{0.98}
  \begin{tabular}{p{0.17\textwidth}p{0.17\textwidth}p{0.25\textwidth}p{0.36\textwidth}}
    \toprule
    Hierarchy & Category & Positive vocabulary & Negative vocabulary \\
    \midrule
    \multicolumn{4}{l}{\textbf{Domain: Transportation}} \\
    \addlinespace[1pt]
    -- & parking lot & parking lot; parking area; car park & multi-level parking garage; bus depot; car wash; drive-through restaurant lane; multilevel parking garage; gas station with pumps; rooftop parking deck; gas station; truck loading yard; car dealership lot; parking garage; multi-level parking structure; bus terminal; truck loading dock; multi-story parking garage; covered parking structure; gas station with fuel pumps \\
    -- & traffic light & traffic light; traffic signal & overhead signal gantry; bus stop shelter; railroad crossing signal; pedestrian shelter or kiosk; stop sign \\
    road & asphalt road & asphalt road; highway; paved road & railway tracks; airport runway; gravel road; elevated bridge overpass; roundabout; unpaved dirt road; cobblestone street; tram tracks; large parking lot; sidewalk; railway track; concrete runway; parking lot; bridge \\
    road & chevron marking & chevron marking; chevron road marking; road chevrons & speed bump; railway tracks; painted pedestrian crosswalk; stop line marking; yield triangle marking; parking bay markings; pedestrian crosswalk; bicycle lane; parking stall markings; chevron road sign; parking stall lines; railroad crossing marking; zebra crosswalk; bicycle lane marking; speed hump; bus stop bay marking \\
    road & crossroad & crossroad; intersection; road intersection & roundabout; overpass; railway crossing; grade-separated overpass; railway level crossing; toll plaza; highway interchange; pedestrian overpass; tram tracks; cloverleaf interchange; grade-separated highway junction; bus depot; grade-separated interchange; pedestrian plaza; elevated highway; pedestrian-only plaza \\
    road & dirt road & dirt road; unpaved road; unsealed road; dirt track & paved road; railway track; concrete runway; river or stream; sidewalk; asphalt road with lane markings; concrete parking lot; railway tracks; paved residential driveway; cobblestone street; asphalt road; parking lot; concrete bridge \\
    road & highway service area & highway service area; rest area; service station; truck stop & railway station; small airport terminal; residential neighborhood; solar panel farm; toll plaza; truck weigh station; bus terminal; campground; weigh station; airport terminal; industrial warehouse; multi-level parking garage; multi-storey parking garage \\
    road & highway toll station & highway toll station; toll plaza & gas station; weigh station; rest area; parking garage; service station; truck weigh station; bus terminal; parking lot \\
    road & median strip & median strip; road median; center divider & tram tracks; railway tracks; bus stop shelter; bike lane; jersey barrier; on-street parking lane; bus-only lane; roundabout; railway line; pedestrian refuge island \\
    road & mountain road & mountain road; winding road; rural road & river; railway track; hiking trail; irrigation canal; airport runway; river channel; multi-lane highway; parking lot; ski lift \\
    road & overpass & overpass; road overpass; cloverleaf interchange; road bridge; road interchange; highway interchange; flyover & suspension bridge; drawbridge; ferry terminal; railway station; pedestrian footbridge; railway bridge; road tunnel entrance; railway viaduct; canal aqueduct; pedestrian bridge; toll plaza; river bridge; rail viaduct \\
    road & railroad crossing & railroad crossing; level crossing; rail crossing & pedestrian bridge; road overpass; train station platform; rail bridge; train station; passenger platform; pedestrian footbridge; railway station; train platform; signal mast; pedestrian crosswalk; road overpass bridge; automatic crossing gate; rail tunnel entrance; freight loading ramp; railway platform; railroad bridge; overpass; switchyard \\
    \bottomrule
  \end{tabular}
  \caption{Category hierarchy and vocabulary of OVEarth-Bench (continued).}
\end{table*}

\begin{table*}[!t]
  \ContinuedFloat
  \centering
  \scriptsize
  \setlength{\tabcolsep}{2pt}
  \renewcommand{\arraystretch}{0.98}
  \begin{tabular}{p{0.17\textwidth}p{0.17\textwidth}p{0.25\textwidth}p{0.36\textwidth}}
    \toprule
    Hierarchy & Category & Positive vocabulary & Negative vocabulary \\
    \midrule
    \multicolumn{4}{l}{\textbf{Domain: Transportation}} \\
    \addlinespace[1pt]
    road & railway & railway; railway tracks; railway track; train tracks & canal; river; airport runway; tram line; subway tunnel; highway interchange; pipeline; tram tracks; freight yard; streetcar stop; subway entrance; railway turntable; freight depot \\
    road & railway traverser & railway traverser; transfer table; track traverser & rail turntable; roundhouse; wagon tippler; container gantry crane; locomotive maintenance pit; rail-mounted gantry crane; wheel lathe; turntable; rail-mounted crane; locomotive; inspection pit \\
    road & railway turntable & railway turntable; locomotive turntable; turntable & railway wye; turning loop; rail-mounted crane; wye junction; overhead gantry crane; circular engine shed with radial stalls; coaling tower; railway roundhouse; water tower; roundhouse; passenger platform \\
    road & roundabout & roundabout; traffic circle & cloverleaf interchange; toll plaza; bus terminal; rotary with pedestrian plaza and benches; roundabout with tram tracks; circular fountain; round pond; helipad; rotunda plaza; fountain; circular plaza; statue on a plinth; bus turnaround; circular pedestrian plaza with benches; tramway circle with embedded tracks; cloverleaf highway ramp; landscaped central park; small pond; bus turning loop; rotary with decorative statue \\
    road & streetlight & streetlight; lamp post & traffic signal; utility pole with power lines; flagpole; tram catenary pole \\
    road > sidewalk & pedestrian crossing & pedestrian crossing; crosswalk; zebra crossing & pedestrian overpass; pedestrian underpass; bicycle lane; railroad crossing; speed bump; bike lane; bicycle box; bus stop shelter; pedestrian footbridge; subway entrance; raised speed table; protected bike lane \\
    station & bus stop & bus stop; bus shelter & tram stop; streetcar platform; taxi rank; parked bus \\
    station & helipad & helipad; helicopter landing pad; landing pad; rooftop helipad & seaplane dock; marina; ferry terminal; floating restaurant; airport runway; roundabout; circular fountain; satellite dish array; runway; seaplane ramp; floatplane dock; air traffic control tower; small aircraft hangar; rooftop helipad; small marina with slips; helicopter refueling station; road roundabout; water tower; parked helicopter; rooftop hangar; airplane runway; control tower \\
    station & slipway & slipway; boat ramp & travel lift (boat hoist); ferry terminal; dry dock; marine railway; boat lift; breakwater; boat hoist; ship lift; travel lift; floating pontoon; marina; boathouse; shipyard crane \\
    station & train station & train station; railway terminal; railway station; rail depot & airport terminal; freight terminal; bus depot; tram depot; freight yard; parking garage; ferry terminal; container port; seaport container terminal; container shipping terminal; oil refinery; freight cargo depot; sports stadium; convention center \\
    station > airport & control tower & control tower; airport control tower & radar dome; fuel storage tank; water tower \\
    station > airport & terminal & terminal; airport terminal & railway station; seaport container terminal; stadium; bus depot; aircraft hangar; cargo warehouse; fuel storage farm; seaport ferry terminal; hangar; cargo terminal; control tower \\
    vehicle & bulldozer & bulldozer & road roller; motor grader; tower crane; concrete mixer truck \\
    vehicle & bus & bus; coach; city bus & tram; light rail vehicle; train carriage; trolleybus \\
    vehicle & car & car; parked car; vehicle & bus; motorcycle; bicycle; tractor; truck \\
    vehicle & excavator & excavator; backhoe loader & tower crane; concrete mixer truck; road roller; forklift; bulldozer; wheel loader; articulated dump truck; skid-steer loader \\
    vehicle & plane & plane; airplane & helicopter; glider; drone; hot air balloon; quadcopter drone; seaplane \\
    \bottomrule
  \end{tabular}
  \caption{Category hierarchy and vocabulary of OVEarth-Bench (continued).}
\end{table*}

\begin{table*}[!t]
  \ContinuedFloat
  \centering
  \scriptsize
  \setlength{\tabcolsep}{2pt}
  \renewcommand{\arraystretch}{0.98}
  \begin{tabular}{p{0.17\textwidth}p{0.17\textwidth}p{0.25\textwidth}p{0.36\textwidth}}
    \toprule
    Hierarchy & Category & Positive vocabulary & Negative vocabulary \\
    \midrule
    \multicolumn{4}{l}{\textbf{Domain: Transportation}} \\
    \addlinespace[1pt]
    vehicle & ship & ship; boat & yacht; submarine; oil platform; sailboat; offshore oil platform; military warship; aircraft carrier; oil tanker \\
    vehicle & train & train; freight train; passenger train; railcar & truck convoy; articulated bus; tram; pipeline; long line of parked cars; subway train; monorail; passenger platform; high-speed bullet train; passenger train; streetcar; subway car; light rail vehicle; freight car \\
    vehicle & truck & truck; box truck; semi-truck & bus; car; motorhome; camper van; pickup truck; delivery van \\
    vehicle > car > with color & pink car & pink car & motorcycle; dam; school bus \\
    vehicle > car > with color & red car & red car & convertible; motorcycle; bicycle; scooter; red bus; red van; red motorcycle; fire truck; red pickup truck; red taxi; red fire engine \\
    vehicle > car > with color & white car & white car & white bus \\
    vehicle > plane & compass rose & compass rose; directional marker; direction marker; orientation marker; navigation marker & helipad; windsock; radar dome; vor station; helipad marking (h); roundabout traffic circle; roundabout; sundial; decorative fountain plaza; decorative fountain; anchor monument; helipad marking; circular decorative fountain; survey control disk \\
    vehicle > plane & hangar & hangar; aircraft hangar & air traffic control tower; airport passenger terminal; cargo warehouse; multistory parking garage; control tower; passenger terminal; fuel storage tanks; terminal concourse; jet bridge \\
    vehicle > plane & helicopter & helicopter; rotorcraft; aircraft & blimp; hot air balloon; seaplane; tiltrotor aircraft (osprey); quadcopter drone; fixed-wing airplane; unmanned aerial vehicle (drone); airliner; airship (blimp); ship; wind turbine; tiltrotor aircraft \\
    vehicle > plane & tiltrotor & tiltrotor; tiltrotor aircraft & seaplane; blimp; quadrotor drone; commercial passenger jet \\
    vehicle > plane > with color & blue plane & blue plane & private propeller plane; helicopter; seaplane; military fighter jet \\
    vehicle > plane > with color & orange plane & orange plane & helicopter; seaplane; hot air balloon; biplane \\
    vehicle > plane > with color & red plane & red plane & helicopter; seaplane; glider; biplane; blue plane; drone; pink plane; red helicopter; hot air balloon \\
    vehicle > ship & hospital ship & hospital ship; medical ship & cruise ship; oil tanker; submarine \\
    vehicle > ship > warship & aircraft carrier & aircraft carrier & cruise ship; oil tanker; commercial ferry; fish; swimming pool; submarine; ferry \\
    vehicle > ship > warship & war ship & war ship; naval vessel & cargo ship; oil tanker; passenger ferry; fishing trawler; submarine; sailboat; cruise ship; container ship; aircraft carrier \\
    vehicle > ship > with color & red ship & red ship; red boat & aircraft carrier; submarine; offshore oil platform; white sailboat; white kayak; black pedal boat \\
    vehicle > truck & pickup & white pickup; white pickup truck & bus; tractor-trailer \\
    vehicle > truck > with color & green semi-trailer truck & green semi-trailer truck & red box truck; refrigerated semi-trailer (reefer); flatbed trailer carrying visible cargo; articulated city bus \\
    vehicle > truck > with color & orange semi-trailer truck & orange semi-trailer truck; orange semi-truck & cement mixer truck; fire engine; double-decker bus; rear-loader garbage truck \\
    vehicle > truck > with color & red semi-trailer truck & red semi-trailer truck & red fuel tanker truck; flatbed truck; refrigerated trailer; dump truck \\
    \midrule
    \multicolumn{4}{l}{\textbf{Domain: Terrain}} \\
    \addlinespace[1pt]
    -- & bareland & bareland; vacant lot & sports field; pond; orchard; railway track \\
    \bottomrule
  \end{tabular}
  \caption{Category hierarchy and vocabulary of OVEarth-Bench (continued).}
\end{table*}

\begin{table*}[!t]
  \ContinuedFloat
  \centering
  \scriptsize
  \setlength{\tabcolsep}{2pt}
  \renewcommand{\arraystretch}{0.98}
  \begin{tabular}{p{0.17\textwidth}p{0.17\textwidth}p{0.25\textwidth}p{0.36\textwidth}}
    \toprule
    Hierarchy & Category & Positive vocabulary & Negative vocabulary \\
    \midrule
    \multicolumn{4}{l}{\textbf{Domain: Terrain}} \\
    \addlinespace[1pt]
    -- & beach & beach; sand beach & picnic table; lifeguard tower; boardwalk kiosk; boardwalk; low concrete wall; deck chair; sun lounger; playground swing set; stone beach; concrete seating block; boat; trash receptacle; bicycle rack \\
    -- & cape & cape; headland & pier; harbor; marina; lighthouse; breakwater \\
    -- & ice & ice; glacier & iceberg; pack ice; frozen river; artificial ice rink \\
    -- & island & island; islet & oil platform; lighthouse; fishing boat; marina \\
    -- & marsh & marsh; wetland & open lake; meandering river channel; pond; mangrove forest \\
    -- & mountain & mountain; mountain ridge & volcano; glacier; snow-capped peak; lava flow; sand dune \\
    -- & volcanic vent & volcanic vent; crater; crater lake & glacier; geothermal power plant; geyser field; impact crater; crater lake; open-pit mine; lava dome; hot spring pool; industrial quarry \\
    \midrule
    \multicolumn{4}{l}{\textbf{Domain: Water infrastructure}} \\
    \addlinespace[1pt]
    -- & breakwater & breakwater; jetty; harbor breakwater & offshore wind turbine; oil drilling platform; container terminal; offshore oil platform; causeway; groin; floating pontoon; coral reef; sandbar shoal; floating breakwater; marina for pleasure boats; natural sandbar \\
    -- & bridge & bridge; road bridge; causeway & rail bridge; causeway; pedestrian footbridge; railway bridge; floating bridge; tunnel; covered wooden bridge; aqueduct; covered bridge; truss bridge \\
    -- & dam & dam; weir; earthen dam; spillway & suspension bridge; steel truss bridge; navigation lock; floating pontoon bridge; hydroelectric power station; fish ladder; boat marina; canal; hydroelectric power plant; earthen levee; sluice gates; boat launch ramp \\
    -- & fish raft house & fish raft house; floating house; floating fish farm & offshore oil platform; large commercial cargo ship; marina with luxury yachts; floating solar farm; moored cargo barge; offshore wind turbine; marina with docked yachts \\
    -- & fishpond & fishpond; aquaculture pond; pond & marina; natural lake; recreational swimming pool; harbor; salt evaporation pond; swimming pool; concrete dam reservoir; wetland marsh; rice paddies; marina docks; recreational swimming pools; sewage treatment basins; large natural lake; concrete reservoir with spillway; concrete reservoir with dam; irrigation canal \\
    -- & floating aquaculture cage & floating aquaculture cage; fish cage; fish farm & offshore oil platform; moored barge; fish processing vessel; offshore wind turbine; oil drilling platform; floating solar panel array; large container ship; cruise ship; marina with moored boats; fishing trawler fleet \\
    -- & inverted siphon & inverted siphon; culvert & lock chamber; aqueduct; bridge with visible girders; boat ramp \\
    -- & lake & lake; water body; pond; body of water & fabric; coastal bay; marina; river; canal; pier; boathouse; swimming pool; reservoir; ocean \\
    -- & moat & moat; water feature; castle moat & canal; river; swimming pool; natural river; natural lake; swamp; canal lock; boathouse; marina; dry moat \\
    -- & navigation lock & navigation lock; boat lock; canal lock & weir; hydroelectric power station; swing bridge; floating marina; dry dock; drawbridge; hydroelectric dam; boat lift; fish ladder; shipyard; ferry terminal; lighthouse; cargo terminal; ship lift; canal basin \\
    -- & pier & pier; boat pier; wharf & ferry terminal; breakwater; floating oil platform; swing bridge; floating pontoon; commercial ship terminal; floating dock; industrial shipyard; floating bridge; lighthouse; offshore oil platform; suspension bridge; offshore wind farm \\
    \bottomrule
  \end{tabular}
  \caption{Category hierarchy and vocabulary of OVEarth-Bench (continued).}
\end{table*}

\begin{table*}[!t]
  \ContinuedFloat
  \centering
  \scriptsize
  \setlength{\tabcolsep}{2pt}
  \renewcommand{\arraystretch}{0.98}
  \begin{tabular}{p{0.17\textwidth}p{0.17\textwidth}p{0.25\textwidth}p{0.36\textwidth}}
    \toprule
    Hierarchy & Category & Positive vocabulary & Negative vocabulary \\
    \midrule
    \multicolumn{4}{l}{\textbf{Domain: Water infrastructure}} \\
    \addlinespace[1pt]
    -- & port container & port container; shipping container; container on land; container at the port & rail freight cars; oil storage tanks; sailboats; flatbed trucks; rail freight wagons; flatbed truck trailers; iso tank containers; wooden shipping crates; storage silos \\
    -- & river & river; water; waterway; stream; river channel; yellow river & lake; reservoir; pond; canal; oxbow lake; swimming pool; marina; waterfall \\
    -- & sea & sea; ocean; water; coastal water & lake; lagoon; harbor; pier; marina; iceberg; cargo ship; river; estuary; inland lake; wetland marsh; volcano; sky; fish \\
    -- & sluice gate & sluice gate; irrigation gate & weir; boat lock; pumping station \\
    floating aquaculture cage & fish farm cage & fish farm cage & moored pontoon dock; anchored cargo barge; navigation buoy field \\
    floating aquaculture cage & floating-raft aquaculture & floating-raft aquaculture & floating solar panels; offshore wind turbines; oil drilling platform \\
    wastewater plant & aeration tank & aeration tank; aeration basin & domed anaerobic digester; cylindrical storage tank; industrial cooling tower; elevated water tower; trickling filter; cooling tower; evaporation pond; membrane bioreactor modules; orchard; solar panel array; open-air swimming pool; wind turbine farm; grain silo; outdoor swimming pool; waterfall; well \\
    wastewater plant & chlorine contact tank & chlorine contact tank; contact tank & uv disinfection unit; sand filtration bed; sludge drying bed; mobile water tanker truck \\
    wastewater plant & clarifier & clarifier; clarifier tank & cooling towers; oil storage tanks; grain silos; circular swimming pools; water towers; cooling tower; gasholder; circular swimming pool; residential swimming pools; parabolic solar dishes; biogas digester; grain silo; oil storage tank; swimming pool; gas holder \\
    wastewater plant & sand filter & sand filter; filter bed; filter beds & pressure filter vessel; membrane bioreactor unit; cooling tower; chlorine contact tank; trickling filter; membrane bioreactor (mbr) unit; rotating biological contactor (rbc) \\
    wastewater plant & sludge thickener & sludge thickener; thickener tank & swimming pool; cooling tower \\
    \midrule
    \multicolumn{4}{l}{\textbf{Domain: Special}} \\
    \addlinespace[1pt]
    military & armoured fighting vehicle & armoured fighting vehicle; tank; military vehicle & armoured personnel carrier; self-propelled artillery; infantry fighting vehicle; military cargo truck; main battle tank; civilian pickup truck; self-propelled howitzer; tank transporter; armoured recovery vehicle; mobile surface-to-air missile launcher; amphibious assault vehicle; armoured bridgelayer; towed artillery piece \\
    military & bunker & bunker; defensive emplacement; military fortification; military bunker; concrete bunker & radar antenna; aircraft hangar; watchtower; fuel storage tank; radar installation; silo; storm shelter; pillbox; air-raid shelter; storm cellar; ammunition cache \\
    \bottomrule
  \end{tabular}
  \caption{Category hierarchy and vocabulary of OVEarth-Bench (continued).}
\end{table*}

\begin{table*}[!t]
  \ContinuedFloat
  \centering
  \scriptsize
  \setlength{\tabcolsep}{2pt}
  \renewcommand{\arraystretch}{0.98}
  \begin{tabular}{p{0.17\textwidth}p{0.17\textwidth}p{0.25\textwidth}p{0.36\textwidth}}
    \toprule
    Hierarchy & Category & Positive vocabulary & Negative vocabulary \\
    \midrule
    \multicolumn{4}{l}{\textbf{Domain: Special}} \\
    \addlinespace[1pt]
    military & military airfield & military airfield; air base; airbase & civilian airport terminal; commercial passenger concourse; large aircraft hangar complex; railway freight yard; commercial passenger terminal; civilian regional airport; cargo seaport; motor racing circuit; container port and ship terminal; aircraft carrier; civilian glider club airstrip; commercial airport terminal; civilian passenger airport; seaport with docks; railway yard; wind farm; commercial passenger terminal with jet bridges; international airport with multiple concourses; major cargo container yard and freight apron; civilian municipal grass-strip airfield with only small hangars; civilian international airport terminal; container shipping port; freight rail classification yard; large commercial solar farm; civilian international airport; civilian municipal airport; container seaport terminal; rail freight yard \\
    military > airplane & beechcraft 1900c & beechcraft 1900c & cessna caravan; beechcraft king air; private business jet; helicopter \\
    military > airplane & boeing c-17 & boeing c-17; military transport aircraft & lockheed c-130 hercules; boeing 747; airbus a380; air-to-air refueling tanker (kc-135) \\
    military > airplane & boeing c-97g & boeing c-97g; transport aircraft & douglas c-124 globemaster ii; lockheed c-130 hercules; douglas dc-4 skymaster; boeing 737 commercial jet \\
    military > airplane & c-130 & c-130; cargo plane & helicopter; commercial jet airliner; c-17 globemaster; seaplane \\
    military > airplane & c-47 & c-47 aircraft & helicopter; glider; seaplane \\
    military > airplane & f-35 & f-35; fighter jet & f-16; f-22 raptor; f/a-18 hornet; c-130 cargo plane \\
    military > airplane & fokker 50 & fokker 50 & dash 8; cessna 172; helicopter; boeing 737 \\
    military > airplane & rc12 & rc12; aircraft & helicopter; glider; seaplane \\
    military > airplane & uh60 & uh60 helicopter; helicopter & ship; v-22 osprey; c-130 hercules \\
    military > airplane & v-22 & v-22; tiltrotor aircraft & helicopter; ch-47 chinook; c-130 hercules; single-engine propeller plane; seaplane; train; airplane model \\
    religion & cathedral & cathedral; church & mosque; synagogue; basilica; castle; sports stadium; stadium; train station; abbey; monastery; palace \\
    religion & stupa & stupa; Buddhist monument; pagoda & pagoda; minaret; bell tower; obelisk; watchtower; water tower \\
    tomb & mausoleum & mausoleum; memorial complex; family tomb; cemetery; tomb & palace; stadium; industrial plant; skyscraper; church; columbarium; mortuary; family vault; funeral chapel; catacomb entrance; chapel; temple; pagoda; barn; boathouse \\
    tomb & pyramid & pyramid; pyramid structure; Egyptian pyramid; step pyramid & obelisk; ziggurat; pagoda; volcanic cone; modern high-rise building \\
    \bottomrule
  \end{tabular}
  \caption{Category hierarchy and vocabulary of OVEarth-Bench (continued).}
\end{table*}

\begin{figure*}[t]
  \centering
  \includegraphics[width=\linewidth]{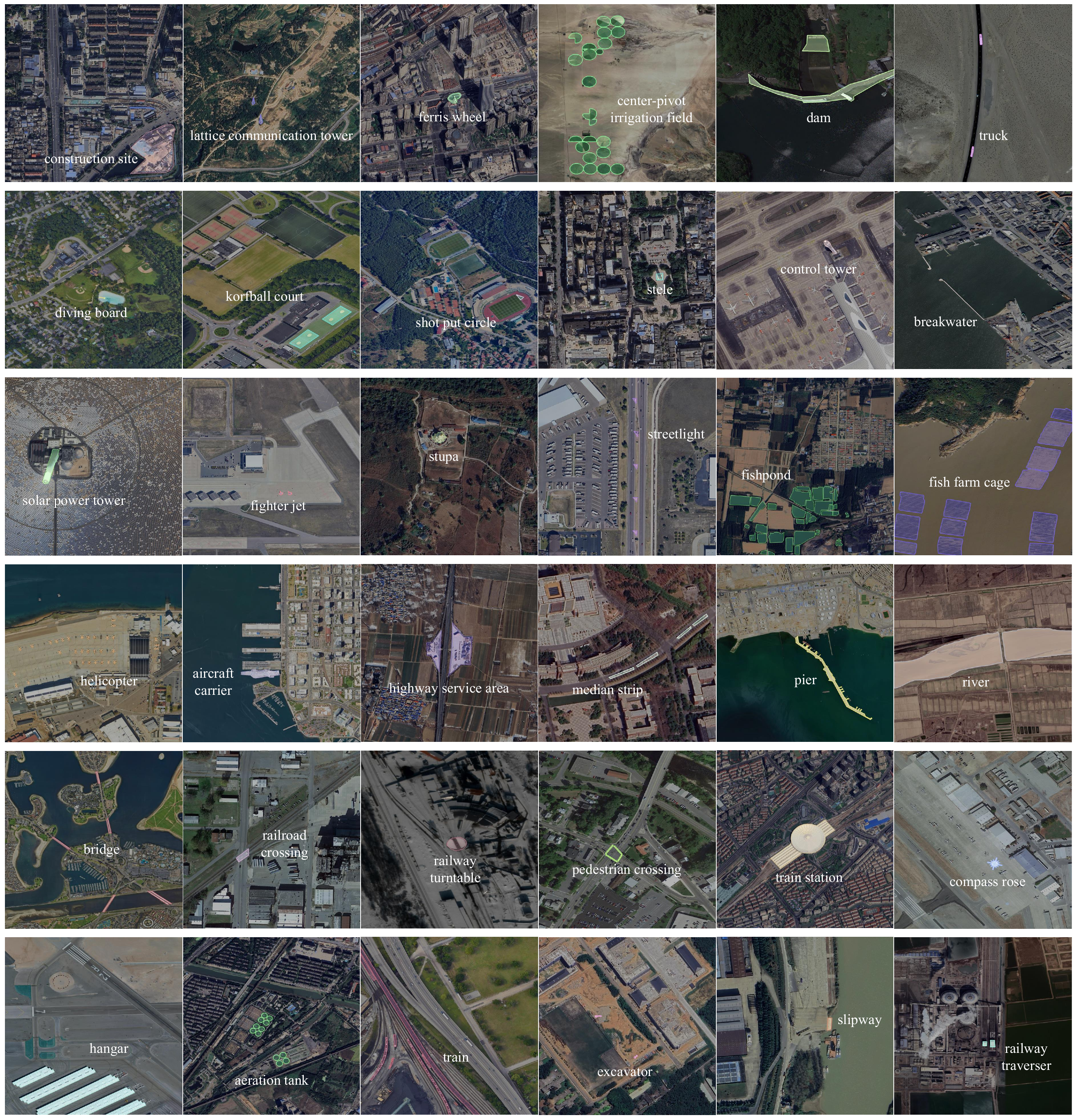}
  \caption{Additional visual examples from OVEarth-Bench, illustrating the breadth of annotated EO concepts across \emph{``plant''}, \emph{``business, service and residential''}, \emph{``industrial, mining and storage''}, \emph{``transportation''}, \emph{``terrain''}, \emph{``water infrastructure''}, and \emph{``special facilities''}.}
  \label{fig:sample3}
\end{figure*}

\end{document}